%% file: main.tex
\documentclass[10pt,journal,compsoc]{IEEEtran}
\ifCLASSOPTIONcompsoc
  \usepackage[nocompress]{cite}
\else
  \usepackage{cite}
\fi
\usepackage{amsmath}
\usepackage{amssymb}
\usepackage{mathtools}
\usepackage{stmaryrd}
\usepackage{graphicx}
\usepackage{algorithmic}
\usepackage{array}
\usepackage{multirow}
\usepackage{url}
\usepackage{hyperref}
\usepackage{color}
\usepackage{colortbl}
\usepackage{framed}
\usepackage{subcaption}
\usepackage{balance}
\usepackage{booktabs}
\usepackage{bm}
\usepackage{bbm}
\usepackage{xspace}
\usepackage{lipsum}
\usepackage{enumitem}
\usepackage[dvipsnames,table]{xcolor}
\usepackage{pifont}

\usepackage{tabularx}
\usepackage{makecell}

\usepackage{scrextend}
\usepackage{enumitem}
\usepackage{graphicx}

\def \ie {\emph{i.e.}}
\def \eg {\emph{e.g.}}
\def \etal {\emph{et al.}}

\newcommand{\tit}[1]{\smallbreak\noindent\textbf{#1.}}
\newcommand{\tyt}[1]{\textit{#1 --}}
\newcommand{\tinytit}[1]{\noindent\textbf{#1.}}

\newcommand{\cmark}{\ding{51}}%

\begin{document}
%
\title{From Show to Tell: A Survey on\\ Deep Learning-based Image Captioning}

%
%
%

\author{Matteo~Stefanini,
        Marcella~Cornia,
        Lorenzo~Baraldi,
        Silvia~Cascianelli,\\
        Giuseppe~Fiameni,
        and~Rita~Cucchiara
\IEEEcompsocitemizethanks{\IEEEcompsocthanksitem M. Stefanini, M. Cornia, L. Baraldi, S. Cascianelli, and R. Cucchiara are with the Department of Engineering ``Enzo Ferrari'', University of Modena and Reggio Emilia, Modena, Italy.\protect\\
E-mail: \{matteo.stefanini, marcella.cornia, lorenzo.baraldi, silvia.cascianelli,  rita.cucchiara\}@unimore.it.
\IEEEcompsocthanksitem G. Fiameni is with NVIDIA AI Technology Centre, Italy.\protect\\
E-mail: gfiameni@nvidia.com}
}

\IEEEtitleabstractindextext{%
\begin{abstract}
Connecting Vision and Language plays an essential role in Generative Intelligence. For this reason, large research efforts have been devoted to image captioning, \ie~describing images with syntactically and semantically meaningful sentences. Starting from 2015 the task has generally been addressed with pipelines composed of a visual encoder and a language model for text generation. During these years, both components have evolved considerably through the exploitation of object regions, attributes, the introduction of multi-modal connections, fully-attentive approaches, and BERT-like early-fusion strategies. However, regardless of the impressive results, research in image captioning has not reached a conclusive answer yet. This work aims at providing a comprehensive overview of image captioning approaches, from visual encoding and text generation to training strategies, datasets, and evaluation metrics. In this respect, we quantitatively compare many relevant state-of-the-art approaches to identify the most impactful technical innovations in architectures and training strategies. Moreover, many variants of the problem and its open challenges are discussed. The final goal of this work is to serve as a tool for understanding the existing literature and highlighting the future directions for a research area where Computer Vision and Natural Language Processing can find an optimal synergy.
\end{abstract}

\begin{IEEEkeywords}
Image Captioning, Vision-and-Language, Deep Learning, Survey.
\end{IEEEkeywords}}

\maketitle

\IEEEdisplaynontitleabstractindextext

%
\IEEEpeerreviewmaketitle


\input{01-introduction}
\input{02-visual_encoding}
\input{03-language_models}
\input{04-training_strategies}
\input{05-evaluation_protocol}
\input{06-results}
\input{07-variants}
\input{08-open_issues}

\ifCLASSOPTIONcompsoc
  \section*{Acknowledgments}
\else
  \section*{Acknowledgment}
\fi
We thank CINECA for providing computational resources. This work has been supported by ``Fondazione di Modena'', by the ``Artificial Intelligence for Cultural Heritage (AI4CH)'' project, co-funded by the Italian Ministry of Foreign Affairs and International Cooperation, and by the H2020 ICT-48-2020 HumanE-AI-NET project. We also want to thank the authors who provided us with the captions and model weights for some of the surveyed approaches.

\ifCLASSOPTIONcaptionsoff
  \newpage
\fi

\bibliographystyle{IEEEtran}
\bibliography{bibliography}

\input{biographies}

\clearpage
\appendices
\input{supplementary_01-metrics}
\input{supplementary_02-evaluation}

\end{document}

%% file: 01-introduction.tex
\IEEEraisesectionheading{\section{Introduction}\label{sec:introduction}}
\IEEEPARstart{I}{mage} captioning is the task of describing the visual content of an image in natural language, employing a visual understanding system and a language model capable of generating meaningful and syntactically correct sentences. Neuroscience research has clarified the link between human vision and language generation only in the last few years~\cite{ardila2015language}. Similarly, in Artificial Intelligence, the design of architectures capable of processing images and generating language is a very recent matter. The goal of these research efforts is to find the most effective pipeline to process an input image, represent its content, and transform that into a sequence of words by generating connections between visual and textual elements while maintaining the fluency of language.

The early-proposed approaches to image captioning have entailed description retrieval~\cite{pan2004automatic,farhadi2010every,ordonez2011im2text,frome2013devise,kiros2014unifying,karpathy2014deep} or template filling and hand-crafted natural language generation techniques~\cite{yao2010i2t,aker2010generating,yang2011corpus,li2011composing,gupta2012choosing,mitchell2012midge,kulkarni2013babytalk,kuznetsova2014treetalk}.
While these have been treated in other surveys~\cite{bernardi2016automatic,bai2018survey,hossain2019comprehensive}, image captioning is currently based on the usage of deep learning-based generative models. In its standard configuration, the task is an image-to-sequence problem whose inputs are pixels. These inputs are encoded as one or multiple feature vectors in the visual encoding step, which prepares the input for a second generative step, called the language model. This produces a sequence of words or sub-words decoded according to a given vocabulary.

In these few years, the research community has improved model design considerably: from the first deep learning-based proposals adopting Recurrent Neural Networks (RNNs) fed with global image descriptors, methods have been enriched with attentive approaches and reinforcement learning up to the breakthroughs of Transformers and self-attention and single-stream BERT-like approaches. At the same time, the Computer Vision and Natural Language Processing (NLP) communities have addressed the challenge of building proper evaluation protocols and metrics to compare results with human-generated ground-truths. However, despite the investigation and improvements achieved in these years, image captioning is still far from being considered a solved task.

Several domain-specific proposals and variants of the task have also been investigated to accommodate for different user needs and descriptions styles. According to~\cite{hodosh2013framing,sharif2020vision}, indeed, image captions can be perceptual, when focusing on low-level visual attributes; non-visual, when reporting implicit and contextual information; conceptual, when describing the actual visual content (\eg~visual entities and their relations).
While the latter is commonly recognized as the target of the image captioning task, this definition encompasses descriptions focusing on different aspects and at various levels of detail (\eg~including attributes or not, mentioning named entities or high-level concepts only, describing salient parts only, or also finer details).

With the aim of providing a testament to the journey that captioning has taken so far, and with that of encouraging novel ideas, we trace a holistic overview of techniques, models, and task variants developed in the last years. Furthermore, we review datasets and evaluation metrics and perform quantitative comparisons of the main approaches. Finally, we discuss open challenges and future directions.

\smallskip
\noindent\tinytit{Contributions} To sum up, the contributions of this survey are as follows:
\begin{itemize}[noitemsep,topsep=0pt,leftmargin=*]
    \item Following the inherent dual nature of captioning models, we develop taxonomies for visual encoding and language modeling approaches and describe their key aspects and limitations. 
    \item We review the training strategies adopted in the literature over the past years and the recent advancement obtained by the pre-training paradigm and masked language model losses.
    \item We review the main datasets used to explore image captioning, both domain-generic benchmarks and domain-specific datasets collected to investigate specific aspects. 
    \item We analyze both standard and non-standard metrics adopted for performance evaluation and the characteristics of the caption they highlight.
    \item We present a quantitative comparison of the main image captioning methods considering both standard and non-standard metrics and a discussion on their relationships, which sheds light on performance, differences, and characteristics of the most important models.
    \item We give an overview of many variants of the task and discuss open challenges and future directions.
\end{itemize}

\smallskip
Compared to previous surveys on image captioning~\cite{hossain2019comprehensive,bai2018survey,liu2019survey,sharma2020image,bernardi2016automatic}, we provide a comprehensive and updated view on deep learning-based generative captioning models. We perform a deeper analysis of proposed approaches and survey a considerably larger number of papers on the topic. Also, we cover non-standard evaluation metrics, which are disregarded by other works, discuss their characteristics, and employ them in a quantitative evaluation of state-of-the-art methods. Moreover, we tackle emerging variants of the task and a broader set of available datasets.

\begin{figure}[t]
\centering
\includegraphics[width=0.98\linewidth]{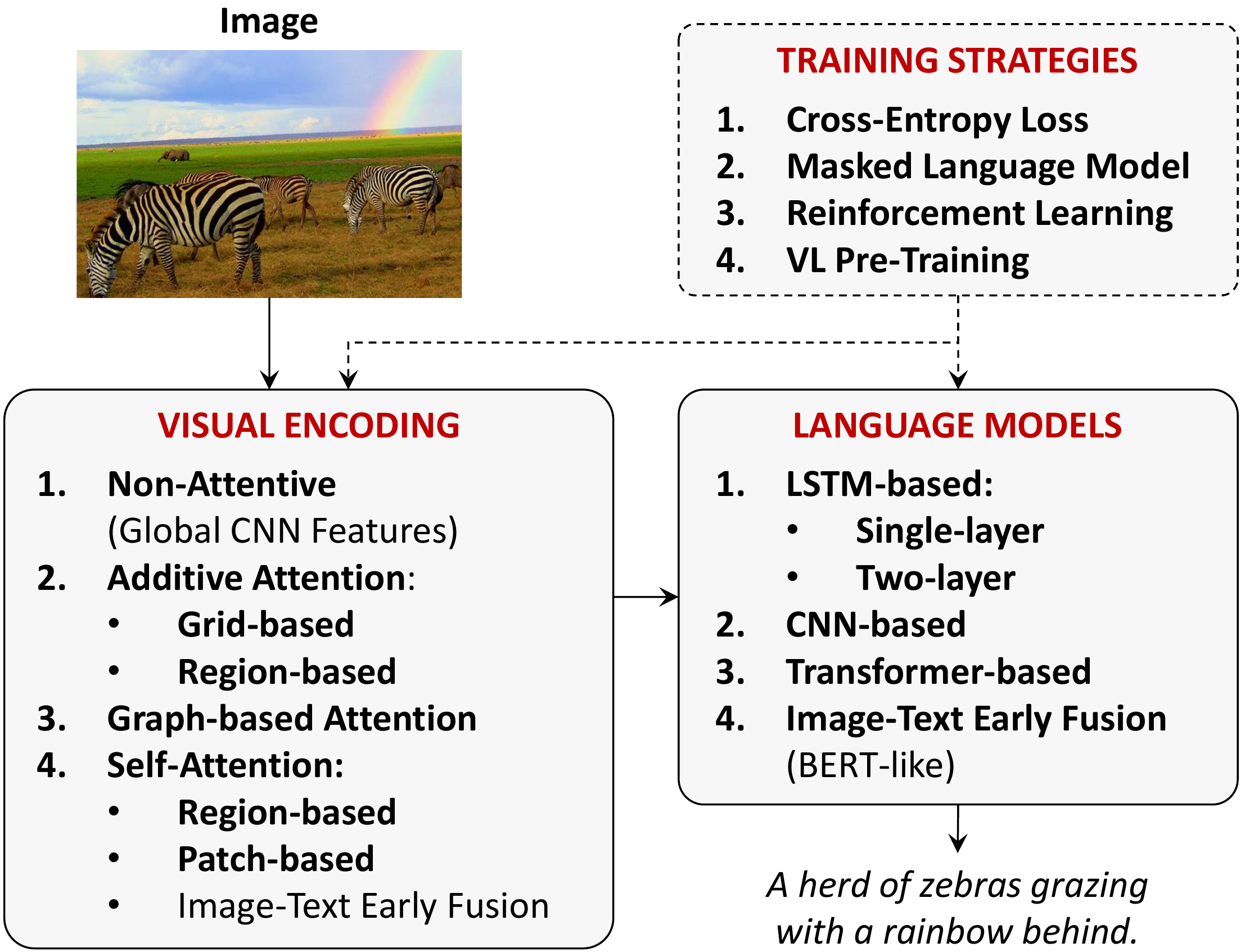}
\caption{Overview of the image captioning task and taxonomy of the most relevant approaches.}
\label{fig:first_page}
\vspace{-0.25cm}
\end{figure}

%% file: 02-visual_encoding.tex
\section{Visual Encoding}\label{sec:visual_encoding}
Providing an effective representation of the visual content is the first challenge of an image captioning pipeline. 
The current approaches for visual encoding can be classified as belonging to four main categories: 1.~\emph{non-attentive methods} based on global CNN features; 2.~\emph{additive attentive methods} that embed the visual content using either grids or regions; 3.~\emph{graph-based methods} adding visual relationships between visual regions; and 4.~\emph{self-attentive methods} that employ Transformer-based paradigms, either by using region-based, patch-based, or image-text early fusion solutions. This taxonomy is visually summarized in Fig.~\ref{fig:first_page}.

\begin{figure*}[tb]
\centering
\resizebox{\linewidth}{!}{
\subfloat[]{
\includegraphics[height=0.1478\textwidth]{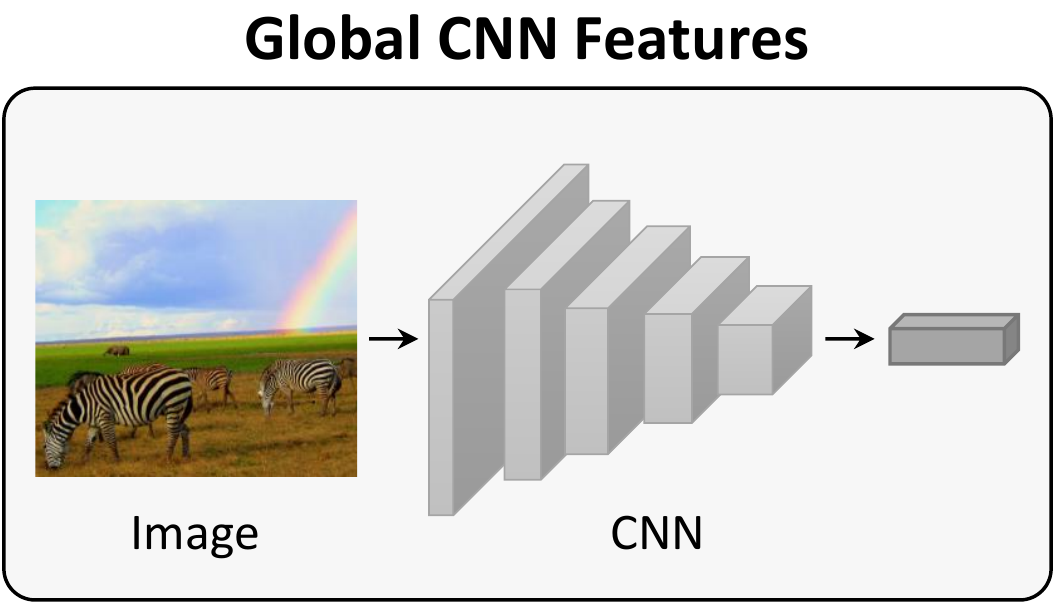}
\label{sfig:global}}
\subfloat[]{
\includegraphics[height=0.1478\textwidth]{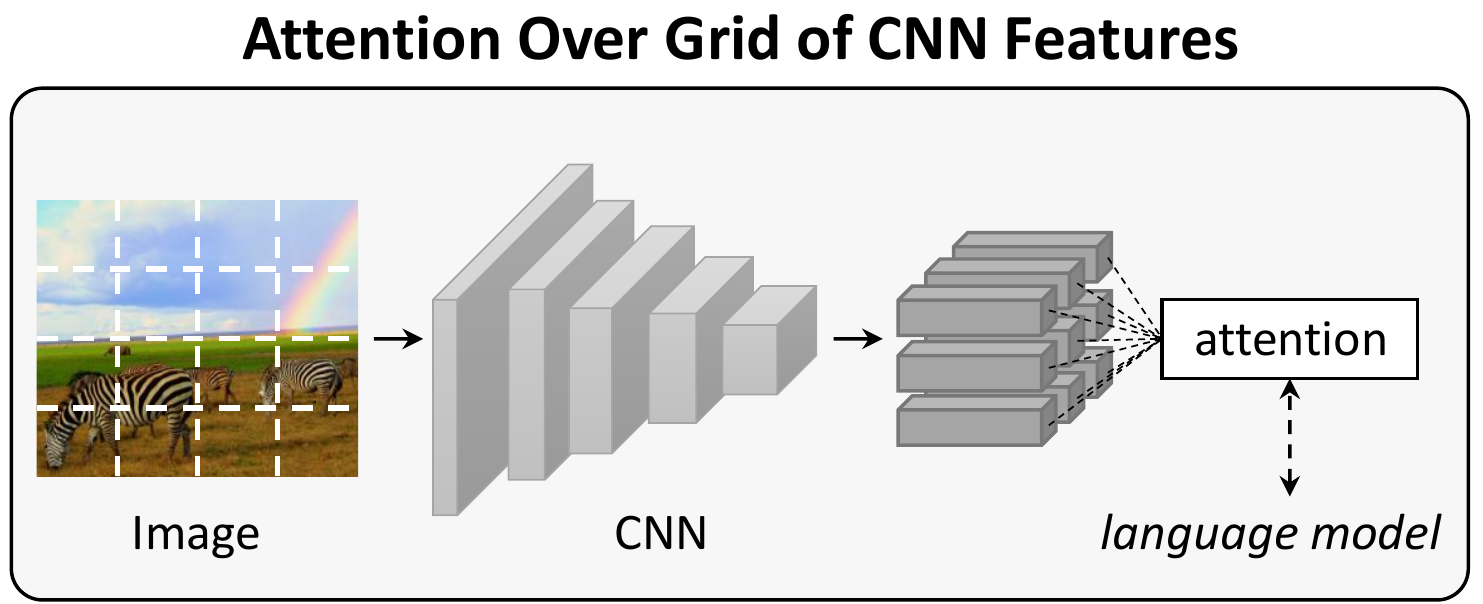}
\label{sfig:grid}}
\subfloat[]{
\includegraphics[height=0.1478\textwidth]{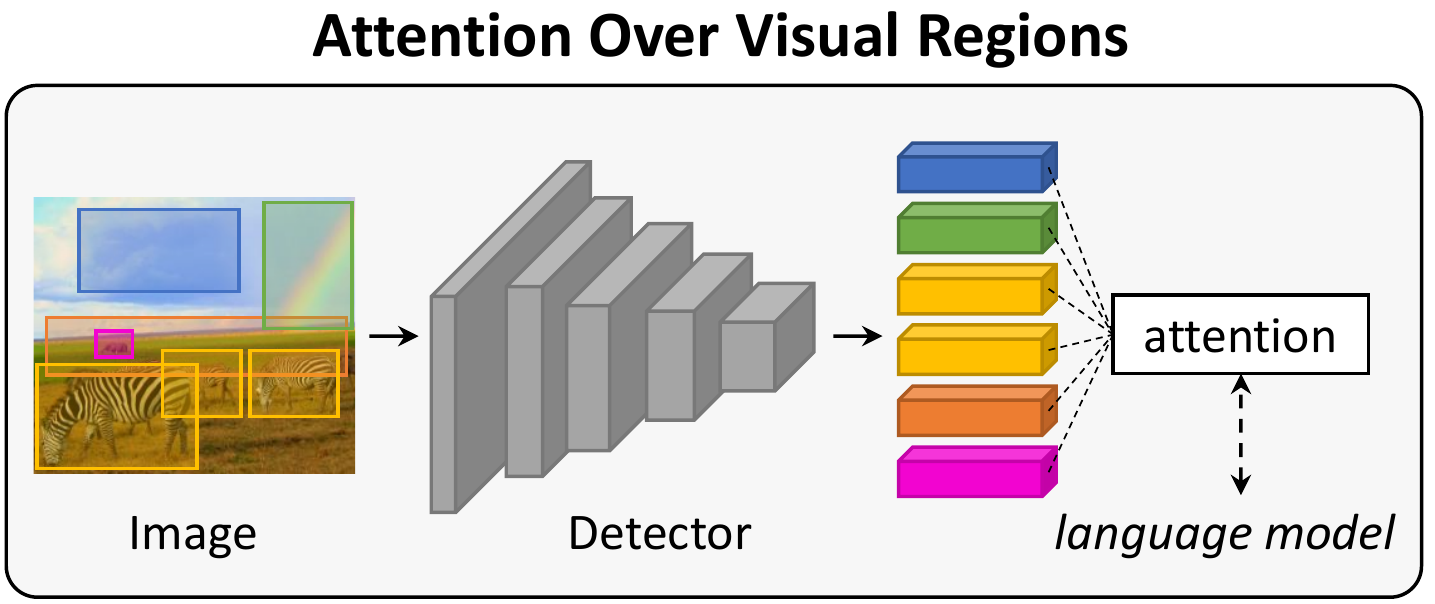}
\label{sfig:regions}}
}
\caption{Three of the most relevant visual encoding strategies for image captioning: \textbf{(a)} global CNN features; \textbf{(b)} fine-grained features extracted from the activation of a convolutional layer, together with an attention mechanism guided by the language model; \textbf{(c)} image region features coming from a detector, together with an attention mechanism.
}
\label{fig:encodings}
\vspace{-0.25cm}
\end{figure*}

\subsection{Global CNN Features}
With the advent of CNNs, all models consuming visual inputs have been improved in terms of performance. The visual encoding step of image captioning is no exception. In the most simple recipe, the activation of one of the last layers of a CNN is employed to extract high-level representations, which are then used as a conditioning element for the language model (Fig.~\ref{sfig:global}). This is the approach employed in the seminal ``Show and Tell'' paper~\cite{vinyals2015show}\footnote{The title of this survey is a tribute of this pioneering work.}, where the output of GoogleNet~\cite{szegedy2015going} is fed to the initial hidden state of the language model. In the same year, Karpathy~\etal~\cite{karpathy2015deep} used global features extracted from AlexNet~\cite{krizhevsky2012imagenet} as the input for a language model. Further, Mao~\etal~\cite{mao2015deep} and Donahue~\etal~\cite{donahue2015long} injected global features extracted from the VGG network~\cite{simonyan2014very} at each time-step of the language model.

Global CNN features were then employed in a large variety of image captioning models~\cite{chen2015mind,fang2015captions,jia2015guiding,you2016image,wu2016value,gu2017empirical,chen2017structcap,chen2018groupcap}. Notably, Rennie~\etal~\cite{rennie2017self} introduced the FC model, in which images are encoded using a ResNet-101~\cite{he2016deep}, preserving their original dimensions. Other approaches~\cite{yao2017boosting,gan2017semantic} integrated high-level attributes or tags, represented as a probability distribution over the most common words of the training captions.

The main advantage of employing global CNN features resides in their simplicity and compactness of representation, which embraces the capacity to extract and condense information from the whole input and to consider the overall context of an image. However, this paradigm also leads to excessive compression of information and lacks granularity, making it hard for a captioning model to produce specific and fine-grained descriptions.

\subsection{Attention Over Grid of CNN Features}
Motivated by the drawbacks of global representations, most of the following approaches have increased the granularity level of visual encoding~\cite{xu2015show,rennie2017self,lu2017knowing}. For instance, Dai~\etal\cite{dai2018rethinking} have employed 2D activation maps in place of 1D global feature vectors to bring spatial structure directly in the language model. 
Drawing from machine translation literature, a big portion of the captioning community has instead employed the additive attention mechanism  (Fig.~\ref{sfig:grid}), which has endowed image captioning architectures with time-varying visual features encoding, enabling greater flexibility and finer granularity.

\tit{Definition of additive attention}
The intuition behind attention boils down to weighted averaging. In the first formulation proposed for sequence alignment by Bahdanau~\etal~\cite{bahdanau2014neural} (also known as \textit{additive attention}), a single-layer feed-forward neural network with a hyperbolic tangent non-linearity is used to compute attention weights. Formally, given two generic sets of vectors $\{\mathbf{x}_{1}, \ldots, \mathbf{x}_{n}\}$ and $\{\mathbf{h}_{1}, \ldots, \mathbf{h}_{m}\}$, the additive attention score between $\mathbf{h}_{i}$ and $\mathbf{x}_{j}$ is computed as follows:
\begin{equation}
f_{\mathrm{att}}\left(\mathbf{h}_{i}, \mathbf{x}_{j}\right)=\mathbf{W}_{3}^{\top} \tanh \left(\mathbf{W}_{1} \mathbf{h}_{i}+\mathbf{W}_{2} \mathbf{x}_{j}\right),
\end{equation}
where $\mathbf{W}_{1}$ and $\mathbf{W}_{2}$ are weight matrices, and $\mathbf{W}_{3}$ is a weight vector that performs a linear combination. A softmax function is then applied to obtain a probability distribution $p\left(\mathbf{x}_{j} \mid \mathbf{h}_{i}\right)$, representing how much the element encoded by $\mathbf{x}_{j}$ is relevant for $\mathbf{h}_{i}$. 

Although the attention mechanism was initially devised for modeling the relationships between two sequences of elements (\ie~hidden states from a recurrent encoder and a decoder), it can be adapted to connect a set of visual representations with the hidden states of a language model.

\tit{Attending convolutional activations}
Xu~\etal~\cite{xu2015show} introduced the first method leveraging the additive attention over the spatial output grid of a convolutional layer. This allows the model to selectively focus on certain elements of the grid by selecting a subset of features for each generated word. Specifically, the model first extracts the activation of the last convolutional layer of a VGG network~\cite{simonyan2014very}, then uses additive attention to compute a weight for each grid element, interpreted as the relative importance of that element for generating the next word.

\tit{Other approaches}
The solution based on additive attention over a grid of features has been widely adopted by several following works with minor improvements in terms of visual encoding~\cite{yao2017boosting,chen2018regularizing,lu2017knowing,wang2017skeleton,ge2019exploring,gu2018stack}.

\tyt{Review networks}
For instance, Yang~\etal~\cite{yang2016review} supplemented the encoder-decoder framework with a recurrent review network. This performs a given number of review steps with attention on the encoder hidden states and outputs a ``thought vector'' after each step, which is then used by the attention mechanism in the decoder. 

\tyt{Multi-level features}
Chen~\etal~\cite{chen2017sca} proposed to employ channel-wise attention over convolutional activations, followed by a more classical spatial attention. They also experimented with using more than one convolutional layer to exploit multi-level features. 
On the same line, Jiang~\etal~\cite{jiang2018recurrent} proposed to use multiple CNNs in order to exploit their complementary information, then fused their representations with a recurrent procedure. 

\tyt{Exploiting human attention}
Some works also integrated saliency information (\ie~what do humans pay more attention to in a scene) to guide caption generation with stimulus-based attention. This idea was first explored by Sugano and Bulling~\cite{sugano2016seeing} who exploited human eye fixations for image captioning by including normalized fixation histograms over the image as an input to the soft-attention module of~\cite{xu2015show} and weighing the attended image regions based on whether these are fixated or not. Subsequent works on this line~\cite{tavakoli2017paying,ramanishka2017top,cornia2018paying,chen2018boosted} employed saliency maps as a form of additional attention source.

\begin{figure*}[tb]
\centering
\subfloat[]{
\includegraphics[height=0.155\textwidth]{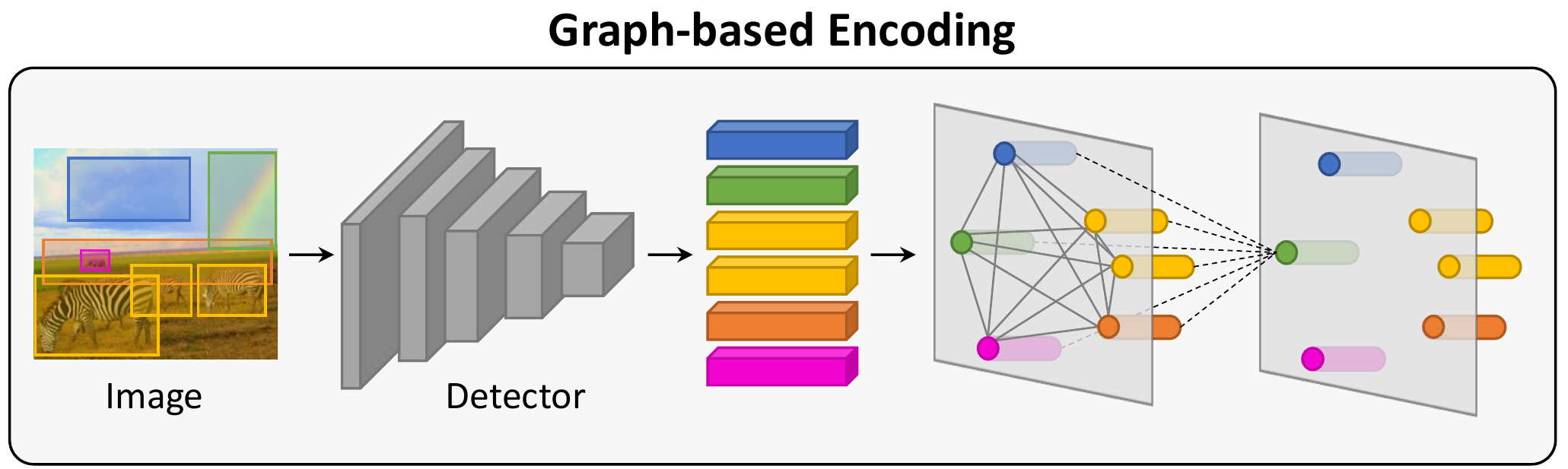}
\label{sfig:graph}}
\hspace{0.3cm}
\subfloat[]{
\includegraphics[height=0.155\textwidth]{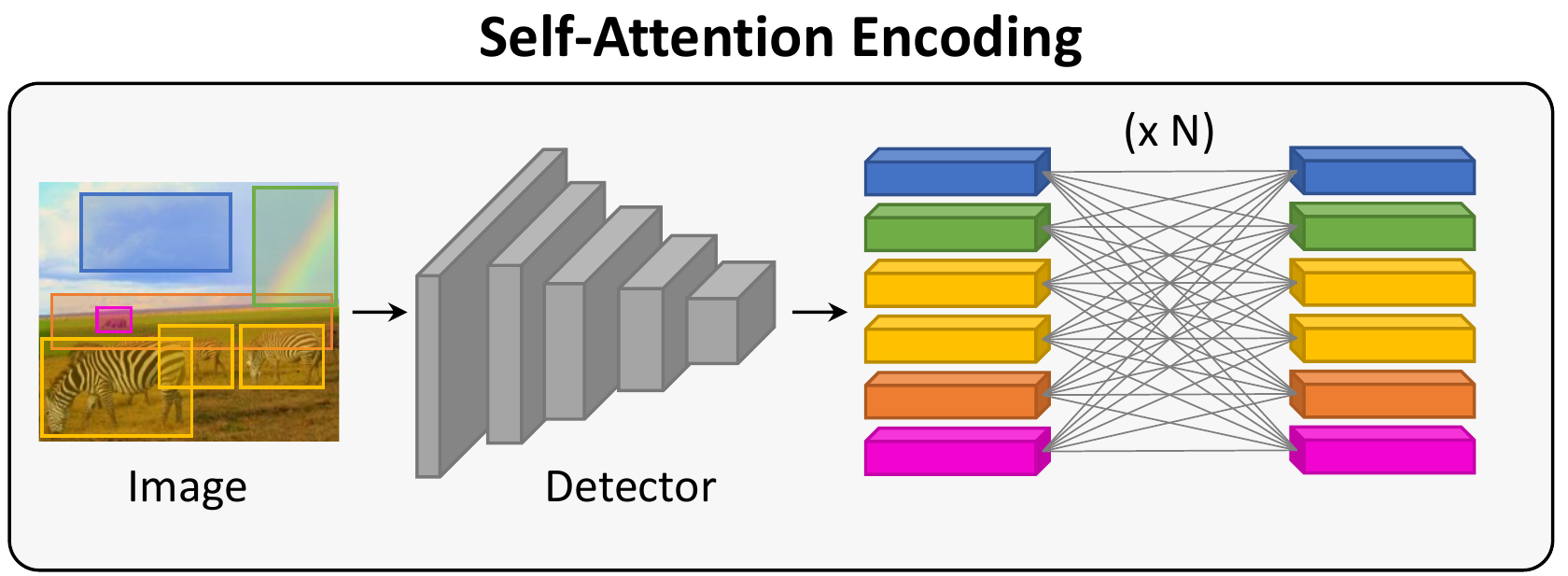}
\label{sfig:self_att}}
\caption{Summary of the two most recent visual encoding strategies for image captioning: \textbf{(a)} graph-based encoding of visual regions; \textbf{(b)} self-attention-based encoding over image region features.}
\label{fig:encodings_2}
\vspace{-0.25cm}
\end{figure*}

\subsection{Attention Over Visual Regions}
The intuition of using saliency boils down to neuroscience, which suggests that our brain integrates a top-down reasoning process with a bottom-up flow of visual signals. The top-down path consists of predicting the upcoming sensory input by leveraging our knowledge and inductive bias, while the bottom-up flow provides visual stimuli adjusting the previous predictions. 
Additive attention can be thought of as a top-down system. In this mechanism, the language model predicts the next word while attending a feature grid, whose geometry is irrespective of the image content.

\tit{Bottom-up and top-down attention}
Differently from saliency-based approaches~\cite{chen2018boosted}, in the solution proposed by Anderson~\etal~\cite{anderson2018bottom} the bottom-up path is defined by an object detector in charge of proposing image regions. This is then coupled with a top-down mechanism that learns to weigh each region for each word prediction (see Fig.~\ref{sfig:regions}). In this approach, Faster R-CNN~\cite{ren2015faster,ren2017faster} is adopted to detect objects, obtaining a pooled feature vector for each region proposal. One of the key elements of this approach resides in its pre-training strategy, where an auxiliary training loss is added for learning to predict attribute classes alongside object classes on the Visual Genome~\cite{krishnavisualgenome} dataset. This allows the model to predict a dense and rich set of detections, including both salient object and contextual regions, and favors the learning of better feature representations.

\tit{Other approaches}
Employing image region features has demonstrated its advantages when dealing with the raw visual input and has been the standard de-facto in image captioning for years. As a result, many of the following works have based the visual encoding phase on this strategy~\cite{ke2019reflective,qin2019look,huang2019adaptively,wang2020show}. Among them, we point out two remarkable variants. 

\tyt{Visual Policy}
While typical visual attention points to a single image region at every step, the approach proposed by Zha~\etal~\cite{zha2019context} introduces a sub-policy network that interprets also the visual part sequentially by encoding historical visual actions (\eg~previously attended regions) via an LSTM to serve as context for the next visual action. 

\tyt{Geometric Transforms} Pedersoli~\etal~\cite{pedersoli2017areas} proposed to use spatial transformers for generating image-specific attention areas by regressing region proposals in a weakly-supervised fashion. Specifically, a localization network learns an affine transformation or each location of the feature map, and then a bilinear interpolation is used to regress a feature vector for each region with respect to anchor boxes.

\subsection{Graph-based Encoding}
To further improve the encoding of image regions and their relationships, some studies consider using graphs built over image regions (see Fig.~\ref{sfig:graph}) to enrich the representation by including semantic and spatial connections. 

\tit{Spatial and semantic graphs} 
The first attempt in this sense is due to Yao~\etal~\cite{yao2018exploring}, followed by Guo~\etal~\cite{guo2019aligning}, who proposed the use of a graph convolutional network (GCN)~\cite{kipf2016semi} to integrate both semantic and spatial relationships between objects. The semantic relationships graph is obtained by applying a classifier pre-trained on Visual Genome~\cite{krishnavisualgenome} that predicts an action or an interaction between object pairs. The spatial relationships graph is instead inferred through geometry measures (\ie~intersection over union, relative distance, and angle) between bounding boxes of object pairs.

\tit{Scene graphs}
With a focus on modeling semantic relations, Yang~\etal~\cite{yang2019auto} proposed to integrate semantic priors learned from text in the image encoding by exploiting a graph-based representation of both images and sentences. The representation used is the scene graph, \ie~a directed graph connecting the objects, their attributes, and their relations. On the same line, Shi~\etal~\cite{shi2020improving} represented the image as a semantic relationship graph but proposed to train the module in charge of predicting the predicate nodes directly on the ground-truth captions rather than on external datasets.

\tit{Hierarchical trees} 
As a special case of a graph-based encoding, Yao~\etal~\cite{yao2019hierarchy} employed a tree to represent the image as a hierarchical structure. The root represents the image as a whole, intermediate nodes represent image regions and their contained sub-regions, and the leaves represent segmented objects in the regions.

Graph encodings brought a mechanism to leverage relationships between detected objects, which allows the exchange of information in adjacent nodes and thus in a local manner. Further, it seamlessly allows the integration of external semantic information. On the other hand, manually building the graph structure can limit the interactions between visual features. This is where self-attention proved to be more successful by connecting all the elements with each other in a complete graph representation.

\subsection{Self-Attention Encoding}
Self-attention is an attentive mechanism where each element of a set is connected with all the others, and that can be adopted to compute a refined representation of the same set of elements through residual connections (Fig.~\ref{sfig:self_att}). It was first introduced by Vaswani~\etal~\cite{vaswani2017attention} for machine translation and language understanding tasks, giving birth to the Transformer architecture and its variants, which have dominated the NLP field and later also Computer Vision.

\tit{Definition of self-attention}
Formally, self-attention makes use of the scaled dot-product mechanism, \ie~a multiplicative attention operator that handles three sets of vectors: a set of $n_q$ query vectors $\bm{Q}$, a set of key vectors $\bm{K}$, and a set of value vectors $\bm{V}$, both containing $n_k$ elements. The operator takes a weighted sum of value vectors according to a similarity distribution between query and key vectors:
\begin{align}
\mathsf{Attention}(\bm{Q}, \bm{K}, \bm{V})=\operatorname{softmax}\left(\frac{\bm{Q} \bm{K}^{T}}{\sqrt{d_{k}}}\right) \bm{V},
\label{eq:attention}
\end{align}
where $d_k$ is a scaling factor. 
In the case of self-attention, the three sets of vectors are obtained as linear projections of the same input set of elements. 
The success of the Transformer demonstrates that leveraging self-attention allows achieving superior performances compared to attentive RNNs. 

\tit{Early self-attention approaches}
Among the first image captioning models leveraging this approach, Yang~\etal~\cite{yang2019learning} used a self-attentive module to encode relationships between features coming from an object detector. Later, Li~\etal~\cite{li2019entangled} proposed a Transformer model with a visual encoder for the region features coupled with a semantic encoder that exploits knowledge from an external tagger. Both encoders are based on self-attention and feed-forward layers. Their output is then fused through a gating mechanism governing the propagation of visual and semantic information.

\tit{Variants of the self-attention operator}
Other works proposed variants or modifications of the self-attention operator tailored for image captioning~\cite{herdade2019image,guo2020normalized,huang2019attention,pan2020x,cornia2020meshed}.

\tyt{Geometry-aware encoding}
Herdade~\etal~\cite{herdade2019image} introduced a modified version of self-attention that takes into account the spatial relationships between regions. In particular, an additional geometric weight is computed between object pairs and is used to scale the attention weights. On a similar line, Guo~\etal~\cite{guo2020normalized} proposed a normalized and geometry-aware version of self-attention that makes use of the relative geometry relationships between input objects.
Further, He~\etal~\cite{he2020image} introduced a spatial graph transformer, which considers different categories of spatial relationship between detections (\eg,~parent, neighbor, child) when performing attention.

\tyt{Attention on Attention} Huang~\etal~\cite{huang2019attention} proposed an extension of the attention operator in which the final attended information is weighted by a gate guided by the context. Specifically, the output of the self-attention is concatenated with the queries, then an information and a gate vector are computed and finally multiplied together. In their encoder, they employed this mechanism to refine the visual features. 
This method is then adopted by later models such as~\cite{liu2020prophet}.

\tyt{X-Linear Attention} Pan~\etal~\cite{pan2020x} proposed to use bilinear pooling techniques to strengthen the representative capacity of the output attended feature. Notably, this mechanism encodes the region-level features with higher-order interaction, leading to a set of enhanced region-level and image-level features.

\tyt{Memory-augmented Attention} Cornia~\etal~\cite{cornia2020meshed,cornia2020smart} proposed a Transformer-based architecture where the self-attention operator of each encoder layer is augmented with a set of memory vectors. Specifically, the set of keys and values is extended with additional ``slots'' learned during training, which can encode multi-level visual relationships. 

\tit{Other self-attention-based approaches}
Ji~\etal~\cite{Ji2020ImprovingIC} proposed to improve self-attention by adding to the sequence of feature vectors a global vector computed as their average. A global vector is computed for each layer, and the resulting global vectors are combined via an LSTM, thus obtaining an inter-layer representation. Luo~\etal~\cite{luo2021dual} proposed a hybrid approach that combines region and grid features to exploit their complementary advantages. Two self-attention modules are applied independently to each kind of features, and a cross-attention module locally fuses their interactions.
On a different line, the architecture proposed by Liu~\etal~\cite{liu2019aligning} is based on an attention module to align grid or detection features with visual words extracted from a concept extractor and to obtain semantic-grounded encodings.

\begin{figure}[tb]
\centering
\includegraphics[width=0.98\linewidth]{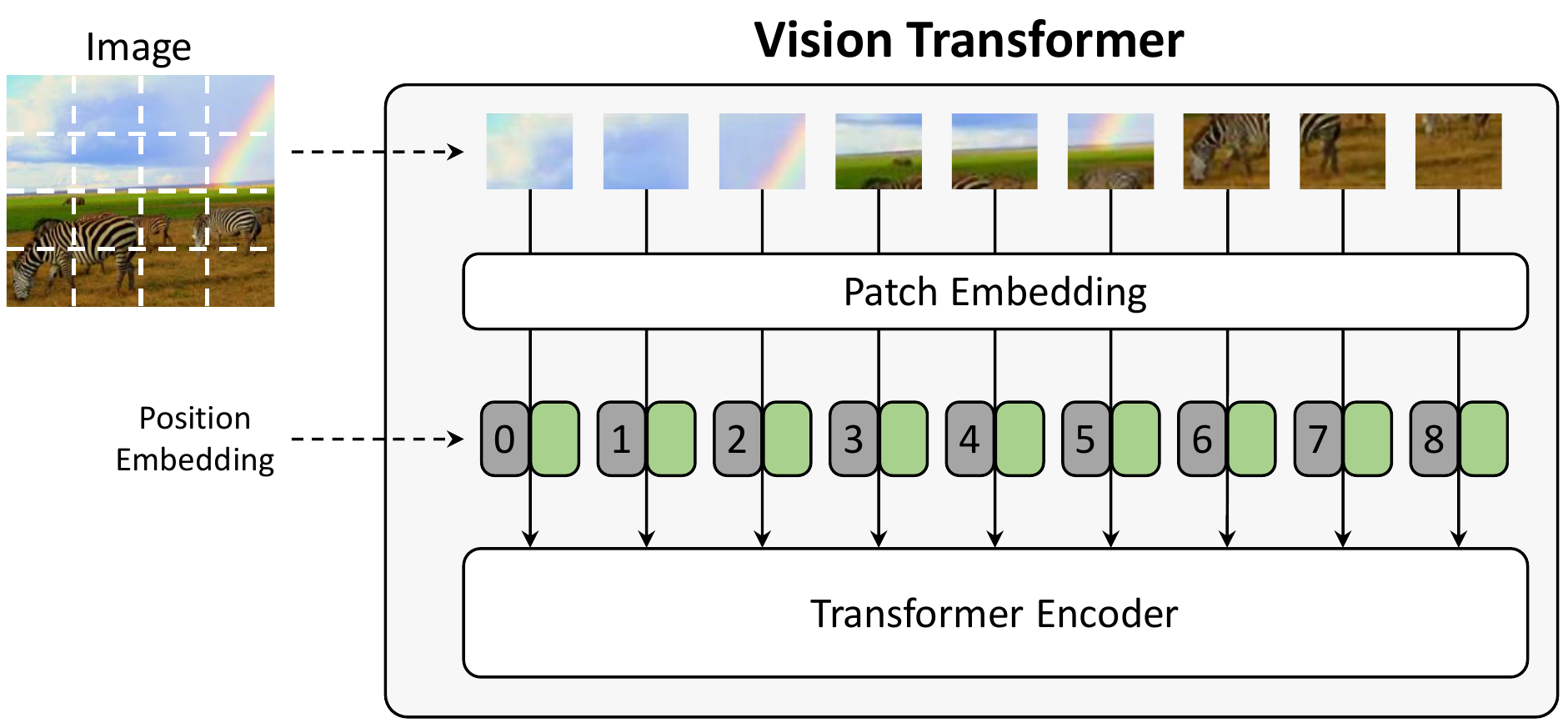}
\caption{Vision Transformer encoding. The image is split into fixed-size patches, linearly embedded, added to position embeddings, and fed to a standard Transformer encoder.}
\label{fig:vision_transformer}
\vspace{-0.25cm}
\end{figure}

\tit{Attention on grid features and patches}
Other than applying the attention operator on detections, the role of grid features has been recently re-evaluated~\cite{jiang2020defense}. For instance, the approach proposed by Zhang~\etal~\cite{zhang2021rstnet} applies self-attention directly to grid features, incorporating their relative geometry relationships into self-attention computation.
Transformer-like architectures can also be applied directly on image patches, thus excluding the usage of the convolutional operator~\cite{dosovitskiy2020image,touvron2020training} (Fig.~\ref{fig:vision_transformer}). On this line, Liu~\etal~\cite{liu2021cptr} devised the first convolution-free architecture for image captioning. Specifically, a pre-trained Vision Transformer network (\ie~ViT~\cite{dosovitskiy2020image}) is adopted as encoder, and a standard Transformer decoder is employed to generate captions. Interestingly, the same visual encoding approach has been adopted in CLIP~\cite{radford2021learning} and SimVLM~\cite{wang2021simvlm}, with the difference that the visual encoder is trained from scratch on large-scale noisy data. CLIP-based features have then been used by subsequent captioning approaches~\cite{shen2021much,mokady2021clipcap,cornia2021universal}.

\tit{Early fusion and vision-and-language pre-training}
Other works using self-attention to encode visual features achieved remarkable performance also thanks to vision-and-language pre-training~\cite{tan2019lxmert,lu2019vilbert} and early-fusion strategies~\cite{li2020oscar,zhou2020unified}. For example, following the BERT architecture~\cite{devlin2018bert}, Zhou~\etal~\cite{zhou2020unified} combined encoder and decoder into a single stream of Transformer layers, where region and word tokens are early fused together into a unique flow. This unified model is first pre-trained on large amounts of image-caption pairs to perform both bidirectional and sequence-to-sequence prediction tasks and then fine-tuned. 

On the same line, Li~\etal~\cite{li2020oscar} proposed OSCAR, a BERT-like architecture that includes object tags as anchor points to ease the semantic alignment between images and text. 
They also performed a large-scale pre-train with $6.5$ million image-text pairs, with a masked token loss similar to the BERT mask language loss and a contrastive loss for distinguishing aligned words-tags-regions triples from polluted ones. 
Later, Zhang~\etal~\cite{zhang2021vinvl} proposed VinVL, built on top of OSCAR, introducing a new object detector capable of extracting better visual features and a modified version of the vision-and-language pre-training objectives. On this line, Hu~\etal~\cite{hu2021scaling} improved the VinVL model by scaling up its size and using larger scale noisy data to pre-train.

\begin{figure*}[tb]
\centering
\subfloat[]{
\includegraphics[height=0.21\textwidth]{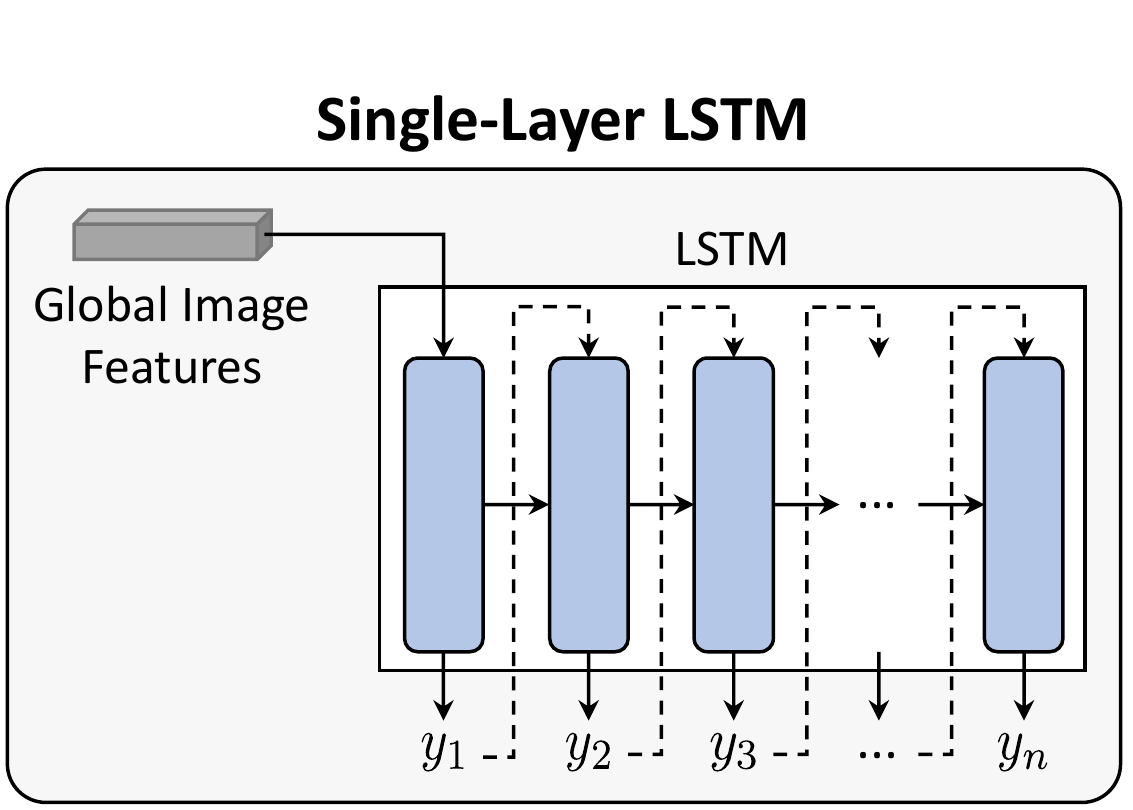}
\label{sfig:lstm1}}
\hspace{0.2cm}
\subfloat[]{
\includegraphics[height=0.21\textwidth]{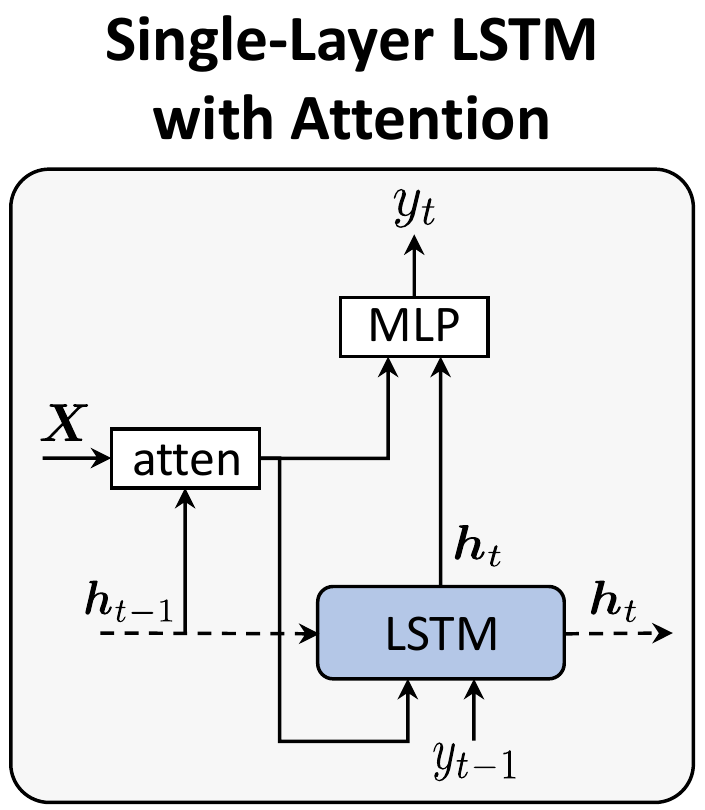}
\label{sfig:lstm2}}
\hspace{0.2cm}
\subfloat[]{
\includegraphics[height=0.21\textwidth]{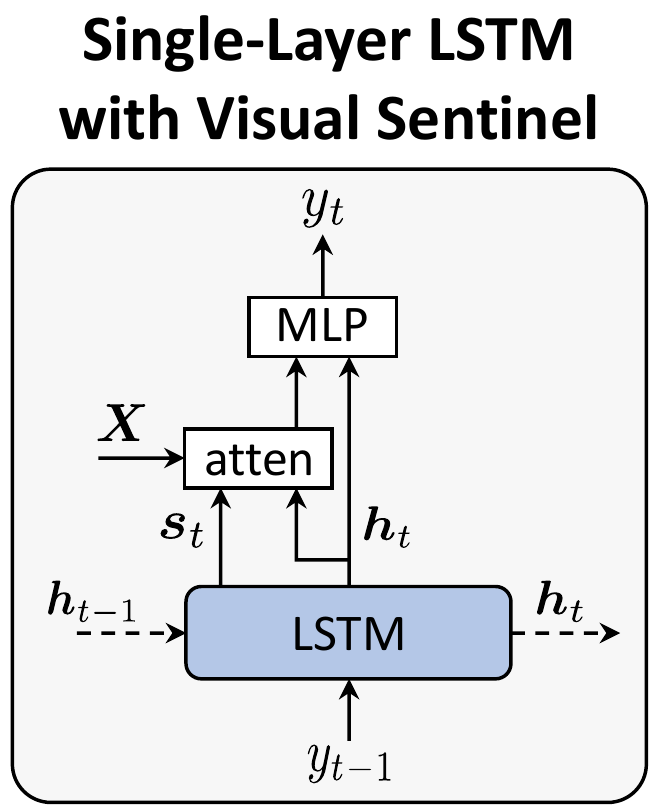}
\label{sfig:lstm3}}
\hspace{0.2cm}
\subfloat[]{
\includegraphics[height=0.21\textwidth]{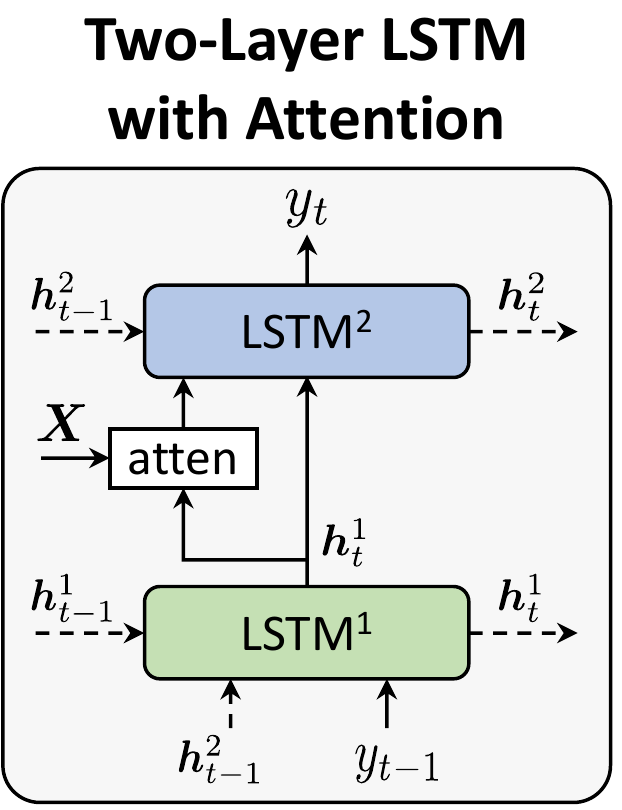}
\label{sfig:lstm4}}
\caption{LSTM-based language modeling strategies: \textbf{(a)} Single-Layer LSTM model conditioned on the visual feature; \textbf{(b)} LSTM with attention, as proposed in the Show, Attend and Tell model~\cite{xu2015show}; \textbf{(c)} LSTM with attention, in the variant proposed in~\cite{lu2017knowing}; \textbf{(d)} two-layer LSTM with attention, in the style of the bottom-up top-down approach by Anderson~\etal~\cite{anderson2018bottom}. In all figures, $\bm{X}$ represents either a grid of CNN features or image region features extracted by an object detector. }
\label{fig:lstm}
\vspace{-0.25cm}
\end{figure*}

\subsection{Discussion}
After the emergence of global features and grid features, region-based features have been the state-of-the-art choice in image captioning for years thanks to their compelling performances. Recently, however, different factors are reopening the discussion on which feature model is most appropriate for image captioning, ranging from the performance of better-trained grid features~\cite{jiang2020defense} to the emergence of self-attentive visual encoders~\cite{dosovitskiy2020image} and large-scale multi-modal models like CLIP~\cite{radford2021learning}. Recent strategies encompass training better object detectors on large-scale data~\cite{zhang2021vinvl} or employing end-to-end visual models trained from scratch~\cite{wang2021simvlm}. Moreover, the success of BERT-like solutions performing image and text early-fusion indicates the suitability of visual representations that also integrate textual information.

%% file: 03-language_models.tex
\section{Language Models}\label{sec:language_model}
The goal of a language model is to predict the probability of a given sequence of words to occur in a sentence. As such, it is a crucial component in image captioning, as it gives the ability to deal with natural language as a stochastic process.

Formally, given a sequence of $n$ words, the language model component of an image captioning algorithm assigns a probability $P\left(y_{1}, y_{2}, \ldots, y_{n} \mid \bm{X} \right)$ to the sequence as:
\begin{equation}
P\left(y_{1}, y_{2}, \ldots y_{n} \mid \bm{X}\right)=\prod_{i=1}^{n} P\left(y_{i} \mid y_{1}, y_{2}, \ldots, y_{i-1}, \bm{X}\right),
\end{equation}
where $\bm{X}$ represents the visual encoding on which the language model is specifically conditioned. Notably, when predicting the next word given the previous ones, the language model is auto-regressive, which means that each predicted word is conditioned on the previous ones. 
The language model usually also decides when to stop generating caption words by outputting a special end-of-sequence token.

The main language modeling strategies applied to image captioning 
can be categorized as: 1.~\emph{LSTM-based} approaches, which can be either single-layer or two-layer; 2.~\emph{CNN-based} methods that constitute a first attempt in surpassing the fully recurrent paradigm; 3.~\emph{Transformer-based} fully-attentive approaches; 4.~\emph{image-text early-fusion} (BERT-like) strategies that directly connect the visual and textual inputs. 
This taxonomy is visually summarized in Fig.~\ref{fig:first_page}.

\subsection{LSTM-based Models}
As language has a sequential structure, RNNs are naturally suited to deal with the generation of sentences. Among RNN variants, LSTM~\cite{hochreiter1997long} has been the predominant option for language modeling.

\subsubsection{Single-layer LSTM}
The most simple LSTM-based captioning architecture is based on a single-layer LSTM and was proposed by Vinyals~\etal~\cite{vinyals2015show}. As shown in Fig.~\ref{sfig:lstm1}, the visual encoding is used as the initial hidden state of the LSTM, which then generates the output caption. At each time step, a word is predicted by applying a softmax activation function over the projection of the hidden state into a vector of the same size as the vocabulary. During training, input words are taken from the ground-truth sentence, while during inference, input words are those generated at the previous step.

Shortly after, Xu~\etal~\cite{xu2015show} introduced the additive attention mechanism. As depicted in Fig.~\ref{sfig:lstm2}, in this case, the previous hidden state guides the attention mechanism over the visual features $\bm{X}$, computing a context vector which is then fed to the MLP in charge of predicting the output word. 

\tit{Other approaches} 
Many subsequent works have adopted a decoder based on a single-layer LSTM, mostly without any architectural changes~\cite{yang2016review,chen2017sca,pedersoli2017areas}, while others have proposed significant modifications, summarized below.

\tyt{Visual sentinel}
Lu~\etal~\cite{lu2017knowing} augmented the spatial image features with an additional learnable vector, called visual sentinel, which can be attended by the decoder in place of visual features while generating ``non-visual'' words (\eg~``the'', ``of'', and ``on''), for which visual features are not needed (Fig.~\ref{sfig:lstm3}). At each time step, the visual sentinel is computed from the previous hidden state and generated word. Then, the model generates a context vector as a combination of attended image features and visual sentinel, whose importance is weighted by a learnable gate.

\tyt{Hidden state reconstruction}
Chen~\etal~\cite{chen2018regularizing} proposed to regularize the transition dynamics of the language model by using a second LSTM for reconstructing the previous hidden state based on the current one. Ge~\etal~\cite{ge2019exploring} enhance context modeling by by using a bidirectional LSTM with an auxiliary module. The auxiliary module in a direction approximates the hidden state of the LSTM in the other direction. Finally, a cross-modal attention mechanism combines grid visual features with the two sentences from the bidirectional LSTM to obtain the final caption.

\tyt{Multi-stage generation}
Wang~\etal~\cite{wang2017skeleton} proposed to generate a caption from coarse central aspects to finer attributes by decomposing the caption generation process into two phases: skeleton sentence generation and attributes enriching, both implemented with single-layer LSTMs. On the same line, Gu~\etal~\cite{gu2018stack} devised a coarse-to-fine multi-stage framework using a sequence of LSTM decoders, each operating on the output of the previous one to produce increasingly refined captions.  

\tyt{Semantic-guided LSTM}
Jia~\etal~\cite{jia2015guiding} proposed an extension of LSTM that includes semantic information extracted from the image to guide the generation. Specifically, the semantic information is used as an extra input to each gate in the LSTM block.

\subsubsection{Two-layer LSTM} 
LSTMs can be expanded to multi-layer structures to augment their capability of capturing higher-order relations. Donahue~\etal~\cite{donahue2015long} firstly proposed a two-layer LSTM as a language model for captioning, stacking two layers, where the hidden states of the first are the input to the second.

\tit{Two-layers and additive attention}
Anderson~\etal~\cite{anderson2018bottom} went further and proposed to specialize the two layers to perform visual attention and the actual language modeling. As shown in Fig.~\ref{sfig:lstm4}, the first LSTM layer acts as a top-down visual attention model which takes the previously generated word, the previous hidden state, and the mean-pooled image features. Then, the current hidden state is used to compute a probability distribution over image regions with an additive attention mechanism. The so-obtained attended image feature vector is fed to the second LSTM layer, which combines it with the hidden state of the first layer to generate a probability distribution over the vocabulary.

\tit{Variants of two-layers LSTM}
Because of their representation power, LSTMs with two-layers and internal attention mechanisms represent the most employed language model approach before the advent of Transformer-based architectures~\cite{yao2018exploring,yang2019auto,yao2019hierarchy,shi2020improving}. As such, many other variants have been proposed to improve the performance of this approach.

\tyt{Neural Baby Talk}
To ground words into image regions, Lu~\etal~\cite{lu2018neural} incorporated a pointing network that modulates the content-based attention mechanism. In particular, during the generation process, the network predicts slots in the caption, which are then filled with the image region classes. For non-visual words, a visual sentinel is used as dummy grounding. This approach leverages the object detector both as a feature region extractor and as a visual word prompter for the language model.

\tyt{Reflective attention}
Ke~\etal~\cite{ke2019reflective} introduced two reflective modules: while the first computes the relevance between hidden states from all the past predicted words and the current one, the second improves the syntactic structure of the sentence by guiding the generation process with words common position information.

\tyt{Look back and predict forward}
On a similar line, Qin~\etal~\cite{qin2019look} used two modules: the look back module that takes into account the previous attended vector to compute the next one, and the predict forward module that predicts the new two words at once, thus alleviating the accumulated errors problem that may occur at inference time.

\tyt{Adaptive attention time}
Huang~\etal~\cite{huang2019adaptively} proposed an adaptive attention time mechanism, in which the decoder can take an arbitrary number of attention steps for each generated word, determined by a confidence network on top of the second-layer LSTM.

\subsubsection{Boosting LSTM with Self-Attention}
Some works adopted the self-attention operator in place of the additive attention one in LSTM-based language models~\cite{huang2019attention,pan2020x,liu2020prophet,zhu2020autocaption}. In particular, Huang~\etal~\cite{huang2019attention} augmented the LSTM with the Attention on Attention operator, which computes another step of attention on top of visual self-attention. Pan~\etal~\cite{pan2020x} introduced the X-Linear attention block, which enhances self-attention with second-order interactions and improves both the visual encoding and the language model. On a different line, Zhu~\etal~\cite{zhu2020autocaption} applied the neural architecture search paradigm to select the connections between layers and the operations within gates of RNN-based image captioning language models, using a decoder enriched with self-attention~\cite{pan2020x}.

\subsection{Convolutional Language Models}
A worth-to-mention approach is that proposed by Aneya~\etal~\cite{aneja2018convolutional}, which uses convolutions as a language model. In particular, a global image feature vector is combined with word embeddings and fed to a CNN, operating on all words in parallel during training and sequentially in inference. Convolutions are right-masked to prevent the model from using the information of future word tokens. Despite the clear advantage of parallel training, the usage of the convolutional operator in language models has not gained popularity due to the poor performance and the advent of Transformer architectures.

\subsection{Transformer-based Architectures}

\begin{figure}[t]
\centering
\includegraphics[width=\linewidth]{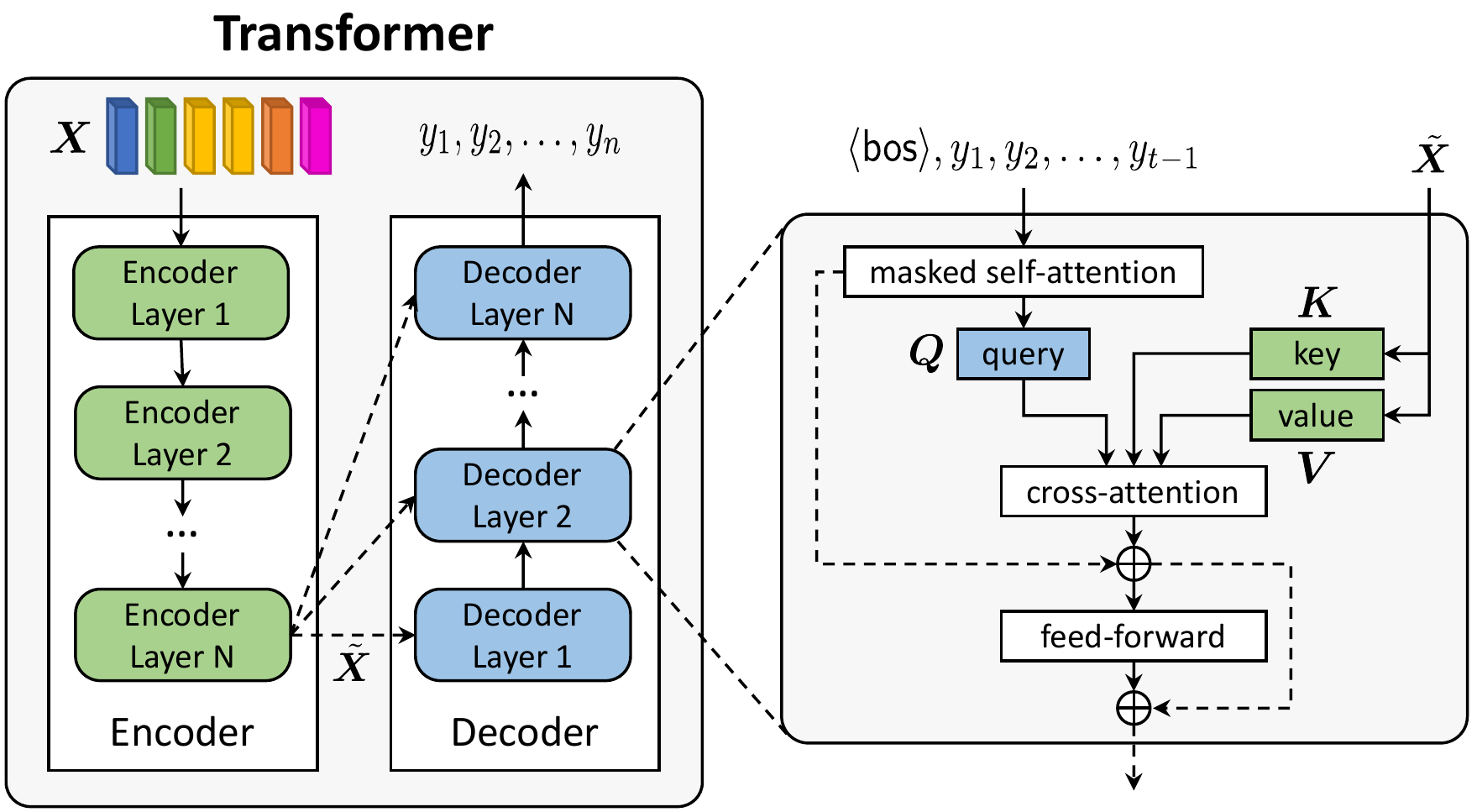}
\caption{Schema of the Transformer-based language model. The caption generation is performed via masked self-attention over previously generated tokens and cross-attention with encoded visual features.}
\label{fig:transformer}
\vspace{-0.25cm}
\end{figure}

The fully-attentive paradigm proposed by Vaswani~\etal~\cite{vaswani2017attention} has completely changed the perspective of language generation. Shortly after, the Transformer model became the building block of other breakthroughs in NLP, such as BERT~\cite{devlin2018bert} and GPT~\cite{radford2018improving}, and the standard de-facto architecture for many language understanding tasks. As image captioning can be cast as a sequence-to-sequence problem, the Transformer architecture has been employed also for this task. The standard Transformer decoder performs a masked self-attention operation, which is applied to words, followed by a cross-attention operation, where words act as queries and the outputs of the last encoder layer act as keys and values, plus a final feed-forward network (Fig.~\ref{fig:transformer}). During training, a masking mechanism is applied to the previous words to constrain a unidirectional generation process. The original Transformer decoder has been employed in some image captioning models without significant architectural modifications~\cite{herdade2019image,guo2020normalized,luo2021dual,wang2021simvlm}. Besides, some variants have been proposed to improve language generation and visual feature encoding.

\tit{Gating mechanisms}
Li~\etal~\cite{li2019entangled} proposed a gating mechanism for the cross-attention operator, which controls the flow of visual and semantic information by combining and modulating image regions representations with semantic attributes coming from an external tagger. On the same line, Ji~\etal~\cite{Ji2020ImprovingIC} integrated a context gating mechanism to modulate the influence of the global image representation on each generated word, modeled via multi-head attention. Cornia~\etal~\cite{cornia2020meshed} proposed to take into account all encoding layers in place of performing cross-attention only on the last one. To this end, they devised the meshed decoder, which contains a mesh operator that modulates the contribution of all the encoding layers independently and a gate that weights these contributions guided by the text query. In~\cite{wang2021simvlm,cornia2021universal}, the decoder architecture is again employed in conjunction with textual prefixes, also extracted from pre-trained visual-semantic models and employed as visual tags.

\subsection{BERT-like Architectures}
\begin{figure}[t]
\centering
\includegraphics[width=0.94\linewidth]{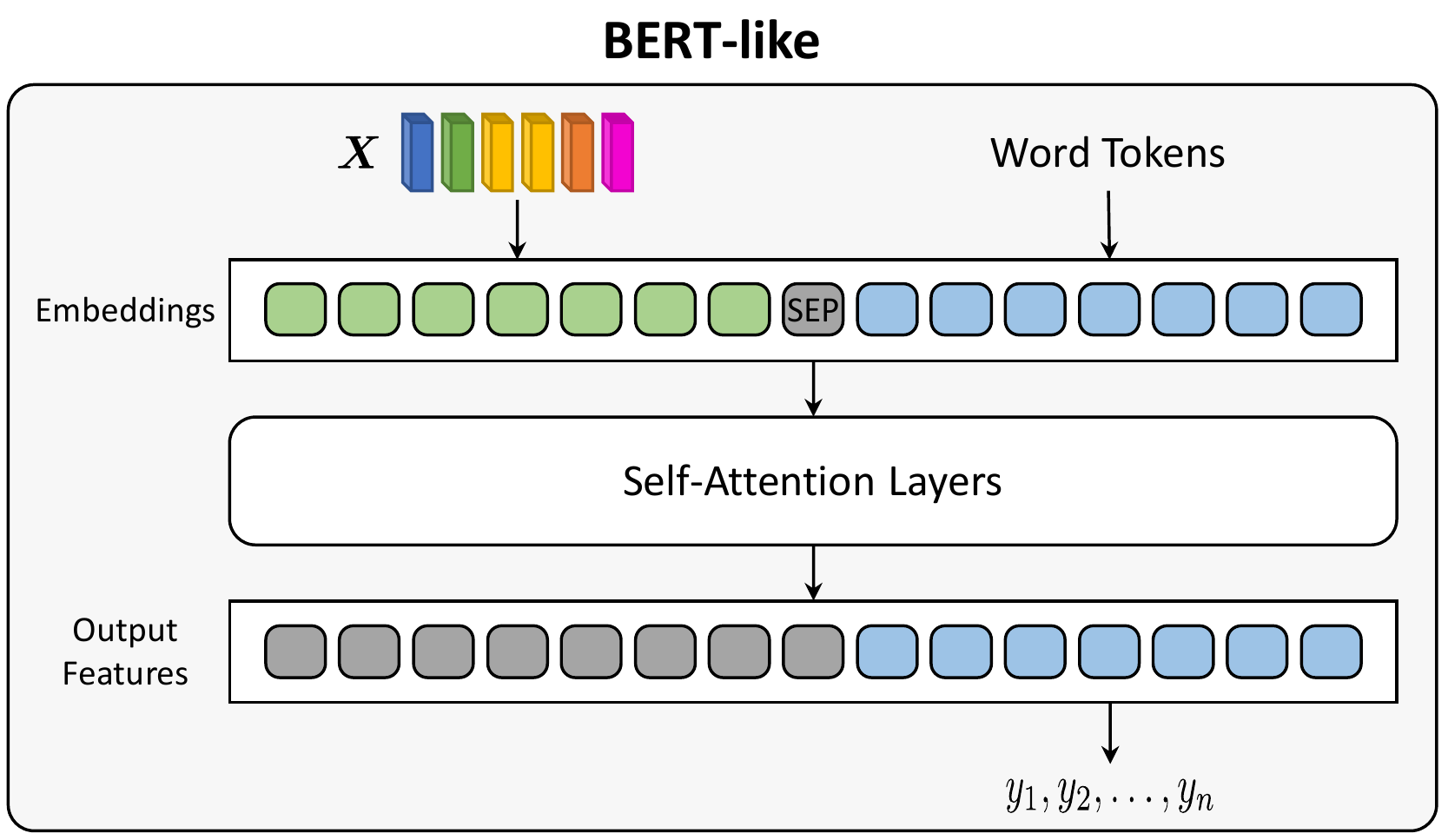}
\caption{Schema of a BERT-like language model. A single stream of attentive layers processes both image regions and word tokens and generates the output caption.}
\label{fig:bert}
\vspace{-0.25cm}
\end{figure}

Despite the encoder-decoder paradigm being a common approach to image captioning, some works have revisited captioning architectures to exploit a BERT-like~\cite{devlin2018bert} structure in which the visual and textual modalities are fused together in the early stages (Fig.~\ref{fig:bert}). The main advantage of this architecture is that layers dealing with text can be initialized with pre-trained parameters learned from massive textual corpora. Therefore, the BERT paradigm has been widely adopted in works that exploit pre-training~\cite{li2020oscar,zhou2020unified,zhang2021vinvl}. The first example is due to Zhou~\etal~\cite{zhou2020unified}, who developed a unified model that fuses visual and textual modalities into a BERT-like architecture for image captioning. The model consists of a shared multi-layer Transformer encoder network for both encoding and decoding, pre-trained on a large corpus of image-caption pairs and then fine-tuned for image captioning by right-masking the tokens sequence to simulate the unidirectional generation process. Further, Li~\etal~\cite{li2020oscar} introduced the usage of object tags detected in the image as anchors points for learning a better alignment in vision-and-language joint representations. To this end, their model represents an input image-text pair as a word tokens-object tags-region features triple, where the object tags are the textual classes proposed by the object detector. 

\subsection{Non-autoregressive Language Models}
Thanks to the parallelism offered by Transformers, non-autoregressive language models have been proposed in machine translation to reduce the inference time by generating all words in parallel. Some efforts have been made to apply this paradigm to image captioning~\cite{fei2019fast,guo2020non,fei2020iterative,guo2021fast}. The first approaches towards a non-autoregressive generation were composed of a number of different generation stages, where all words were predicted in parallel and refined at each stage. Subsequent methods, instead, employ reinforcement learning techniques to improve the final results. Specifically, these approaches treat the generation process as a cooperative multi-agent reinforcement system, where the positions in of the words in the target sequence are viewed as agents that learn to cooperatively maximize a sentence-level reward~\cite{guo2020non,guo2021fast}. These works also leverage knowledge distillation on unlabeled data and a post-processing step to remove identical consecutive tokens.

\subsection{Discussion}
Recurrent models have been the standard for many years, and their application brought to the development of clever and successful ideas that can be integrated also into non-recurrent solutions. However, they are slow to train and struggle to maintain long-term dependencies: these drawbacks are alleviated by autoregressive and Transformer-based solutions that recently gained popularity. Inspired by the success of pre-training on large, unsupervised corpora for NLP tasks, massive pre-training has been applied also for image captioning by employing either encoder-decoder or BERT-like architectures, often in conjunction with textual tags. This strategy led to impressive performance, suggesting that visual and textual semantic relations can be inferred and learned also from not well-curated data~\cite{li2020oscar,wang2021simvlm,hu2021scaling}. BERT-like architectures are suitable for such a massive pre-training but are not generative architectures by design. Massive pre-training on generative-oriented architectures~\cite{wang2021simvlm,cornia2021universal} is currently a worth-exploring direction, which leads to performances that are at least on-pair with the early-fusion counterparts.

%% file: 04-training_strategies.tex
\section{Training Strategies}\label{sec:training_strategies}
An image captioning model is commonly expected to generate a caption word by word by taking into account the previous words and the image. At each step, the output word is sampled from a learned distribution over the vocabulary words. In the most simple scenario, \ie~the greedy decoding mechanism, the word with the highest probability is output. The main drawback of this setting is that possible prediction errors quickly accumulate along the way. To alleviate this drawback, one effective strategy is to use the beam search algorithm~\cite{koehn2009statistical} that, instead of outputting the word with maximum probability at each time step, maintains $k$ sequence candidates (those with the highest probability at each step) and finally outputs the most probable one. 

During training, the captioning model must learn to properly predict the probabilities of the words to appear in the caption. To this end, the most common training strategies are based on 1.~\emph{cross-entropy loss}; 2.~\emph{masked language model}; 3.~\emph{reinforcement learning} that allows directly optimizing for captioning-specific non-differentiable metrics; 4.~\emph{vision-and-language pre-training} objectives (see Fig.~\ref{fig:first_page}).

\subsection{Cross-Entropy Loss}
The cross-entropy loss is the first proposed and most used objective for image captioning models. With this loss, the goal of the training, at each timestep, is to minimize the negative log-likelihood of the current word given the previous ground-truth words. Given a sequence of target words $y_{1:T}$, the loss is formally defined as:
\begin{equation}
L_{XE}(\theta)=-\sum_{i=1}^{n} \log \left(P\left(y_{i} \mid y_{1: i-1},\bm{X}\right)\right),
\end{equation}
where $P$ is the probability distribution induced by the language model, $y_{i}$ the ground-truth word at time $i$, $y_{1: i-1}$ indicate the previous ground-truth words, and $\bm{X}$ the visual encoding. The cross-entropy loss is designed to operate at word level and optimize the probability of each word in the ground-truth sequence without considering longer range dependencies between generated words. The traditional training setting with cross-entropy also suffers from the exposure bias problem~\cite{ranzato2015sequence} caused by the discrepancy between the training data distribution as opposed to the distribution of its own predicted words. 

\subsection{Masked Language Model (MLM)}
The first masked language model has been proposed for training the BERT~\cite{devlin2018bert} architecture. The main idea behind this optimization function consists in randomly masking out a small subset of the input tokens sequence and training the model to predict masked tokens while relying on the rest of the sequence, \ie~both previous and subsequent tokens. As a consequence, the model learns to employ contextual information to infer missing tokens, which allows building a robust sentence representation where the context plays an essential role. Since this strategy considers only the prediction of the masked tokens and ignores the prediction of the non-masked ones, training with it is much slower than training for complete left-to-right or right-to-left generation. Notably, some works have employed this strategy as a pre-training objective, sometimes completely avoiding the combination with the cross-entropy~\cite{li2020oscar,zhang2021vinvl}.

\subsection{Reinforcement Learning}
Given the limitations of word-level training strategies observed when using limited amounts of data, a significant improvement was achieved by applying the reinforcement learning paradigm for training image captioning models. Within this framework, the image captioning model is considered as an agent whose parameters determine a policy. At each time step, the agent executes the policy to choose an action, \ie~the prediction of the next word in the generated sentence. Once the end-of-sequence is reached, the agent receives a reward, and the aim of the training is to optimize the agent parameters to maximize the expected reward.

Many works harnessed this paradigm and explored different sequence-level metrics as rewards. The first proposal is due to Ranzato~\etal~\cite{ranzato2015sequence}, which introduced the usage of the REINFORCE algorithm~\cite{williams1992simple} adopting BLEU~\cite{papineni2002bleu} and ROUGE~\cite{lin2004rouge} as reward signals. Ren~\etal~\cite{ren2017deep} experimented using visual-semantic embeddings obtained from a network that encodes the image and the so far generated caption in order to compute a similarity score to be used as reward. Liu~\etal~\cite{liu2017improved} proposed to use as reward a linear combination of SPICE~\cite{spice2016} and CIDEr~\cite{vedantam2015cider}, called SPIDEr. Finally, the most widely adopted strategy~\cite{zhang2017actor,gao2019self,cornia2020meshed}, introduced by Rennie~\etal~\cite{rennie2017self}, entails using the CIDEr score, as it correlates better with human judgment~\cite{vedantam2015cider}. The reward is normalized with respect to a baseline value to reduce variance. Formally, to compute the loss gradient, beam search and greedy decoding are leveraged as follows:
\begin{equation}
    \nabla_\theta L(\theta) = -\frac{1}{k}\sum_{i=1}^k \left((r(\bm{w}^i)-b) \nabla_\theta \log P(\bm{w}^i)\right),
\end{equation}
where $\bm{w}^i$ is the $i$-th sentence in the beam or a sampled collection, $r(\cdot)$ is the reward function, \ie~the CIDEr computation, and $b$ is the baseline, computed as the reward of the sentence obtained via greedy decoding~\cite{rennie2017self}, or as the average reward of the beam candidates~\cite{cornia2020meshed}. 

Note that, since it would be difficult for a random policy to improve in an acceptable amount of time, the usual procedure entails pre-training with cross-entropy or masked language model first, and then fine-tuning stage with reinforcement learning by employing a sequence level metric as reward. This ensures the initial reinforcement learning policy to be more suitable than the random one.

\begin{figure*}[t]
\centering
\subfloat[]{
\includegraphics[height=0.35\linewidth]{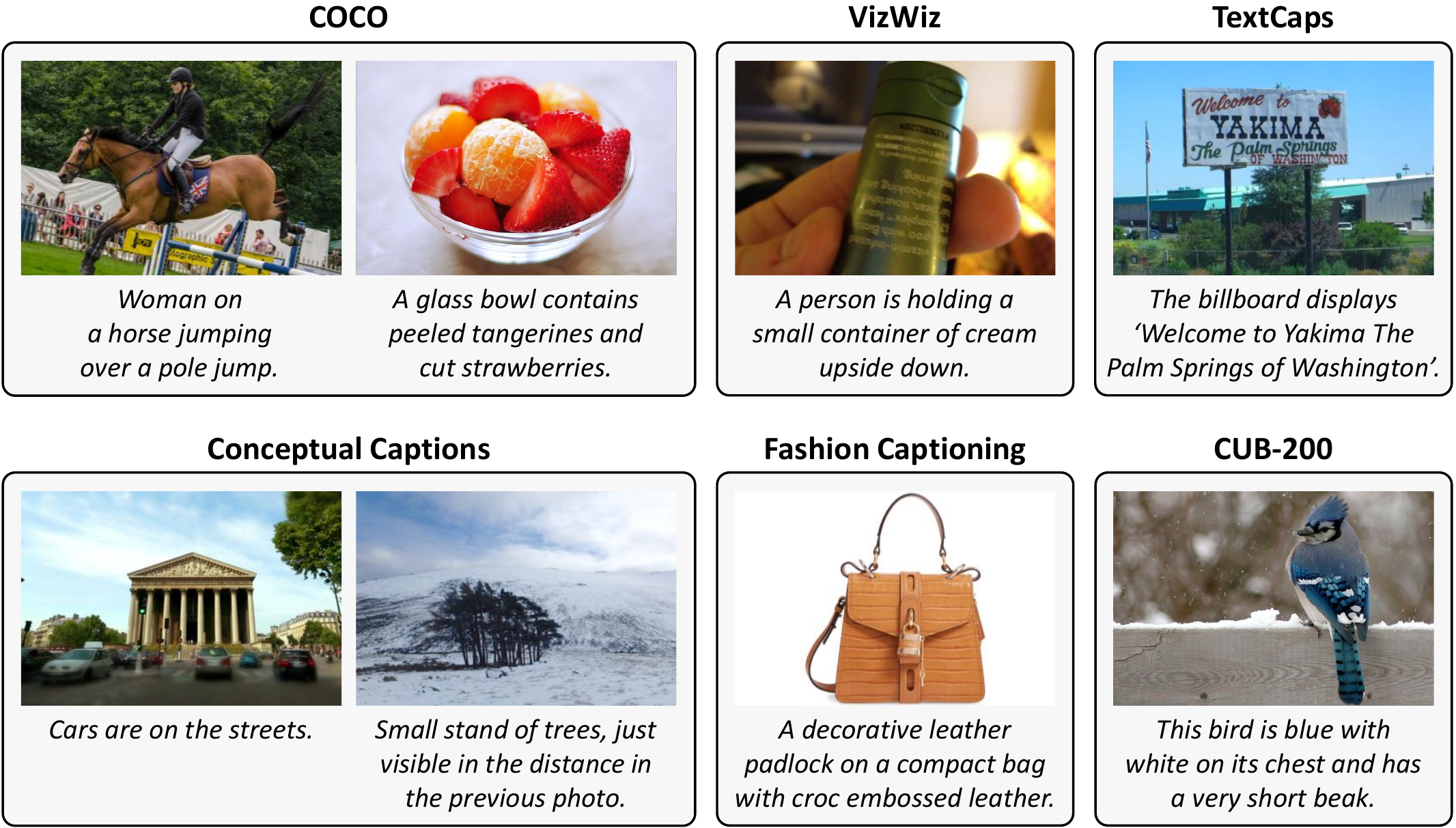}
\label{sfig:datasets}}
\subfloat[]{
\includegraphics[height=0.35\linewidth]{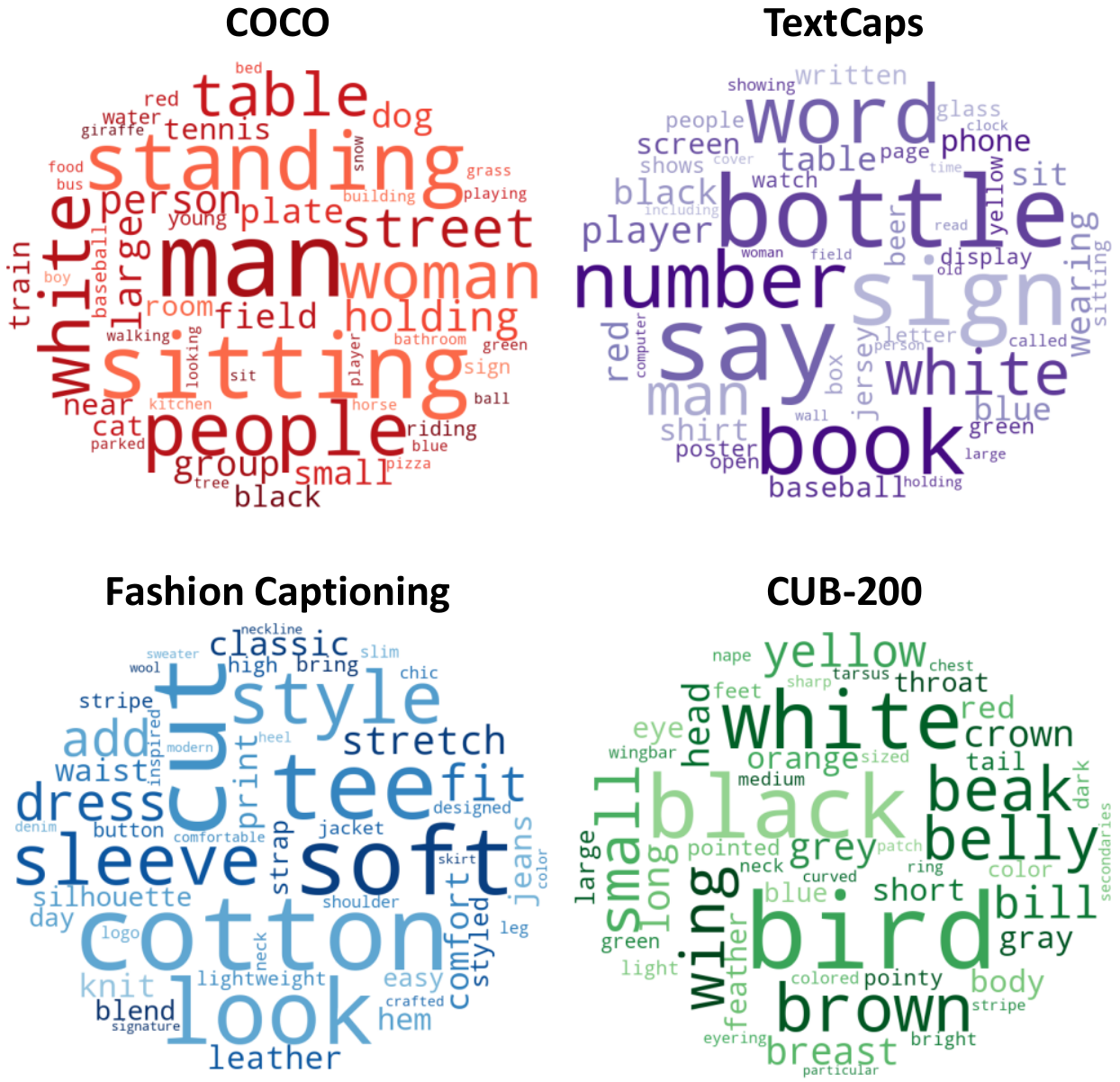}
\label{sfig:clouds}}
\vspace{-0.15cm}
\caption{Qualitative examples from some of the most common image captioning datasets: \textbf{(a)} image-caption pairs; \textbf{(b)} word clouds of the captions most common visual words.}
\label{fig:datasets}
\vspace{-0.25cm}
\end{figure*}

\subsection{Large-scale Pre-Training}
In the context of vision-and-language pre-training in early-fusion architectures, one of the most common pre-training objectives is the masked contextual token loss, where tokens of each modality (visual and textual) are randomly masked following the BERT strategy~\cite{devlin2018bert}, and the model has to predict the masked input based on the context of both modalities, thus connecting their joint representation. Another largely adopted strategy entails using a contrastive loss, where the inputs are organized as image regions-captions words-object tags triples, and the model is asked to discriminate correct triples from polluted ones, in which tags are randomly replaced~\cite{li2020oscar,zhang2021vinvl}. Other objectives take into account the text-image alignment at a word-region level and entail predicting the original word sequence given a corrupted one~\cite{xia2020xgpt}.

On the other hand, cross-entropy has also been used when pre-training on noisy captions~\cite{cornia2021universal,wang2021simvlm}, sometimes also employing prefixes. PrefixLM~\cite{wang2021simvlm} has indeed proved to be a valuable strategy that enables bidirectional attention within the prefix sequence, and thus, it is applicable for both decoder-only and encoder-decoder sequence-to-sequence language models. Noticeably, some large-scale models pre-trained on noisy data under this setting can achieve state-of-the-art performance without requiring a fine-tuning stage with Reinforcement~\cite{wang2021simvlm}.

Finally, we notice that image captioning can be used as a pre-training task to efficiently learn visual representations, which can benefit downstream tasks such as image classification, object detection, and instance segmentation~\cite{desai2021virtex}.

%% file: 05-evaluation_protocol.tex
\section{Evaluation Protocol}
As for any data-driven task, the development of image captioning has been enabled by the collection of large datasets and the definition of quantitative scores to evaluate the performance and monitor the advancement of the field.

\subsection{Datasets}
Image captioning datasets contain images and one or multiple captions associated with them. 
Having multiple ground-truth captions for each image helps to capture the variability of human descriptions. Other than the number of available captions, also their characteristics (\eg~average caption length and vocabulary size) highly influence the design and the performance of image captioning algorithms. 
Note that the distribution of the terms in the datasets captions is usually long-tailed, thus, when using word-level dictionaries, the common practice is to include in the vocabulary only those terms whose frequency is above a pre-defined threshold. Recently, however, using subword-based tokenization approaches like BPE~\cite{sennrich2016neural} is a popular choice that allows avoiding dataset pre-processing. The available datasets differ both on the images contained (for their domain and visual quality) and on the captions associated with the images (for their length, number, relevance, and style). A summary of the most used public datasets is reported in Table~\ref{tab:datasets}, and some sample image-caption pairs are reported in Fig.~\ref{fig:datasets}, along with some word clouds obtained from the 50 most used visual words in the captions.

\subsubsection{Standard captioning datasets}
Standard benchmark datasets are used by the community to compare their approaches on a common test-bed, a procedure that guides the development of image captioning strategies by allowing to identify suitable directions. Datasets used as benchmarks should be representative of the task at hand, both in terms of the challenges and ideal expected results (\ie~achievable human performance). Further, they should contain a large number of generic-domain images, each associated with multiple captions.

\definecolor{lightgray}{gray}{0.97}
\begin{table}
\caption{Overview of the main image captioning datasets.}
\label{tab:datasets}
\setlength{\tabcolsep}{.3em}
\renewcommand{\arraystretch}{1.1}
\resizebox{\linewidth}{!}{
\rowcolors{3}{}{lightgray}
\begin{tabular}{lccccccc}
\toprule
& & \multirow{2}{*}{\textbf{Domain}} & \multirow{2}{*}{\textbf{Nb. Images}} & \textbf{Nb. Caps} & \multirow{2}{*}{\textbf{Vocab Size}} & \textbf{Nb. Words}\\
& & & & \textbf{(per Image)} & & \textbf{(per Cap.)} \\
\midrule
COCO~\cite{lin2014microsoft} & & Generic & $132$K & $5$ & $27$K ($10$K) & $10.5$\\
Flickr30K~\cite{young2014image} & & Generic & $31$K & $5$ & $18$K ($7$K) & $12.4$ \\
Flickr8K~\cite{hodosh2013framing} & & Generic & $8$K & $5$ & $8$K ($3$K) & $10.9$ \\
\midrule
CC3M~\cite{sharma2018conceptual} & & Generic & $3.3$M & $1$ & $48$K ($25$K) & $10.3$\\
CC12M~\cite{changpinyo2021conceptual} & & Generic & $12.4$M & $1$ & $523$K ($163$K) & $20.0$\\
SBU Captions~\cite{ordonez2011im2text} & & Generic & $1$M & $1$ & $238$K ($46$K) & $12.1$ \\
\midrule
VizWiz~\cite{gurari2020captioning} & & Assistive & $70$K & $5$ & $20$K ($8$K)  & $13.0$ \\
CUB-200~\cite{reed2016learning} & & Birds & $12$K & $10$ & $6$K ($2$K) & $15.2$ \\
Oxford-102~\cite{reed2016learning} & & Flowers & $8$K & $10$ & $5$K ($2$K) & $14.1$ \\
Fashion Cap.~\cite{yang2020fashion} & & Fashion & $130$K & $1$ & $17$K ($16$K) & $21.0$ \\
BreakingNews~\cite{ramisa2017breakingnews} & & News & $115$K & $1$ & $85$K ($10$K) & $28.1$ \\
GoodNews~\cite{biten2019good} & & News & $466$K & $1$ & $192$K ($54$K) & $18.2$ \\
TextCaps~\cite{sidorov2020textcaps} & & OCR & $28$K & $5/6$ & $44$K ($13$K) & $12.4$ \\
Loc. Narratives~\cite{pont2020connecting} & & Generic & $849$K & $1/5$ & $16$K ($7$K) & $41.8$ \\
\bottomrule
\end{tabular}}
\vspace{-0.25cm}
\end{table}

Early image captioning architectures~\cite{mao2015deep,donahue2015long,karpathy2015deep} were commonly trained and tested on the \textbf{Flickr30K}~\cite{young2014image} and \textbf{Flickr8K}~\cite{hodosh2013framing} datasets, consisting of pictures collected from the Flickr website, containing everyday activities, events, and scenes, paired with five captions each. Currently, the most commonly used dataset is \textbf{Microsoft COCO}~\cite{lin2014microsoft}, which consists of images of complex scenes with people, animals, and common everyday objects in their context. It contains more than 120,000 images, each annotated with five captions, divided into 82,783 images for training and 40,504 for validation. For ease of evaluation, most of the literature follows the splits defined by Karpathy~\etal~\cite{karpathy2015deep}, where 5,000 images of the original validation set are used for validation, 5,000 for test, and the rest for training. The dataset has also an official test set, composed of 40,775 images paired with 40 private captions each, and a public evaluation server\footnote{\url{https://competitions.codalab.org/competitions/3221}}. 

\subsubsection{Pre-training datasets} 
Although training on large well-curated datasets is a sound approach, some works~\cite{lu2019vilbert, li2020oscar,wang2021simvlm,hu2021scaling} have demonstrated the benefits of pre-training on even bigger vision-and-language datasets, which can be either image captioning datasets of lower-quality captions or datasets collected for other tasks (\eg~visual question answering~\cite{li2020oscar,zhou2020unified}, text-to-image generation~\cite{ramesh2021zero}, image-caption association~\cite{radford2021learning}). 
Among the datasets used for pre-training, that have been specifically collected for image captioning, it is worth mentioning \textbf{SBU Captions}~\cite{ordonez2011im2text}, originally used for tackling image captioning as a retrieval task~\cite{hodosh2013framing}, which contains around 1 million image-text pairs, collected from the Flickr website. Similarly, \textbf{YFCC100M}~\cite{thomee2016yfcc100m} is composed of 100 million media objects in which 14.8 million images are available with automatically-collected textual descriptions.
Later, the \textbf{Conceptual Captions}~\cite{sharma2018conceptual,changpinyo2021conceptual} datasets have been proposed, which are collections of around 3.3 million (CC3M) and 12 million (CC12M) images paired with one weakly-associated description automatically collected from the web with a relaxed filtering procedure. Differently from previous datasets, \textbf{Wikipedia-based Image Text} (WIT)~\cite{srinivasan2021wit} provides images coming from Wikipedia together with various metadata extracted from the original pages, with approximately 5.3 million images available with the corresponding descriptions in English.
Although the large scale and variety in caption style make all these datasets particularly interesting for pre-training, the contained captions can be noisy, and the availability of images is not always guaranteed since most of them are provided as URLs.

Pre-training on such datasets requires significant computational resources and effort to collect the data needed. Nevertheless, this strategy represents an asset to obtain state-of-the-art performances. Accordingly, some pre-training datasets are currently not publicly available, such as \textbf{ALIGN}~\cite{jia2021scaling,wang2021simvlm} and \textbf{ALT-200}~\cite{hu2021scaling}, respectively containing 1.8 billion and 200 million noisy image-text pairs, or the datasets used to train DALL-E~\cite{ramesh2021zero} and CLIP~\cite{radford2021learning} consisting of 250 and 400 million pairs.

\subsubsection{Domain-specific datasets}
While domain-generic benchmark datasets are important to capture the main aspects of the image captioning task, domain-specific datasets are also important to highlight and target specific challenges. These may relate to the visual domain (\eg~type and style of the images) and the semantic domain. In particular, the distribution of the terms used to describe domain-specific images can be significantly different from that of the terms used for domain-generic images.

An example of dataset-specific in terms of the visual domain is the \textbf{VizWiz Captions}~\cite{gurari2020captioning} dataset, collected to favor the image captioning research towards assistive technologies. The images in this dataset have been taken by visually-impaired people with their phones, thus, they can be of low quality and concern a wide variety of everyday activities, most of which entail reading some text. 

Some examples of specific semantic domain are the \textbf{CUB-200}~\cite{welinder2010caltech} and the \textbf{Oxford-102}~\cite{nilsback2008automated} datasets, which contain images of birds and flowers, respectively, that have been paired with ten captions each by Reed~\etal~\cite{reed2016learning}. Given the specificity of these datasets, rather than for standard image captioning, they are usually adopted for different related tasks such as cross-domain captioning~\cite{chen2017show}, visual explanation generation~\cite{hendricks2016generating,hendricks2018grounding}, and text-to-image synthesis~\cite{reed2016generative}. Another domain-specific dataset is \textbf{Fashion Captioning}~\cite{yang2020fashion} that contains images of clothing items in different poses and colors that may share the same caption. The vocabulary for describing these images is somewhat smaller and more specific than for generic datasets. Differently, datasets as \textbf{BreakingNews}~\cite{ramisa2017breakingnews} and \textbf{GoodNews}~\cite{biten2019good} enforce using a richer vocabulary since their images, taken from news articles, have long associated captions written by expert journalists. The same applies to the \textbf{TextCaps}~\cite{sidorov2020textcaps} dataset, which contains images with text, that must be ``read'' and included in the caption, and to \textbf{Localized Narratives}~\cite{pont2020connecting}, whose captions have been collected by recording people freely narrating what they see in the images.

Collecting domain-specific datasets and developing solutions to tackle the challenges they pose is crucial to extend the applicability of image captioning algorithms.

\subsection{Evaluation Metrics}\label{ssec:metrics}
Evaluating the quality of a generated caption is a tricky and subjective task~\cite{vedantam2015cider, spice2016}, complicated by the fact that captions cannot only be grammatical and fluent but need to properly refer to the input image. Arguably, the best way to measure the quality of the caption for an image is still carefully designing a human evaluation campaign in which multiple users score the produced sentences~\cite{kasai2021transparent}. However, human evaluation is costly and not reproducible -- which prevents a fair comparison between different approaches. Automatic scoring methods exist that are used to assess the quality of system-produced captions, usually by comparing them with human-produced reference sentences, although some metrics do not rely on reference captions.  Table~\ref{tab:metrics}. 

\subsubsection{Standard evaluation metrics}
The first strategy adopted to evaluate image captioning performance consists of exploiting metrics designed for NLP tasks. For example, the \textbf{BLEU} score~\cite{papineni2002bleu} and the \textbf{METEOR}~\cite{banerjee2005meteor} score were introduced for machine translation. The former is based on \emph{n}-gram precision considering \emph{n}-grams up to length four; the latter favors the recall of matching unigrams from the candidate and reference sentences in their exact form stemmed form and meaning. Moreover, the \textbf{ROUGE} score~\cite{lin2004rouge} was designed for summarization and applied also for image captioning in its variant considering the longest subsequence of tokens in the same relative order, possibly with other tokens in-between, that appears in both candidate and reference caption.
Later, specific image captioning metrics have been proposed~\cite{vedantam2015cider, spice2016}. The reference \textbf{CIDEr} score~\cite{vedantam2015cider} is based on the cosine similarity between the Term Frequency-Inverse Document Frequency weighted \emph{n}-grams in the candidate caption and in the set of reference captions associated with the image, thus taking into account both precision and recall. The \textbf{SPICE} score~\cite{spice2016} considers matching tuples extracted from the candidate and the reference (or possibly directly the image) scene graphs, thus favoring the semantic content rather than the fluency.

\begin{table*}
\caption{Performance analysis of representative image captioning approaches in terms of different evaluation metrics. The $\dagger$ marker indicates models trained by us with ResNet-152 features, while the $\ddagger$ marker indicates unofficial implementations. For all the metrics, the higher the value, the better ($\uparrow$).}
\label{tab:results}
\setlength{\tabcolsep}{.3em}
\renewcommand{\arraystretch}{1.1}
\rowcolors{7}{}{lightgray}
\resizebox{\linewidth}{!}{
\begin{tabular}{lccccccccccccccccccccccccccc}
\toprule
& & & & \multicolumn{6}{c}{\textbf{Standard Metrics}} & & & \multicolumn{4}{c}{\textbf{Diversity Metrics}} & & & \multicolumn{3}{c}{\textbf{Embedding-based Metrics}} & & & \multicolumn{4}{c}{\textbf{Learning-based Metrics}}  \\
\cmidrule{5-10} \cmidrule{13-16} \cmidrule{19-21} \cmidrule{24-27}
& & \textbf{\#Params (M)} & & \textbf{B-1} & \textbf{B-4} & \textbf{M} & \textbf{R} & \textbf{C} & \textbf{S} & & & \textbf{Div-1} & \textbf{Div-2} & \textbf{Vocab} & \textbf{\%Novel} & & & \textbf{WMD} & \textbf{Alignment} & \textbf{Coverage} & & & \textbf{TIGEr} & \textbf{BERT-S} & \textbf{CLIP-S} & \textbf{CLIP-S$^\text{Ref}$}\\
\midrule
Show and Tell$^\dagger$~\cite{vinyals2015show} & & 13.6 & & 72.4 & 31.4 & 25.0 & 53.1 & 97.2 & 18.1 & & & 0.014 & 0.045 & 635 & 36.1 & & & 16.5 & 0.199 & 71.7 & & & 71.8 & 93.4 & 0.697 & 0.762 \\
SCST (FC)$^\ddagger$~\cite{rennie2017self} & & 13.4 & & 74.7 & 31.7 & 25.2 & 54.0 & 104.5 & 18.4 & & & 0.008 & 0.023 & 376 & 60.7 & & & 16.8 & 0.218 & 74.7 & & & 71.9 & 89.0 & 0.691 & 0.758 \\
Show, Attend and Tell$^\dagger$~\cite{xu2015show} & & 18.1 & & 74.1 & 33.4 & 26.2 & 54.6 & 104.6 & 19.3 & & & 0.017 & 0.060 & 771 & 47.0 & & & 17.6 & 0.209 & 72.1 & & & 73.2 & 93.6 & 0.710 & 0.773 \\
\midrule
SCST (Att2in)$^\ddagger$~\cite{rennie2017self} & & 14.5 & & 78.0 & 35.3 & 27.1 & 56.7 & 117.4 & 20.5 & & & 0.010 & 0.031 & 445 & 64.9 & & & 18.5 & 0.238 & 76.0 & & & 73.9 & 88.9 & 0.712 & 0.779 \\
Up-Down$^\ddagger$~\cite{anderson2018bottom} & & 52.1 & & 79.4 & 36.7 & 27.9 & 57.6 & 122.7 & 21.5 & & & 0.012 & 0.044 & 577 & 67.6 & & & 19.1 & 0.248 & 76.7 & & & 74.6 & 88.8 & 0.723 & 0.787 \\
SGAE~\cite{yang2019auto} & & 125.7 & & 81.0 & 39.0 & 28.4 & 58.9 & 129.1 & 22.2 & & & 0.014 & 0.054 & 647 & 71.4 & & & 20.0 & 0.255 & 76.9 & & & 74.6 & 94.1 & 0.734 & 0.796 \\
MT~\cite{shi2020improving} & & 63.2 & & 80.8 & 38.9 & 28.8 & 58.7 & 129.6 & 22.3 & & & 0.011 & 0.048 & 530 & 70.4 & & & 20.2 & 0.253 & 77.0 & & & 74.8 & 88.8 & 0.726 & 0.791 \\
AoANet~\cite{huang2019attention} & & 87.4 & & 80.2 & 38.9 & 29.2 & 58.8 & 129.8 & 22.4 & & & 0.016 & 0.062 & 740 & 69.3 & & & 20.0 & 0.254 & 77.3 & & & 75.1 & 94.3 & 0.737 & 0.797 \\
X-LAN~\cite{pan2020x}  & & 75.2 & & 80.8 & 39.5 & 29.5 & 59.2 & 132.0 & 23.4 & & & 0.018 & 0.078 & 858 & 73.9 & & & 20.6 & 0.261 & 77.9 & & & 75.4 & 94.3 & 0.746 & 0.803 \\
DPA~\cite{liu2020prophet} & & 111.8 & & 80.3 & 40.5 & 29.6 & 59.2 & 133.4 & 23.3 & & & 0.019 & 0.079 & 937 & 65.9 & & & 20.5 & 0.261 & 77.3 & & & 75.0 & 94.3 & 0.738 & 0.802 \\
AutoCaption~\cite{zhu2020autocaption}  & & - & & 81.5 & 40.2 & 29.9 & 59.5 & 135.8 & 23.8 & & & 0.022 & 0.096 & 1064 & 75.8 & & & 20.9 & 0.262 & 77.7 & & & 75.4 & 94.3 & 0.752 & 0.808 \\
\midrule
ORT~\cite{herdade2019image} & & 54.9 & & 80.5 & 38.6 & 28.7 & 58.4 & 128.3 & 22.6 & & & 0.021 & 0.072 & 1002 & 73.8 & & & 19.8 & 0.255 & 76.9 & & & 75.1 & 94.1 & 0.736 & 0.796    \\
CPTR~\cite{liu2021cptr} & & 138.5 & & 81.7 & 40.0 & 29.1 & 59.4 & 129.4 & - & & & 0.014 & 0.068 & 667 & 75.6 & & & 20.2 & 0.261 & 77.0 & & & 74.8 & 94.3 & 0.745 & 0.802    \\
$\mathcal{M}^2$ Transformer~\cite{cornia2020meshed} & & 38.4 & & 80.8 & 39.1 & 29.2 & 58.6 & 131.2 & 22.6 & & & 0.017 & 0.079 & 847 & \textbf{78.9} & & & 20.3 & 0.256 & 76.0 & & & 75.3 & 93.7 & 0.734 & 0.792\\
X-Transformer~\cite{pan2020x}  & & 137.5 & & 80.9 & 39.7 & 29.5 & 59.1 & 132.8 & 23.4 & & & 0.018 & 0.081 & 878 & 74.3 & & & 20.6 & 0.257 & 77.7 & & & 75.5 & 94.3 & 0.747 & 0.803 \\
\midrule
Unified VLP~\cite{zhou2020unified} & & 138.2 & & 80.9 & 39.5 & 29.3 & 59.6 & 129.3 & 23.2 & & & 0.019 & 0.081 & 898 & 74.1 & & & \textbf{26.6} & 0.258 & 77.1 & & & 75.1 & \textbf{94.4} & 0.750 & 0.807    \\
VinVL~\cite{zhang2021vinvl}  & & 369.6 & & \textbf{82.0} & \textbf{41.0} & \textbf{31.1} & \textbf{60.9} & \textbf{140.9} & \textbf{25.2} & & & \textbf{0.023} & \textbf{0.099} & \textbf{1125} & 77.9 & & & 20.5 & \textbf{0.265} & \textbf{79.6} & & & \textbf{75.7} & 88.5 & \textbf{0.766} & \textbf{0.820} \\
\bottomrule
\end{tabular}
}
\vspace{-0.25cm}
\end{table*}

As expected, metrics designed for image captioning usually correlate better with human judgment than those borrowed from other NLP tasks (with the exception of METEOR~\cite{banerjee2005meteor}), both at corpus-level and caption-level~\cite{spice2016, sharif2018nneval, cui2018learning}. Correlation with human judgment is measured via statistical correlation coefficients (such as Pearson's, Kendall's, and Spearman's correlation coefficients) and via the agreement with humans' preferred caption in a pair of candidates, all evaluated on sample captioned images.

\subsubsection{Diversity metrics}
To better assess the performance of a captioning system, it is common practice to consider a set of the above-mentioned standard metrics. Nevertheless, these are somehow gameable because they favor word similarity rather than meaning correctness~\cite{caglayan2020curious}.
Another drawback of the standard metrics is that they do not capture (but rather disfavor) the desirable capability of the system to produce novel and diverse captions, which is more in line with the variability with which humans describe complex images. This consideration brought to the development of diversity metrics~\cite{shetty2017speaking,van2018measuring,wang2019describing,wang2020diversity}. Most of these metrics can potentially be calculated even when no ground-truth captions are available at test time. However, since they overlook the syntactic correctness of the captions and their relatedness with the image, it is advisable to combine them with other metrics.

The overall performance of a captioning system can be evaluated in terms of corpus-level diversity or, when the system can output multiple captions for the same image, single image-level diversity (termed as \emph{global diversity} and \emph{local diversity}, respectively, in~\cite{van2018measuring}). To quantify the former, it can be considered the number of unique words used in all the generated captions (\textbf{Vocab}) and the percentage of generated captions that were not present in the training set (\textbf{\%Novel}). For the latter, it can be used the ratio of unique captions unigrams or bigrams to the total number of captions unigrams (\textbf{Div-1} and \textbf{Div-2}). 

\subsubsection{Embedding-based metrics}
An alternative approach to captioning evaluation consists in relying on captions semantic similarity or other specific aspects of caption quality, which are estimated via embedding-based metrics~\cite{rohrbach2018object,jiang2019reo,wang2020towards}.
For example, the \textbf{WMD} score~\cite{kusner2015word}, originally introduced to evaluate document semantic dissimilarity, can also be applied to captioning evaluation by considering generated captions and ground-truth captions as the compared documents~\cite{kilickaya2017re}. Moreover, the \textbf{Alignment} score~\cite{cornia2019show} is based on the alignment between the sequences of nouns in the candidate and reference sentence and captures whether concepts are mentioned in a human-like order. Finally, the \textbf{Coverage} score~\cite{cornia2020smart,bigazzi2020explore} expresses the completeness of a caption, which is evaluated by considering the mentioned scene visual entities. Since this score considers visual objects directly, it can be applied even when no ground-truth caption is available.

\subsubsection{Learning-based evaluation}
As a further development towards captions quality assessment, learning-based evaluation strategies~\cite{sharif2018nneval,cui2018learning,lee2020vilbertscore,yi2020improving,wang2021faier,lee2021umic} are being investigated. To this end, it can be exploited a component of a complete captioning approach, in charge to evaluate the produced caption completeness~\cite{dai2018neural} or how human-like it is~\cite{dai2017towards}. Alternatively, learning-based evaluation is usually based on a pre-trained model. For example, the \textbf{BERT-S} score~\cite{zhang2020bertscore}, which is used to evaluate various language generation tasks~\cite{unanue2021berttune}, exploits pre-trained BERT embeddings~\cite{devlin2018bert} to represent and match the tokens in the reference and candidate sentences via cosine similarity. Moreover, the \textbf{TIGEr} score~\cite{jiang2019tiger} represents the reference and candidate captions as grounding score vectors obtained from a pre-trained model~\cite{lee2018stacked} that grounds their words on the image regions and scores the candidate caption based on the similarity of the grounding vectors. 
Further, the \textbf{CLIP-S} score~\cite{hessel2021clipscore} is a direct application of the CLIP~\cite{radford2021learning} model to image captioning evaluation and consists of an adjusted cosine similarity between image and candidate caption representation. Thus, CLIP-S is designed to work without reference captions, although the CLIP-S$^\text{Ref}$ variant can exploit also the reference captions.

\noindent We refer the reader to Appendix~\ref{sec:appendix_metrics} for a deeper discussion on diversity, embedding-based, and learning-based metrics.

%% file: 06-results.tex
\begin{figure*}[t]
    \centering
    \includegraphics[width=\textwidth]{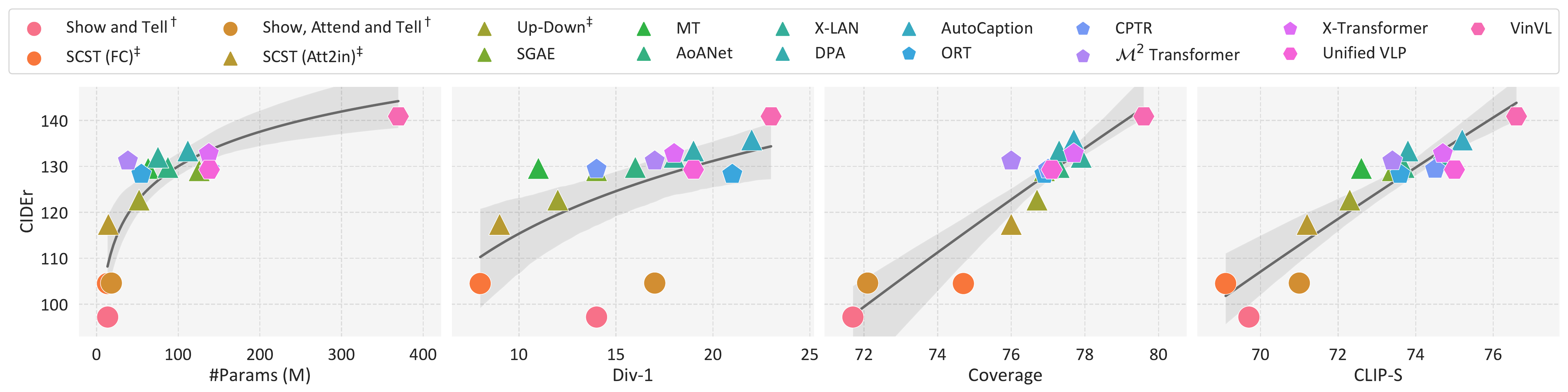}
    \caption{Relationship between CIDEr, number of parameters and other scores. 
    Values of Div-1 and CLIP-S are multiplied by powers of 10 for readability.} 
    \label{fig:cider_vs_all}
    \vspace{-0.25cm}
\end{figure*}

\section{Experimental Evaluation}\label{sec:results}
In Table~\ref{tab:results}, we analyze the performance of some of the main approaches in terms of all the evaluation scores presented in Section~\ref{ssec:metrics} to take into account the different aspects of caption quality these express and report their number of parameters to give an idea of the computational complexity and memory occupancy of the models. The data in the table have been obtained either from the model weights and captions files provided by the original authors or from our best implementation. Given its large use as a benchmark in the field, we consider the domain-generic COCO dataset also for this analysis. In the table, methods are clustered based on the information included in the visual encoding and ordered by CIDEr score. 
It can be observed that standard and embedding-based metrics all had a substantial improvement with the introduction of region-based visual encodings. Further improvement was due to the integration of information on inter-objects relations, either expressed via graphs or self-attention. Notably, CIDEr, SPICE, and Coverage most reflect the benefit of vision-and-language pre-training. Moreover, as expected, it emerges that the diversity-based scores are correlated, especially Div-1 and Div-2 and the Vocab Size. The correlation of this family of scores and the others is almost linear, except for early approaches, which perform averagely well in terms of Diversity despite lower values for standard metrics. From the trend of learning-based scores, it emerges that exploiting models trained on textual data only (BERT-S, reported in the table as its F1-score variant) does not help discriminating among image captioning approaches. On the other hand, considering as reference only the visual information and disregarding the ground-truth captions is possible with the appropriate vision-and-language pre-trained model (consider that CLIP-S and CLIP-S$^\text{Ref}$ are linearly correlated). This is a desirable property for an image captioning evaluation score since it allows estimating the performance of a model without relying on reference captions that can be limited in number and somehow subjective.

For readability, in Fig.~\ref{fig:cider_vs_all} we highlight the relation between the CIDEr score and other characteristics from Table~\ref{tab:results}. 
We chose CIDEr as this score is commonly regarded as one of the most relevant indicators of image captioning systems performance. The first plot, depicting the relation between model complexity and performance, shows that more complex models do not necessarily bring to better performance. The other plots describe an almost-linear relation between CIDEr and the other scores, with some flattening for high CIDEr values. These trends confirm the suitability of the CIDEr score as an indicator of the overall performance of an image captioning algorithm, whose specific characteristics in terms of the produced captions would still be expressed more precisely in terms of non-standard metrics.

\noindent We refer the reader to Appendix~\ref{sec:appendix_performance} for additional performance analyses and qualitative results.

%% file: 07-variants.tex
\section{Image Captioning Variants}
Beyond general-purpose image captioning, several specific sub-tasks have been explored in the literature. These can be classified into four categories according to their scope: 1. \emph{dealing with the lack of training data}; 2. \emph{focusing on the visual input}; 3. \emph{focusing on the textual output}; 4. \emph{addressing user requirements}.

\subsection{Dealing with the lack of training data}
Paired image-caption datasets are very expensive to obtain. Thus, some image captioning variants are being explored that limit the need for full supervision information.
\tit{Novel Object Captioning}
Novel object captioning focuses on describing objects not appearing in the training set, thus enabling a zero-shot learning setting that can increase the applicability of the models in the real world. Early approaches to this task~\cite{hendricks16cvpr,venugopalan17cvpr} tried to transfer knowledge from out-domain images by conditioning the model on external unpaired visual and textual data at training time. To explore this strategy, Hendricks~\etal~\cite{hendricks16cvpr} introduced a variant of the COCO dataset~\cite{lin2014microsoft}, called \emph{held-out COCO}, in which image-caption pairs containing one of eight pre-selected object classes were removed from the training set but not from the test set. To further encourage research on this task, the more challenging \emph{nocaps} dataset, with nearly 400 novel objects, has been introduced~\cite{agrawal2019nocaps}. 
Some approaches to this variant~\cite{yao2017incorporating,li2019lstmp} integrate copying mechanisms in the language model to select novel objects predicted from a tagger or generate a caption template with placeholders to be filled with novel objects~\cite{wu2018dnoc,lu2018neural}. 
On a different line, Anderson~\etal~\cite{cbs2017emnlp} devised the Constrained Beam Search algorithm to force the inclusion of selected tag words in the output caption, following the predictions of a tagger. Moreover, following the pre-training trend with BERT-like architectures, Hu~\etal~\cite{hu2020vivo} proposed a multi-layer Transformer model pre-trained by randomly masking one or more tags from image-tag pairs.
\tit{Unpaired Image Captioning}
Unpaired Image Captioning approaches can be either unsupervised or semi-supervised. Unsupervised captioning aims at understanding and describing images without paired image-text training data. Following unpaired machine translation approaches, the early work~\cite{gu2018unpaired} proposes to generate captions in a pivot language and then translate predicted captions to the target language. After this work, the most common approach focuses on adversarial learning by training an LSTM-based discriminator to distinguish whether a caption is real or generated~\cite{feng2019unsupervised,laina2019towards}. As alternative approaches, it is worth mentioning~\cite{gu2019unpaired} that generates a caption from the image scene-graph and~\cite{guo2020recurrent} that leverages a memory-based network.
Moreover, semi-supervised approaches have been proposed, such as~\cite{kim2019image}, which uses both paired and unpaired data with adversarial learning, and~\cite{ben2021unpaired}, which performs iterative self-learning.
\tit{Continual Captioning}
Continual captioning aims to deal with partially unavailable data by following the continual learning paradigm to incrementally learn new tasks without forgetting what has been learned before. In this respect, new tasks can be represented as sequences of captioning tasks with different vocabularies, as proposed in~\cite{del2020ratt}, and the model should be able to transfer visual concepts from one to the other while enlarging its vocabulary. 

\subsection{Focusing on the visual input}
Some sub-tasks focus on making the textual description more correlated with visual data.
\tit{Dense Captioning}
Dense captioning was proposed by Johnson~\etal~\cite{johnson2016densecap} and consists of concurrently localizing and describing salient image regions with short natural language sentences. In this respect, the task can be conceived as a generalization of object detection, where caption replaces object tags, or image captioning, where single regions replace the full image. 
To address this task, contextual and global features~\cite{yang2017dense,li2019learning} and attribute generators~\cite{yin2019context,kim2019dense} can be exploited. Related to this variant, an important line of works~\cite{krause2017hierarchical,liang2017recurrent,mao2018show,chatterjee2018diverse,zha2019context,luo2019curiosity} focuses on the generation of textual paragraphs that densely describe the visual content as a coherent story.
\tit{Text-based Image Captioning}
Text-based image captioning, also known as OCR-based image captioning or image captioning with reading comprehension, aims at reading and including the text appearing in images in the generated descriptions. The task was introduced by Sidorov~\etal~\cite{sidorov2020textcaps} with the TextCaps dataset. Another dataset designed for pre-training for this variant is \emph{OCR-CC}~\cite{yang2021tap}, which is a subset of images containing meaningful text taken from the CC3M dataset~\cite{sharma2018conceptual} and automatically annotated through a commercial OCR system. The common approach to this variant entails combining image regions and text tokens, \ie~groups of characters from an OCR, possibly enriched with mutual spatial information~\cite{wang2020multimodal,wang2021improving}, in the visual encoding~\cite{sidorov2020textcaps,zhu2021simple}. Another direction entails generating multiple captions describing different parts of the image, including the contained text~\cite{xu2021towards}.
\tit{Change Captioning}
Change captioning targets changes that occurred in a scene, thus requiring both accurate change detection and effective natural language description. The task was first presented in~\cite{jhamtani2018learning} with the \emph{Spot-the-Diff} dataset, composed of pairs of frames extracted from video surveillance footages and the corresponding textual descriptions of visual changes. To further explore this variant, the \emph{CLEVR-Change} dataset~\cite{park2019robust} has been introduced, which contains five scene change types on almost 80K image pairs. The proposed approaches for this variant apply attention mechanisms to focus on semantically relevant aspects without being deceived by distractors such as viewpoint changes~\cite{shi2020finding,huang2021image,kim2021viewpoint} or perform multi-task learning with image retrieval as an auxiliary task~\cite{hosseinzadeh2021image}, where an image must be retrieved from its paired image and the description of the occurred changes.

\subsection{Focusing on the textual output}
Since every image captures a wide variety of entities with complex interactions, human descriptions tend to be diverse and grounded to different objects and details. Some image captioning variants explicitly focus on these aspects.
\tit{Diverse Captioning} 
Diverse image captioning tries to replicate the quality and variability of the sentences produced by humans. The most common technique to achieve diversity is based on variants of the beam search algorithm~\cite{vijayakumar2018diverse} that entail dividing the beams into similar groups and encouraging diversity between groups. Other solutions have been investigated, such as contrastive learning~\cite{dai2017contrastive}, conditional GANs~\cite{dai2017towards,shetty2017speaking}, and paraphrasing~\cite{liu2019generating}. However, these solutions tend to underperform in terms of caption quality, which is partially recovered by using variational auto-encoders~\cite{wang2017diverse,aneja2019sequential,chen2019variational,mahajan2020diverse}. Another approach is exploiting multiple part-of-speech tags sequences predicted from image region classes~\cite{deshpande2019fast} and forcing the model to produce different captions based on these sequences.
\tit{Multilingual Captioning}
Since image captioning is commonly performed in English, multilingual captioning~\cite{elliott2015multilingual} aims to extend the applicability of captioning systems to other languages. The two main strategies entail collecting captions in different languages for commonly used datasets (\eg~Chinese and Japanese captions for COCO images~\cite{li2019coco,miyazaki2016cross}, German captions for Flick30K~\cite{elliott2016multi30k}), or directly training multilingual captioning systems with unpaired captions~\cite{elliott2015multilingual,lan2017fluency,gu2018unpaired,song2019unpaired}. 
\tit{Application-specific Captioning}
Image captioning can be applied to ease and automate activities involving text generation from images. For example, captioning systems can be applied for medical report generation, for which they need to predict disease tags and try to imitate the style of real medical reports~\cite{jing2018automatic,liu2021exploring,yang2021writing}.
Another interesting application is art description generation, which entails describing not only factual aspects of the artworks, but also their context and style, and conveyed message
art description~\cite{bai2021explain}. To this end, captioning systems could also rely on external knowledge, \eg~metadata. A similar application is automatic caption generation for news articles~\cite{ramisa2017breakingnews,biten2019good}, for which named entities from the article should be described~\cite{feng2012automatic,tran2020transform}, and the rich journalistic style should be maintained~\cite{liu2021visual,yang2021journalistic}. Another important application domain is assistive technology for the visually impaired~\cite{wu2017automatic}, where image captioning approaches must be able to provide informative descriptions even for low-quality visual inputs~\cite{gurari2020captioning}.

\subsection{Addressing user requirements}
Regular image captioning models generate factual captions with a neutral tone and no interaction with end-users. Instead, some image captioning sub-tasks are devoted to coping with user requests.
\tit{Personalized Captioning}
Humans consider more effective the captions that avoid stating the obvious and that are written in a style that catches their interest. Personalized image captioning aims at fulfilling this requirement by generating descriptions that take into account the user's prior knowledge, active vocabulary, and writing style. To this end, early approaches exploit a memory block as a repository for this contextual information~\cite{chunseong2017attend,park2018towards}. On another line, Zhang~\etal~\cite{zhang2020learning} proposed a multi-modal Transformer network that personalizes captions conditioned on the user's recent captions and a learned user representation. Other works have instead focused on the style of captions as an additional controllable input and proposed to solve this task by exploiting unpaired stylized textual corpus~\cite{gan2017stylenet,mathews2018semstyle,guo2019mscap,zhao2020memcap}. Some datasets have been collected to explore this variant, such as \emph{InstaPIC}~\cite{chunseong2017attend}, which is composed of multiple Instagram posts from the same users, \emph{FlickrStyle10K}~\cite{gan2017stylenet}, which contains images and textual sentences with two different styles, and \emph{Personality-Captions}~\cite{shuster2019engaging}, which contains triples of images, captions, and one among 215 personality traits to be used to condition the caption generation.
\tit{Controllable Captioning}
Controllable captioning puts the users in the loop by asking them to select and give priorities to what should be described in an image. This information is exploited as a guiding signal for the generation process. 
The signal can be sparse, as selected image regions~\cite{zheng2019intention,cornia2019show} and user-provided visual words~\cite{deshpande2019fast}, or dense, as mouse traces~\cite{pont2020connecting,meng2021connecting}. Eventually, the guiding signal can incorporate some form of structure, such as sequences that encode the mentioning order of concepts (part-of-speech tag as in~\cite{deshpande2019fast}) or visual objects~\cite{cornia2019show}. Guiding inputs can also encode the relation between objects that is most of interest for the user, as done for example in~\cite{chen2021human} via verbs and semantic roles (verbs represent activities in the image and semantic roles determine how objects engage in these activities) and in \cite{chen2020say,zhong2020comprehensive} via user-generated or user-selected scene graphs.
A different control signal is introduced by~\cite{deng2020length}, which consist of a length-level embedding added as an additional token to each textual word, providing existing models the ability to generate length-controllable image captions.
\tit{Image Captioning Editing}
Image captioning editing was proposed by Sammani~\etal~\cite{sammani2020show}, following the consideration that generated captions may have repetitions and inconsistencies. This variant focuses on decoupling the decoding stage in a caption generation step and a caption polishing one to correct syntactic errors.

%% file: 08-open_issues.tex
\section{Conclusions and Future Directions}
Image captioning is an intrinsically complex challenge for machine intelligence as it integrates difficulties from both Computer Vision and NLP. Further, as mentioned in the Introduction, the task itself is vaguely defined and captions can, in principle, be generated with many different styles and objectives. The presented literature review and experimental comparison show the performance improvement over the last few years on standard datasets. However, many open challenges remain since accuracy, robustness, and generalization results are far from satisfactory. Similarly, requirements of fidelity, naturalness, and diversity are not yet met. Based on the analysis presented, we can trace three main developmental directions for the image captioning field, which are discussed in the following.

\tit{Procedural and architectural challenges}
Since image captioning models are data greedy, pre-training on large-scale datasets, even if not well-curated, is becoming a solid strategy, as demonstrated in~\cite{li2020oscar,zhou2020unified,zhang2021vinvl,cornia2021universal}. In this regard, promoting the public release of such datasets will be fundamental to fostering reproducibility and allowing fair comparisons. The growing size of pre-training models is also a concern, and the community will probably need to investigate less computationally-intensive alternatives to promote equality in the community. In architectural terms, instead, the growing dichotomy between early-fusion strategies and the encoder-decoder paradigm is still to be solved and is currently one of the main open issues. On the other side, the supremacy of detection features is leaving space to a variety of visual encoding strategies (pre-training from scratch, using detections, using features from multi-modal models) which all appear to be on pair in terms of performance.

\tit{Generalization, diversity, long-tail concepts}
While pre-training on web-scale datasets provides a promising direction to increase generalization and promoting long-tail concepts~\cite{cornia2021universal}, specializing in particular domains and generating captions with different styles and aims is still among the main open challenges for image captioning. Although we discussed some attempts to encourage naturalness and diversity~\cite{dai2017towards,dai2017contrastive,wang2017diverse}, further research is needed to design models that are suitable for real-world applications. In this sense, the emergence of models which can deal with long-tail concepts~\cite{cornia2021universal,hu2021scaling}
offers a valuable promise of modeling real-life scenarios and generalizing to different contexts. Additionally, developments in image captioning variants such as novel objects captioning or controllable captioning could help to tackle this open issue. Notably, the emergence of subword-based tokenization techniques has made it possible to handle and generate rare words.

\tit{Design of trustworthy AI solutions}
Due to its potential in human-machine interaction, image captioning needs solutions that are transparent and acceptable for end-users, framed as overcome bias, and interpretable. Since most vision-and-language datasets share common patterns and regularities, datasets bias and overrepresented visual concepts are major issues for any vision-and-language task. In this sense, some effort should be devoted to the study of fairness and bias: two possible directions entail designing specific evaluation metrics and focusing on the robustness to unwanted correlations. Further, despite the promising performance on the benchmark datasets, state-of-the-art approaches are not yet satisfactory when applied in the wild. A possible reason for this is the evaluation procedures used and their impact on the training approaches currently adopted. In this sense, the design of appropriate and reproducible evaluation protocols~\cite{hodosh2016focused,xie2019going,alikhani2020cross} and insightful metrics remains an open challenge in image captioning. Moreover, since the task is currently defined as a supervised one and thus is strongly influenced by the training data, the development of scores that do not need reference captions for assessing the performance would be key for a shift towards unsupervised image captioning. Finally, since existing image captioning algorithms lack reliable and interpretable means for determining the cause of a particular output, further research is needed to shed more light on model explainability, focusing on how these deal with different modalities or novel concepts.

%% file: biographies.tex
\begin{IEEEbiography}[{\includegraphics[width=1in,height=1.25in,clip,keepaspectratio]{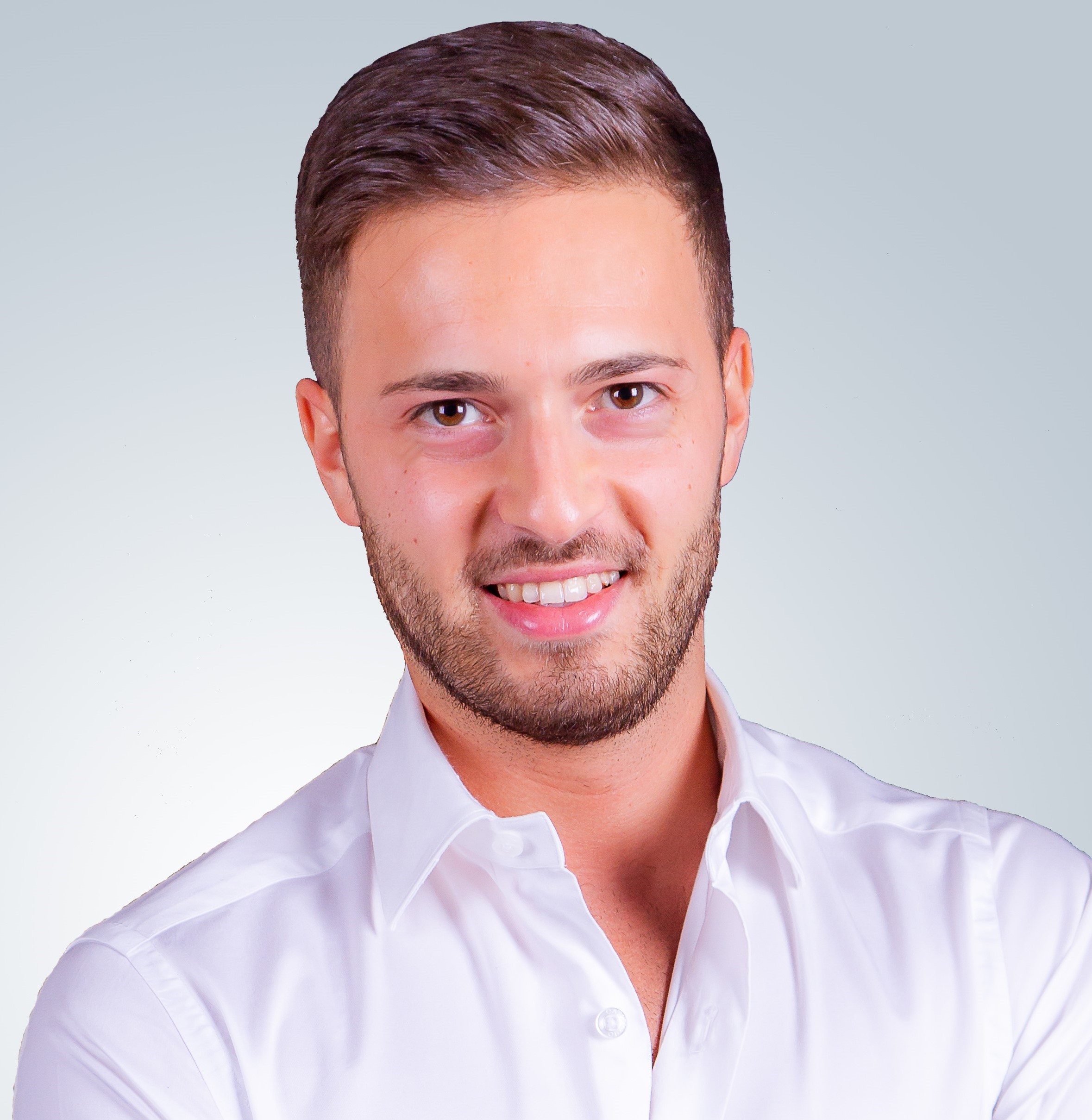}}]{Matteo Stefanini} received the M.Sc. degree in Computer Engineering cum laude from the University of Modena and Reggio Emilia, in 2018. He is currently pursuing a PhD degree in Information and Communication Technologies at the Department of Engineering ``Enzo Ferrari'', University of Modena and Reggio Emilia. 
His research activities involve the integration of vision and language modalities, focusing on image captioning and Transformer-based architectures.
\end{IEEEbiography}

\begin{IEEEbiography}[{\includegraphics[width=1in,height=1.25in,clip,keepaspectratio]{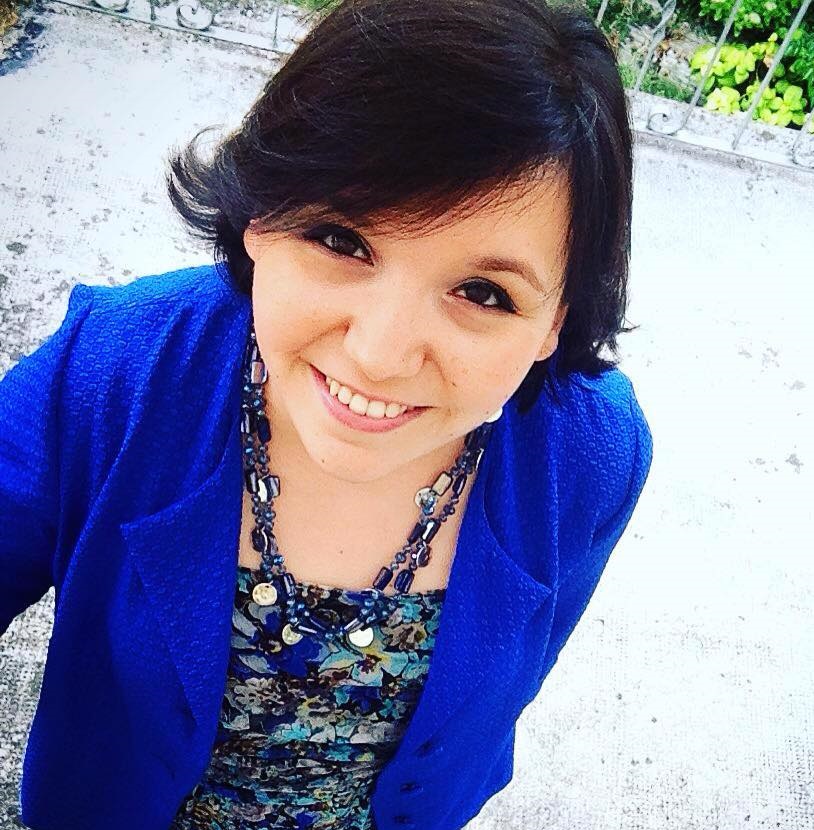}}]{Marcella Cornia} received the M.Sc. degree in Computer Engineering and the Ph.D. degree cum laude in Information and Communication Technologies from the University of Modena and Reggio Emilia, in 2016 and 2020, respectively. She is currently a Postdoctoral Researcher with the University of Modena and Reggio Emilia. She has authored or coauthored more than 30 publications in scientific journals and international conference proceedings.
\end{IEEEbiography}

\begin{IEEEbiography}[{\includegraphics[width=1in,height=1.25in,clip,keepaspectratio]{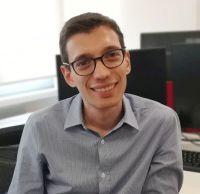}}]{Lorenzo Baraldi} received the M.Sc. degree in Computer Engineering and the Ph.D. degree cum laude in Information and Communication Technologies from the University of Modena and Reggio Emilia, in 2014 and 2018. He is currently Tenure Track Assistant Professor with the University of Modena and Reggio Emilia. He was a Research Intern at Facebook AI Research (FAIR) in 2017. He has authored or coauthored more than 70 publications in scientific journals and international conference proceedings. 
\end{IEEEbiography}

\begin{IEEEbiography}[{\includegraphics[width=1in,height=1.25in,clip,keepaspectratio]{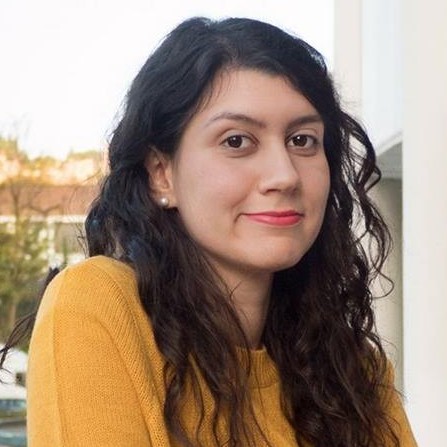}}]{Silvia Cascianelli} received the M.Sc. degree in Information and Automation Engineering and the Ph.D. degree cum laude in Information and Industrial Engineering from the University of Perugia, in 2015 and 2019, respectively. She is an Assistant Professor with the University of Modena and Reggio Emilia. She was a Visitor Researcher at the Queen Mary University of London in 2018. She has authored or coauthored more than 30 publications in scientific journals and international conference proceedings. 
\end{IEEEbiography}

\begin{IEEEbiography}[{\includegraphics[width=1in,height=1.25in,clip,keepaspectratio]{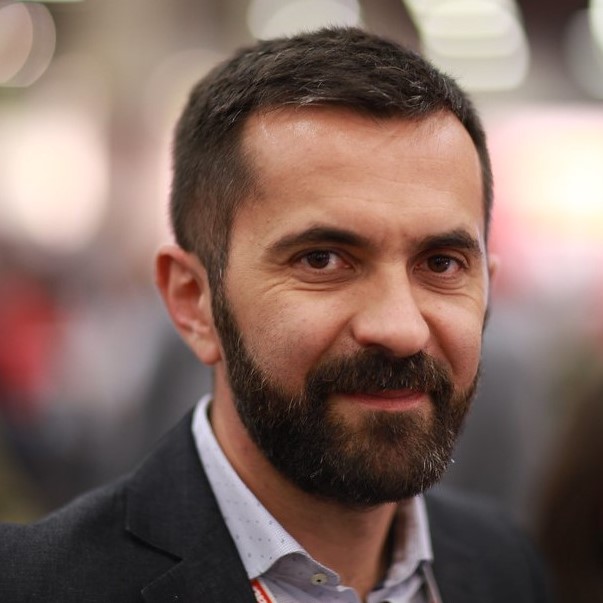}}]{Giuseppe Fiameni} is a Data Scientist at NVIDIA where he oversees the NVIDIA AI Technology Centre in Italy, a collaboration among NVIDIA, CINI and CINECA to accelerate academic research in the field of AI. He has been working as HPC specialist at CINECA, the largest HPC facility in Italy, for more than 14 years providing support for large-scale data analytics workloads. Research interests include large scale deep learning models, system architectures, massive data engineering, video action detection.
\end{IEEEbiography}

\begin{IEEEbiography}[{\includegraphics[width=1in,height=1.25in,clip,keepaspectratio]{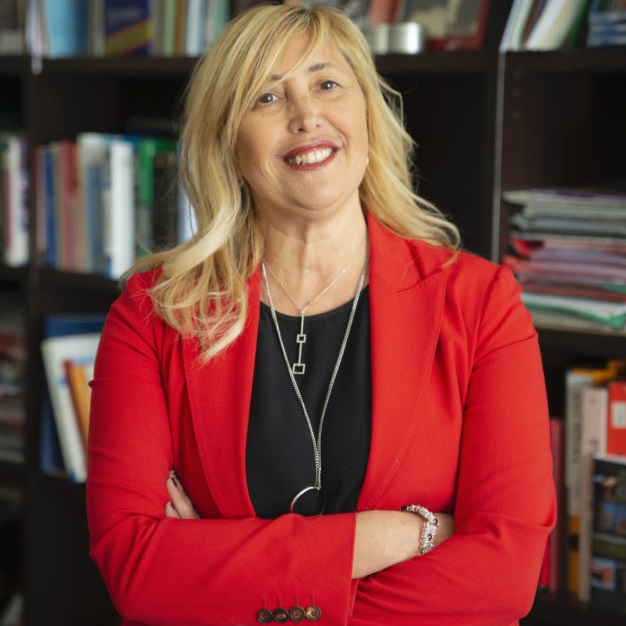}}]{Rita Cucchiara} received the M.Sc. degree in Electronics Engineering and the Ph.D. degree in Computer Engineering from the University of Bologna, in 1989 and 1992. She is currently Full Professor with the University of Modena and Reggio Emilia, where she heads the AImageLab Laboratory. She has authored or coauthored more than 400 papers in journals and international proceedings, and has been a coordinator of several projects in computer vision. She is Member of the Advisory Board of the Computer Vision Foundation, and Director of the ELLIS Unit of Modena.
\end{IEEEbiography}

%% file: supplementary_01-metrics.tex
\section{Further analysis of the evaluation metrics}
\label{sec:appendix_metrics}
In this section, we extend our analysis of the evaluation metrics for image captioning. In particular, in Table~\ref{tab:metrics}, we provide a taxonomy and summarize the main characteristics of the metrics presented. Moreover, in the following, we describe in more detail additional diversity metrics, embedding-based metrics, and learning-based metrics, which were only mentioned in the main paper.

\tit{Diversity metrics}
Local Diversity can be quantified via the average BLEU score between each caption and the others (\textbf{mBLEU}: the lower the mBLEU, the more diverse the produced caption set is)~\cite{shetty2017speaking}.
Another approach to quantify captions diversity is to focus on the mention of all the relevant words in the produced caption, despite their rareness in the training set. In this respect, in \cite{van2018measuring}, two recall-based diversity metrics have been proposed, namely \textbf{Global Recall} and \textbf{Local Recall}. The former is computed as the fraction of generated words with respect to the words appearing in both the training and validation set. The latter is computed for each test image as the fraction of generated words with respect to the words in the reference captions. 
Moreover, when the system can produce multiple captions for the same image, diversity can be quantified at the topics level by using the Latent Semantic Analysis-based metric \textbf{LSA} and the kernelized version of the CIDEr score \textbf{Self-CIDEr}, proposed in~\cite{wang2019describing,wang2020diversity}.

\tit{Embedding-based metrics}
Among the specific aspects of a produced caption that can e evaluated via embedding-based metrics, the hallucination rate is measured via the \textbf{CHAIR} score~\cite{rohrbach2018object}, which is expressed as the fraction of hallucinated objects among those mentioned in a caption (in the CHAIR$_\text{i}$ variant) or the fraction of captions with at least a hallucinated object among all the produced captions (in the CHAIR$_\text{s}$ variant). Other aspects that can be measured by exploiting embedding-based representation of candidate caption, reference captions, and image are \textbf{Relevance} (via cosine similarity), \textbf{Extraness}, and \textbf{Omission} (via orthogonal projections), as proposed in~\cite{jiang2019reo}. Moreover, to take into account the uniqueness of the generated captions, in~\cite{wang2020towards} the SPICE-U score is proposed as the harmonic mean of the SPICE score and a measure of the caption uniqueness, which considers the fraction of images in the training set not containing the mentioned concepts.

\tit{Learning-based metrics}
With respect to learning-based evaluation scores, \textbf{NNEval}~\cite{sharif2018nneval} was proposed as the first learning-based image captioning evaluation approach. It considers classical metrics (BLEU, METEOR, CIDEr, SPICE, and WMD) as features describing the candidate caption when compared to reference captions and predicts its probability of being human-generated.
Another early-proposed learning-based evaluation strategy is \textbf{LEIC}~\cite{cui2018learning}, which directly scores the probability of a caption being human-generated, conditioned on the image and eventually on a reference caption, by using a binary classifier fed with pre-trained ResNet image features and an LSTM-based encoding of the caption. 
As a refinement of BERT-S specific for image captioning, the \textbf{ViLBERT-S}~\cite{lee2020vilbertscore} exploits the image-conditioned embedding obtained from the vision-and-language representation model ViLBERT~\cite{lu2019vilbert}. Similar to the BERT-S, the matching between these tokens is expressed via the cosine similarity of their embeddings, and the best matching token pairs are used for computing precision, recall, and F1-score. 
Another variant of the BERT-S, to which we here refer to as \textbf{BERT-S$^\text{IRV}$}~\cite{yi2020improving}, takes into account the variability of the reference captions associated with the same image by combining them in a unique embedding vector that contains all the mentioned concepts and against which the candidate caption is compared.
To evaluate the candidate caption fidelity and adequacy, the \textbf{FAIEr}~\cite{wang2021faier} score exploits scene graphs matching. The references and the image scene graphs are fused in a unique scene graph, whose more relevant nodes (representing the concepts more often mentioned in the references) get more weight, and the score is obtained based on the similarity between the candidate scene graph and the unique scene graph.
On a similar line, the \textbf{UMIC} score~\cite{lee2021umic} exploits the pre-trained vision-and-language representation model UNITER~\cite{chen2020uniter}, fine-tuned on carefully-designed negative samples, to score a candidate caption without the need for reference captions.

\definecolor{lightgray}{gray}{0.97}
\begin{table}[t]
\centering
\caption{Taxonomy and main characteristics of image captioning metrics.}
\label{tab:metrics}
\rowcolors{4}{lightgray}{}
\setlength{\tabcolsep}{.4em}
\resizebox{\linewidth}{!}{
\begin{tabular}{lll c ccc}
\toprule
& & & & \multicolumn{3}{c}{\textbf{Inputs}} \\
\cmidrule{5-7}
& & & \textbf{Original Task} & \textbf{Pred} & \textbf{Refs} & \textbf{Image} \\
\midrule
\cellcolor{white}  & \cellcolor{white}  & BLEU~\cite{papineni2002bleu} & Translation & \cmark & \cmark & \\
\cellcolor{white}  & \cellcolor{white}  & METEOR~\cite{banerjee2005meteor} & Translation & \cmark & \cmark & \\
\cellcolor{white}  \textbf{Standard} & \cellcolor{white}  & ROUGE~\cite{lin2004rouge} & Summarization & \cmark & \cmark & \\
\cellcolor{white}  & \cellcolor{white}  &  CIDEr~\cite{vedantam2015cider} & Captioning & \cmark & \cmark & \\
\cellcolor{white}  & \cellcolor{white}  & SPICE~\cite{spice2016} & Captioning & \cmark & (\cmark) & (\cmark) \\
\midrule
\cellcolor{white}  & \cellcolor{white}  & Div~\cite{shetty2017speaking} & Captioning & \cmark & & \\
\cellcolor{white}  & \cellcolor{white}  & Vocab~\cite{shetty2017speaking} & Captioning & \cmark & & \\
\cellcolor{white}  & \cellcolor{white}  & \%Novel~\cite{shetty2017speaking} & Captioning & \cmark & & \\
\cellcolor{white}  \textbf{Diversity}   & \cellcolor{white}  & mBLEU~\cite{shetty2017speaking} & Captioning & \cmark & & \\
\cellcolor{white}  & \cellcolor{white}  & LSA~\cite{wang2019describing,wang2020diversity} & Captioning & \cmark & & \\
\cellcolor{white}  & \cellcolor{white}  & Self-CIDEr~\cite{wang2019describing,wang2020diversity} & Captioning & \cmark & & \\
\cellcolor{white}  & \cellcolor{white}  & Recall~\cite{van2018measuring} & Captioning & \cmark & (\cmark) & \\
\midrule
\cellcolor{white}& \cellcolor{white}  & WMD~\cite{kusner2015word} & Doc. Dissimilarity & \cmark & \cmark &  \\
\cellcolor{white}  & \cellcolor{white}  & Alignment~\cite{cornia2019show} & Captioning & \cmark & \cmark &  \\
\cellcolor{white} & \cellcolor{white}  & Coverage~\cite{cornia2019show,bigazzi2020explore} & Captioning & \cmark & (\cmark) & (\cmark) \\
\cellcolor{white} & \cellcolor{white}  & Relevance~\cite{jiang2019reo} & Captioning & \cmark & (\cmark) & \cmark \\
\cellcolor{white} & \cellcolor{white}  & Extraness~\cite{jiang2019reo} & Captioning & \cmark & (\cmark) & \cmark \\
\cellcolor{white} & \cellcolor{white}  & Omission~\cite{jiang2019reo} & Captioning & \cmark & (\cmark) & \cmark \\
\cellcolor{white} & \cellcolor{white}  & SPICE-U~\cite{wang2020towards} & Captioning & \cmark & (\cmark) & \cmark \\
\multirow{-8}{*}{\cellcolor{white}\textbf{Embedding-based}} & \cellcolor{white}  & CHAIR~\cite{rohrbach2018object} & Captioning & \cmark & \cmark & \cmark \\
\midrule
\cellcolor{white}  & \cellcolor{white}  & NNEval~\cite{sharif2018nneval} & Captioning & \cmark & \cmark &  \\
\cellcolor{white}  & \cellcolor{white}  & BERT-S~\cite{zhang2020bertscore} & Text Similarity & \cmark & \cmark & \\
\cellcolor{white}  & \cellcolor{white}  & BERT-S$^\text{IRV}$~\cite{yi2020improving} & Captioning & \cmark & \cmark & \\
\cellcolor{white}  & \cellcolor{white}  & UMIC~\cite{lee2021umic} & Captioning & \cmark &  & \cmark \\
\cellcolor{white}  \textbf{Learning-based} & \cellcolor{white}  & LEIC~\cite{cui2018learning} & Captioning & \cmark & (\cmark) & \cmark \\
\cellcolor{white}  & \cellcolor{white}  & CLIP-S~\cite{hessel2021clipscore} & Captioning & \cmark & (\cmark) & \cmark \\
\cellcolor{white}  & \cellcolor{white}  & TIGEr~\cite{jiang2019tiger} & Captioning & \cmark & \cmark & \cmark \\
\cellcolor{white}  & \cellcolor{white}  & ViLBERT-S~\cite{lee2020vilbertscore} & Captioning &  \cmark & \cmark & \cmark \\
\cellcolor{white}  & \cellcolor{white}  & FAIEr~\cite{wang2021faier} & Captioning & \cmark & \cmark & \cmark \\
\bottomrule
\end{tabular}
}
\vspace{-0.2cm}
\end{table}

%% file: supplementary_02-evaluation.tex
\section{Further Performance Analysis}
\label{sec:appendix_performance}

According to the taxonomies proposed in Sections~\ref{sec:visual_encoding},~\ref{sec:language_model}, and~\ref{sec:training_strategies}, in Table~\ref{tab:overview}, we overview the most relevant surveyed methods. We report their performance in terms of BLEU-4, METEOR, and CIDEr on the COCO Karpathy test set and their main features in terms of visual encoding, language modeling, and training strategies. In the table, methods are clustered based primarily on their visual encoding strategy and ordered based on the obtained scores. Methods exploiting vision-and-language pre-training are further separated from the others. Image captioning models have reached impressive performance in just a few years: from an average BLEU-4 of 25.1 for the methods using global CNN features to an average BLEU-4 of 35.3 and 40.0 for those exploiting the attention and self-attention mechanisms, peaking at 42.6 in the case of vision-and-language pre-training.
By looking at the performance in terms of the CIDEr score, we can notice that, as for the visual encoding, the more complete and structured information about semantic visual concepts and their mutual relation is included, the better is the performance (consider that methods applying attention over a grid of features reach an average CIDEr score of 105.8, while those performing attention over visual regions 121.8, further increased for graph-based approaches and methods using self-attention, which reach 133.2 on average). As for the language model, LSTM-based approaches combined with strong visual encoders are still competitive with subsequent fully-attentive methods in terms of performance. These methods are slower to train but are generally smaller than Transformer-based ones. As for the training strategy, sentence-level fine-tuning with reinforcement learning leads to significant performance improvement (consider that methods relying only on the cross-entropy loss obtain an average CIDEr score of 92.3, while those combining it with reinforcement learning fine-tuning reach 125.1 on average). Moreover, it emerges that vision-and-language pre-training on large datasets allows boosting the performance and deserves further investigation (with an average CIDEr score of 140.4).

\begin{figure*}[t]
\centering
\includegraphics[width=\textwidth]{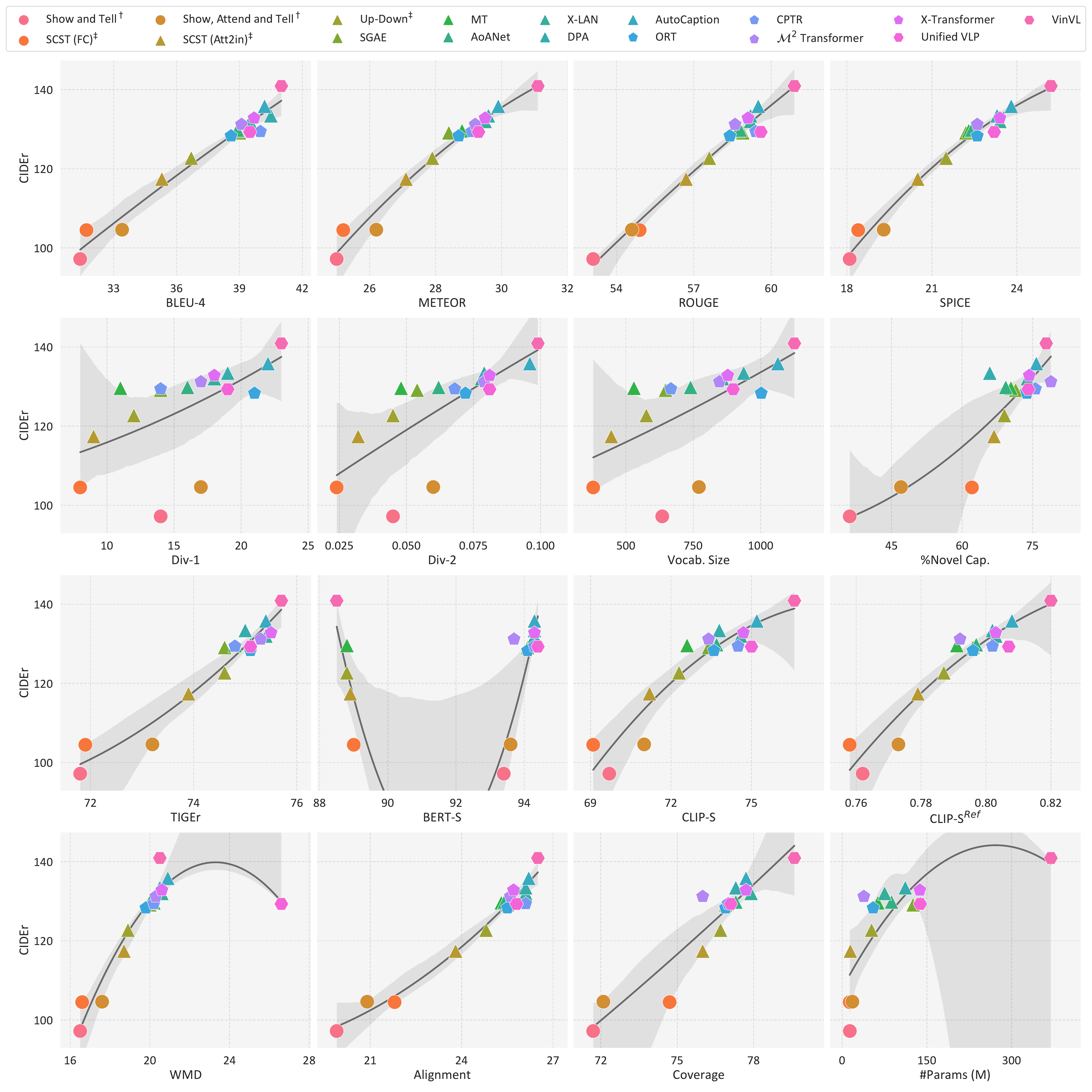}
\caption{Relationship between CIDEr, number of parameters and other scores. Values of Div-1, Div-2, Alignment, CLIP-S, and CLIP-S$^\text{Ref}$ are multiplied by powers of 10 for readability.} 
\label{fig:cider_vs_all_all}
\vspace{-0.2cm}
\end{figure*}

For completeness, in Fig.~\ref{fig:cider_vs_all_all} we report the relation between the CIDEr score and all the other characteristics from Table~\ref{tab:results}. The almost-linear relation with the CIDEr score is observable also for the scores not reported in Fig.~\ref{fig:cider_vs_all} in the main paper, with the only exception of the BERT-S score.

\definecolor{lightgray}{gray}{0.97}
\begin{table*}
\caption{Overview of deep learning-based image captioning models. Scores are taken from the respective papers. For all the metrics, the higher the value, the better ($\uparrow$).}
\label{tab:overview}
\setlength{\tabcolsep}{.3em}
\renewcommand{\arraystretch}{1.1}
\resizebox{\linewidth}{!}{
\rowcolors{7}{}{lightgray}
\begin{tabular}{lccccccccccccccccccccccc}
\toprule
& & & \multicolumn{5}{c}{\textbf{Visual Encoding}} & & & \multicolumn{3}{c}{\textbf{Language Model}} & & &  \multicolumn{4}{c}{\textbf{Training Strategies}} & & &  \multicolumn{3}{c}{\textbf{Main Results}} \\
\cmidrule{4-8} \cmidrule{11-13} \cmidrule{16-19} \cmidrule{22-24}
\textbf{Model} & & & Global & Grid & Regions & Graph & Self-Attention & & & RNN/LSTM & Transformer & BERT & & & XE & MLM & Reinforce & VL Pre-Training & & & BLEU-4 & METEOR & CIDEr\\
\midrule
LEMON~\cite{hu2021scaling} & & & & & \cmark & & \cmark & & & & & \cmark & & & & \cmark & \cmark & \cmark & & & \textbf{42.6} & 31.4 & \textbf{145.5} \\
UniversalCap~\cite{cornia2021universal} & & & & \cmark & & & \cmark & & & & \cmark & & & & \cmark & & \cmark & \cmark & & & 40.8 & 30.4 & 143.4 \\
SimVLM~\cite{wang2021simvlm} & & & & & & & \cmark & & & & \cmark & & & & \cmark & & & \cmark & & & 40.6 & \textbf{33.7} & 143.3 \\
VinVL~\cite{zhang2021vinvl} & & & & & \cmark & & \cmark & & & & & \cmark & & & & \cmark & \cmark & \cmark & & & 41.0 & 31.1 & 140.9 \\
Oscar~\cite{li2020oscar} & & & & & \cmark & & \cmark & & & & & \cmark & & & & \cmark & \cmark & \cmark & & & 41.7 & 30.6 & 140.0 \\
Unified VLP~\cite{zhou2020unified} & & & & & \cmark & & \cmark & & & & & \cmark & & & \cmark & & \cmark & \cmark & & & 39.5 & 29.3 & 129.3 \\
\midrule
AutoCaption~\cite{zhu2020autocaption} & & & & & \cmark & & \cmark & & & \cmark & & & & & \cmark & & \cmark & & & & 40.2 & 29.9 & 135.8 \\
RSTNet~\cite{zhang2021rstnet} & & & & \cmark & & & \cmark & & & & \cmark & & & & \cmark & & \cmark & & & & 40.1 & 29.8 & 135.6 \\
DLCT~\cite{luo2021dual}  & & & & \cmark & \cmark & & \cmark & & & & \cmark & & & & \cmark & & \cmark & & & & 39.8 & 29.5 & 133.8 \\
DPA~\cite{liu2020prophet} & & & & & \cmark & & \cmark & & & \cmark & & & & & \cmark & & \cmark & & & & 40.5 & 29.6 & 133.4 \\
X-Transformer~\cite{pan2020x} & & & & & \cmark & & \cmark & & & & \cmark & & & & \cmark & & \cmark & & & & 39.7 & 29.5 & 132.8 \\
NG-SAN~\cite{guo2020normalized} & & & & & \cmark & & \cmark & & & & \cmark & & & & \cmark & & \cmark & & & & 39.9 & 29.3 & 132.1 \\
X-LAN~\cite{pan2020x} & & & & & \cmark & & \cmark & & & \cmark & & & & & \cmark & & \cmark & & & & 39.5 & 29.5 & 132.0 \\
GET~\cite{Ji2020ImprovingIC} & & & & & \cmark & & \cmark & & & & \cmark & & & & \cmark & & \cmark & & & & 39.5 & 29.3 & 131.6 \\
$\mathcal{M}^2$ Transformer~\cite{cornia2020meshed} & & & & & \cmark & & \cmark & & & & \cmark & & & & \cmark & & \cmark & & & & 39.1 & 29.2 & 131.2 \\
AoANet~\cite{huang2019attention} & & & & & \cmark & & \cmark & & & \cmark & & & & & \cmark & & \cmark & & & & 38.9 & 29.2 & 129.8 \\
CPTR~\cite{liu2021cptr} & & & & & & & \cmark & & & & \cmark & & & & \cmark & & \cmark & & & & 40.0 & 29.1 & 129.4 \\
ORT~\cite{herdade2019image} & & & & & \cmark & & \cmark & & & & \cmark & & & & \cmark & & \cmark & & & & 38.6 & 28.7 & 128.3 \\
CNM~\cite{yang2019learning}  & & & & & \cmark & & \cmark & & & \cmark & & & & & \cmark & & \cmark & & & & 38.9 & 28.4 & 127.9 \\
ETA~\cite{li2019entangled} & & & & & \cmark & & \cmark & & & & \cmark & & & & \cmark & & \cmark & & & & 39.9 & 28.9 & 127.6 \\
\midrule
GCN-LSTM+HIP~\cite{yao2019hierarchy} & & & & & \cmark & \cmark & & & & \cmark & & & & & \cmark & & \cmark & & & & 39.1 & 28.9 & 130.6 \\
MT~\cite{shi2020improving} & & & & & \cmark & \cmark & & & & \cmark & & & & & \cmark & & \cmark & & & & 38.9 & 28.8 & 129.6 \\
SGAE~\cite{yang2019auto} & & & & & \cmark & \cmark & & & & \cmark & & & & & \cmark & & \cmark & & & & 39.0 & 28.4 & 129.1 \\
GCN-LSTM~\cite{yao2018exploring} & & & & & \cmark & \cmark & & & & \cmark & & & & & \cmark & & \cmark & & & & 38.3 & 28.6 & 128.7 \\
VSUA~\cite{guo2019aligning} & & & & & \cmark & \cmark & & & & \cmark & & & & & \cmark & & \cmark & & & & 38.4 & 28.5 & 128.6 \\
\midrule
SG-RWS~\cite{wang2020show} & & & & & \cmark & & & & & \cmark & & & & & \cmark & & \cmark & & & & 38.5 & 28.7 & 129.1 \\
LBPF~\cite{qin2019look} & & & & & \cmark & & & & & \cmark & & & & & \cmark & & \cmark & & & & 38.3 & 28.5 & 127.6 \\
AAT~\cite{huang2019adaptively} & & & & & \cmark & & & & & \cmark & & & & & \cmark & & \cmark & & & & 38.2 & 28.3 & 126.7  \\
CAVP~\cite{zha2019context}  & & & & & \cmark & & & & & \cmark & & & & & \cmark & & \cmark & & & & 38.6 & 28.3 & 126.3\\
Up-Down~\cite{anderson2018bottom} & & & & & \cmark & & & & & \cmark & & & & & \cmark & & \cmark & & & & 36.3 & 27.7 & 120.1 \\
RDN~\cite{ke2019reflective} & & & & & \cmark & & & & & \cmark & & & & & \cmark & & & & & & 36.8 & 27.2&  115.3 \\
Neural Baby Talk~\cite{lu2018neural} & & & & & \cmark & & & & & \cmark & & & & & \cmark & & & & & & 34.7 & 27.1 & 107.2 \\
\midrule
Stack-Cap~\cite{gu2018stack} & & & & \cmark & & & & & & \cmark & & & & & \cmark & & \cmark & & & & 36.1 & 27.4 & 120.4 \\
MaBi-LSTM~\cite{ge2019exploring} & & & & \cmark & & & & & & \cmark & & & & & \cmark & & & & & & 36.8 & 28.1 & 116.6 \\
RFNet~\cite{jiang2018recurrent} & & & & \cmark & & & & & & \cmark & & & & & \cmark & & \cmark & & & & 35.8 & 27.4 & 112.5 \\
SCST (Att2in)~\cite{rennie2017self} & & & & \cmark & & & & & & \cmark & & & & & \cmark & & \cmark & & & & 33.3 & 26.3 & 111.4 \\  
Adaptive Attention~\cite{lu2017knowing} & & & & \cmark & & & & & & \cmark & & & & & \cmark & & & & & & 33.2 & 26.6 & 108.5 \\  
Skeleton~\cite{wang2017skeleton} & & & & \cmark & & & & & & \cmark & & & & & \cmark & & & & & & 33.6 & 26.8 & 107.3 \\
ARNet~\cite{chen2018regularizing} & & & & \cmark & & & & & & \cmark & & & & & \cmark & & & & & & 33.5 & 26.1 & 103.4 \\
SCA-CNN~\cite{chen2017sca} & & & & \cmark & & & & & & \cmark & & & & & \cmark & & & & & & 31.1 & 25.0 & 95.2 \\
Areas of Attention~\cite{pedersoli2017areas} & & & & \cmark & & & & & & \cmark & & & & & \cmark & & & & & & 30.7 & 24.5 & 93.8 \\
Review Net~\cite{yang2016review} & & & & \cmark & & & & & & \cmark & & & & & \cmark & & & & & & 29.0 & 23.7 & 88.6 \\
Show, Attend and Tell~\cite{xu2015show} & & & & \cmark & & & & & & \cmark & & & & & \cmark & & & & & & 24.3 & 23.9 & - \\
\midrule
SCST (FC)~\cite{rennie2017self} & & & \cmark & & & & & & & \cmark & & & & & \cmark & & \cmark & & & & 31.9 & 25.5 & 106.3 \\ 
PG-SPIDEr~\cite{liu2017improved} & & & \cmark & & & & & & & \cmark & & & & & \cmark & & \cmark & & & & 33.2 & 25.7 & 101.3 \\
SCN-LSTM~\cite{gan2017semantic} & & & \cmark & & & & & & & \cmark & & & & & \cmark & & & & & & 33.0 & 25.7 & 101.2 \\ 
LSTM-A~\cite{yao2017boosting} & & & \cmark & & & & & & & \cmark & & & & & \cmark & & & & & & 32.6 & 25.4 & 100.2 \\
CNN$_\mathcal{L}$+RNH~\cite{gu2017empirical} & & & \cmark & & & & & & & \cmark & & & & & \cmark & & & & & & 30.6 & 25.2 & 98.9 \\
Att-CNN+LSTM~\cite{wu2016value} & & & \cmark & & & & & & & \cmark & & & & & \cmark & & & & & & 31.0 & 26.0 & 94.0 \\
GroupCap~\cite{chen2018groupcap} & & & \cmark & & & & & & & \cmark & & & & & \cmark & & & & & & 33.0 & 26.0 & - \\
StructCap~\cite{chen2017structcap} & & & \cmark & & & & & & & \cmark & & & & & \cmark & & & & & & 32.9 & 25.4 & - \\
Embedding Reward~\cite{ren2017deep} & & & \cmark & & & & & & & \cmark & & & & & \cmark & & \cmark & & & & 30.4 & 25.1 & 93.7 \\
ATT-FCN~\cite{you2016image} & & & \cmark & & & & & & & \cmark & & & & & \cmark & & & & & & 30.4 & 24.3 & - \\
MIXER~\cite{ranzato2015sequence} & & & \cmark & & & & & & & \cmark & & & & & \cmark & & \cmark & & & & 29.0 & - & - \\
MSR~\cite{fang2015captions} & & & \cmark & & & & & & & \cmark & & & & & \cmark & & & & & & 25.7 & 23.6 & - \\
gLSTM~\cite{jia2015guiding} & & & \cmark & & & & & & & \cmark & & & & & \cmark & & & & & & 26.4 & 22.7 & 81.3 \\
m-RNN~\cite{mao2015deep} & & & \cmark & & & & & & & \cmark & & & & & \cmark & & & & & & 25.0 & - & - \\
Show and Tell~\cite{vinyals2015show}  & & & \cmark & & & & & & & \cmark & & & & & \cmark & & & & & & 24.6 & - & - \\
Mind’s Eye~\cite{chen2015mind}  & & & \cmark & & & & & & & \cmark & & & & & \cmark & & & & & & 19.0 & 20.4 & -\\
DeepVS~\cite{karpathy2015deep}  & & & \cmark & & & & & & & \cmark & & & & & \cmark & & & & & & 23.0 & 19.5 & 66.0 \\
LRCN~\cite{donahue2015long}  & & & \cmark & & & & & & & \cmark & & & & & \cmark & & & & & & 21.0 & - & - \\
\bottomrule
\end{tabular}
}
\end{table*}

We deepen our analysis by considering also the Flickr30K dataset, and evaluate the performance of methods trained on COCO, and tested on the test set of Flickr30K, and of methods developed for COCO, but trained and tested on Flickr30K. The results of this study are reported in Table~\ref{tab:results_f30k}. Compared to what is obtained by the same models when trained and tested on COCO (in Table~\ref{tab:results}), the standard metrics and the embedding-based metrics significantly drop while diversity metrics increase. This can be imputed to the smaller size of Flickr30K compared to COCO. Learning-based metrics, especially BERT-S and CLIP-S, are more stable. As expected, training on COCO and directly testing on Flickr30K results in a performance drop under all the metrics when compared with the case in which both training and test are performed on data from Flickr30K. This confirms that generalization is still an open issue for image captioning approaches. Interestingly, VinVL surpasses the other considered approaches under all metrics in the case of the test on Flickr30K, suggesting the benefits of pre-training in terms of generalization capabilities of the resulting model.

\begin{figure*}[t]
\centering
\resizebox{\linewidth}{!}{
\begin{tabular}{c}
\includegraphics[width=\textwidth]{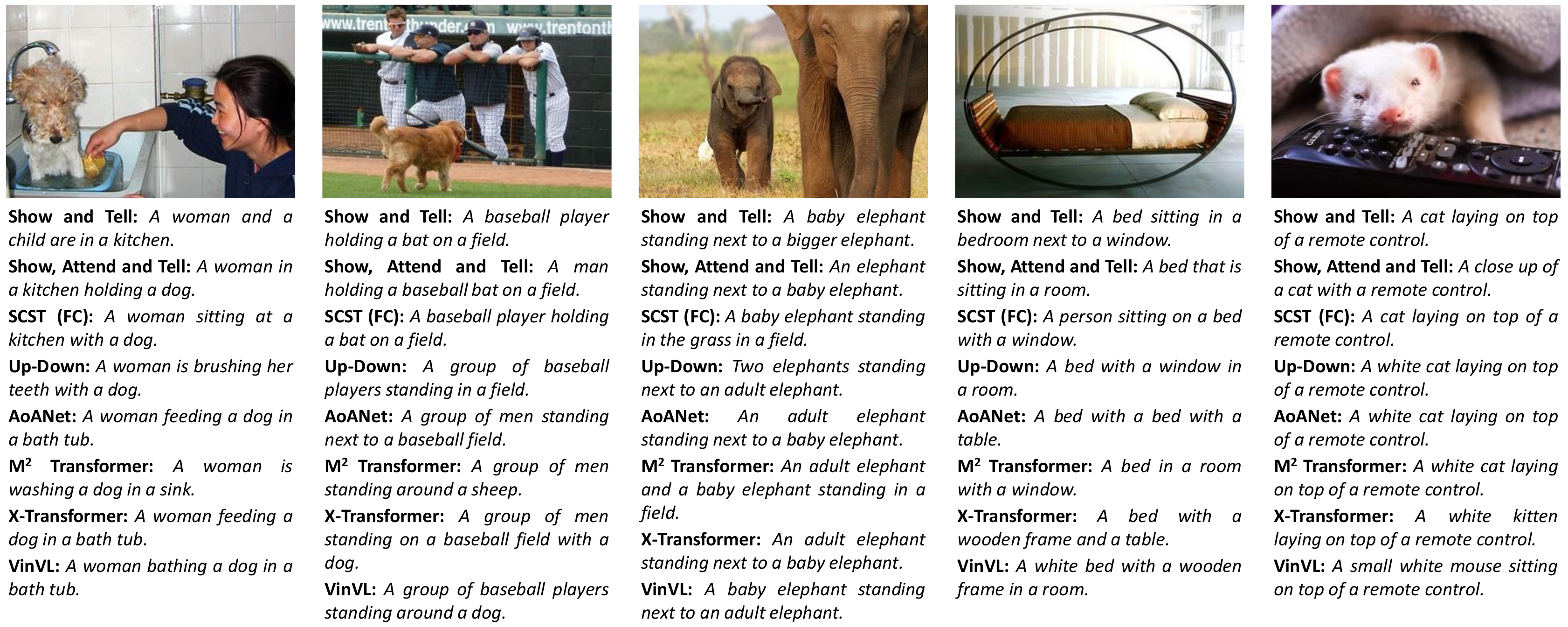} \\
\addlinespace[0.08cm]
\includegraphics[width=\textwidth]{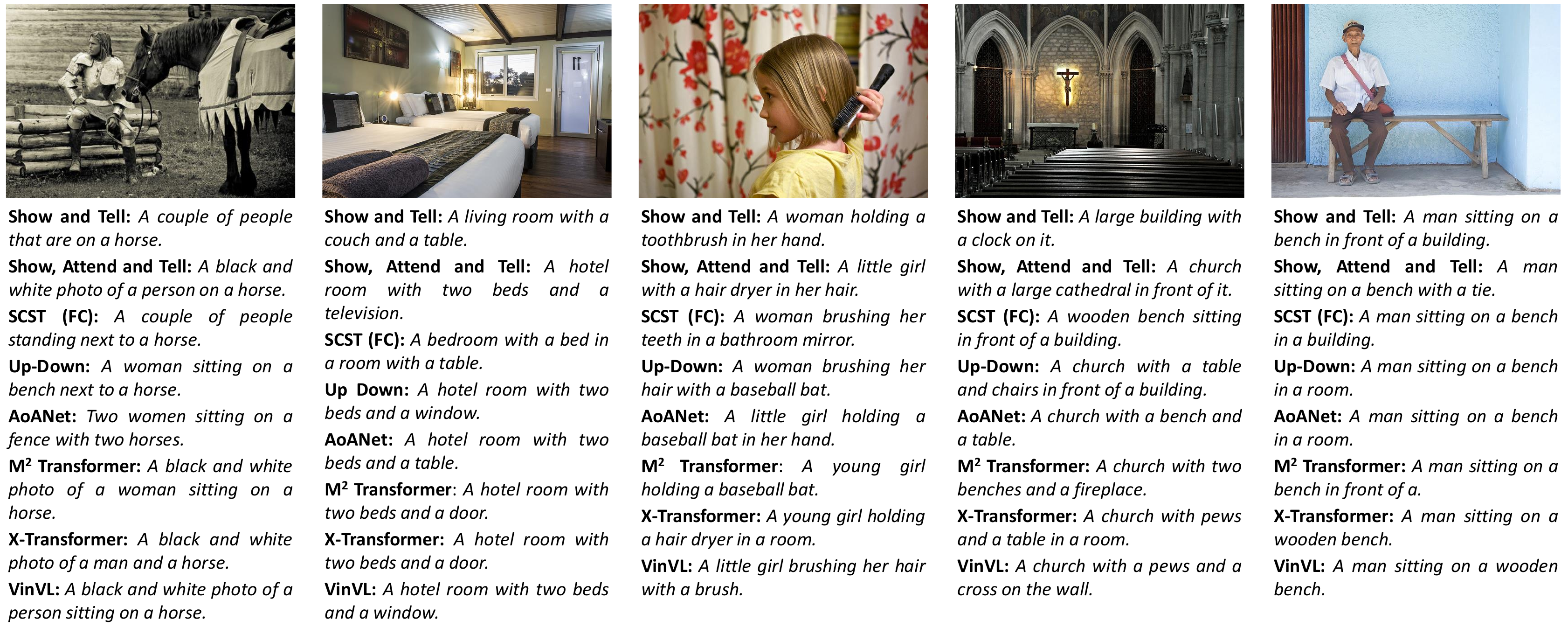} \\
\addlinespace[0.08cm]
\includegraphics[width=\textwidth]{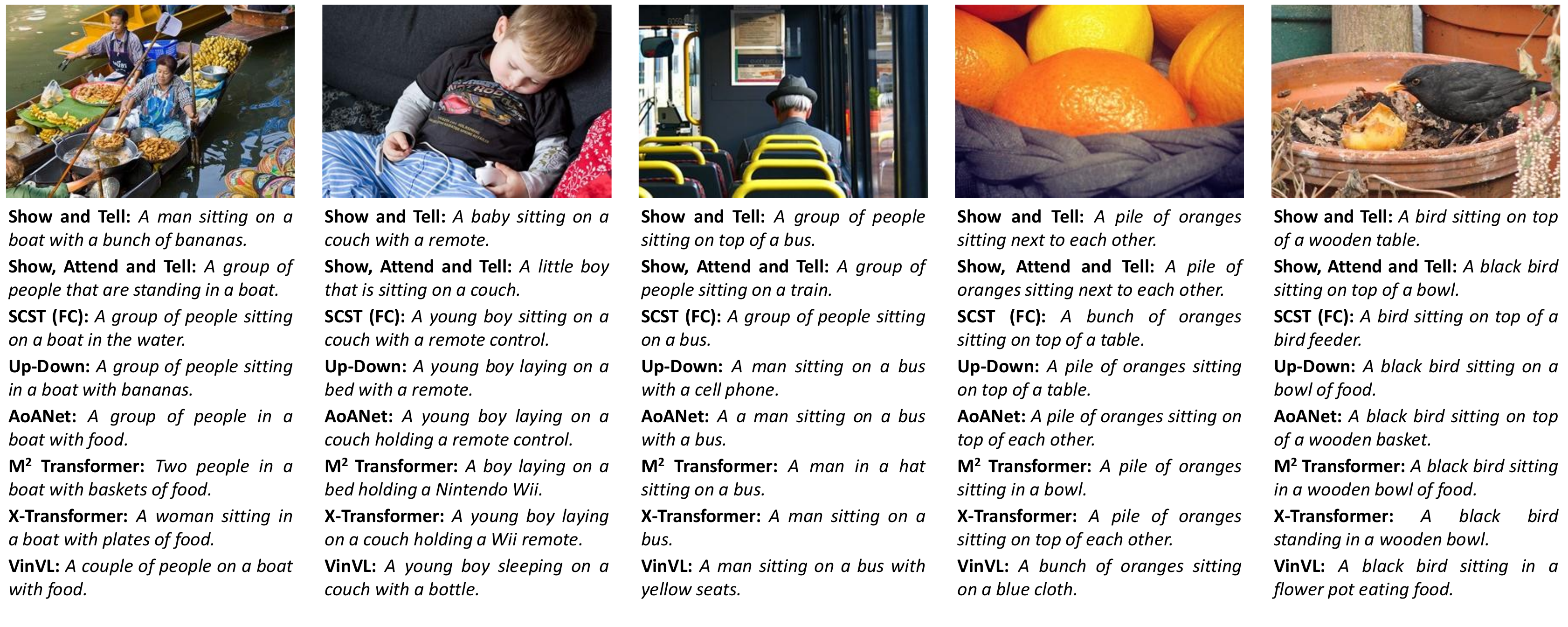} \\
\end{tabular}
}
\vspace{-0.3cm}
\caption{Additional qualitative examples from eight popular captioning models on COCO test images.}
\label{fig:qualitatives_01}
\end{figure*}

\begin{figure*}[t]
\centering
\resizebox{\linewidth}{!}{
\begin{tabular}{c}
\includegraphics[width=\textwidth]{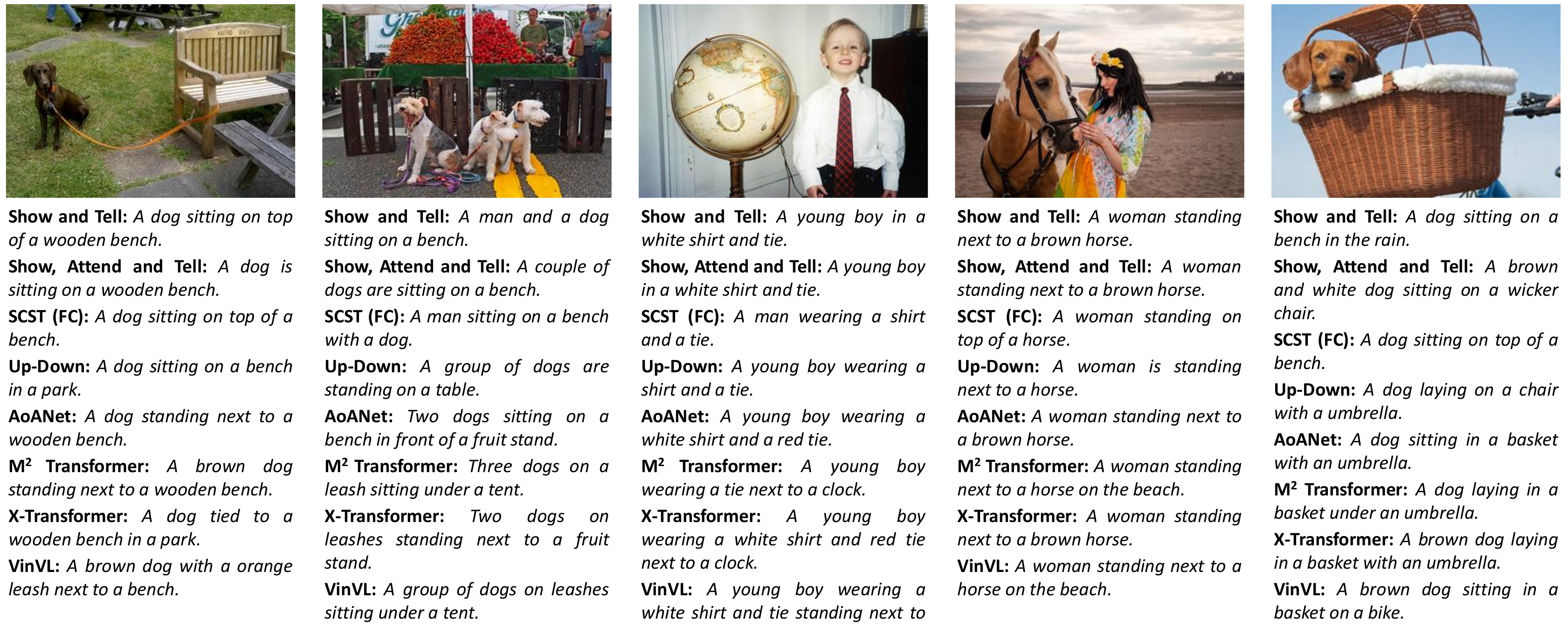} \\
\addlinespace[0.08cm]
\includegraphics[width=\textwidth]{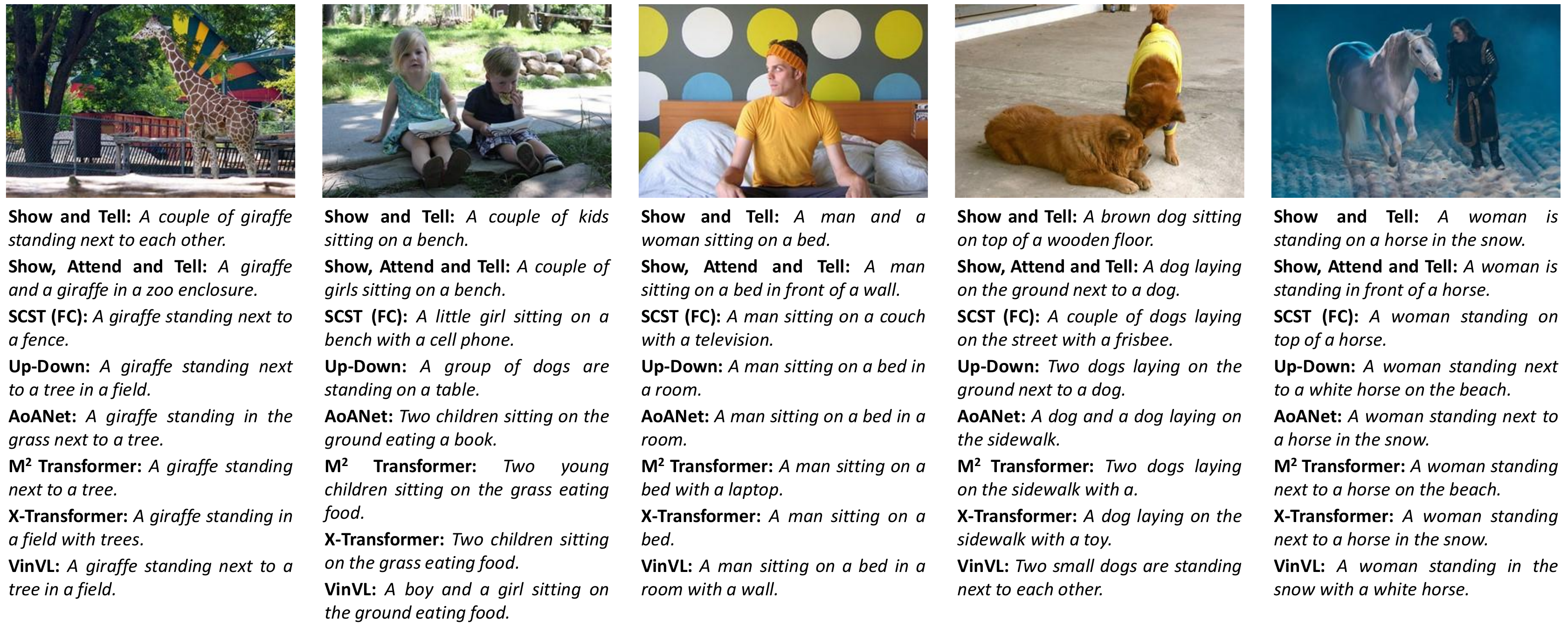} \\
\addlinespace[0.08cm]
\includegraphics[width=\textwidth]{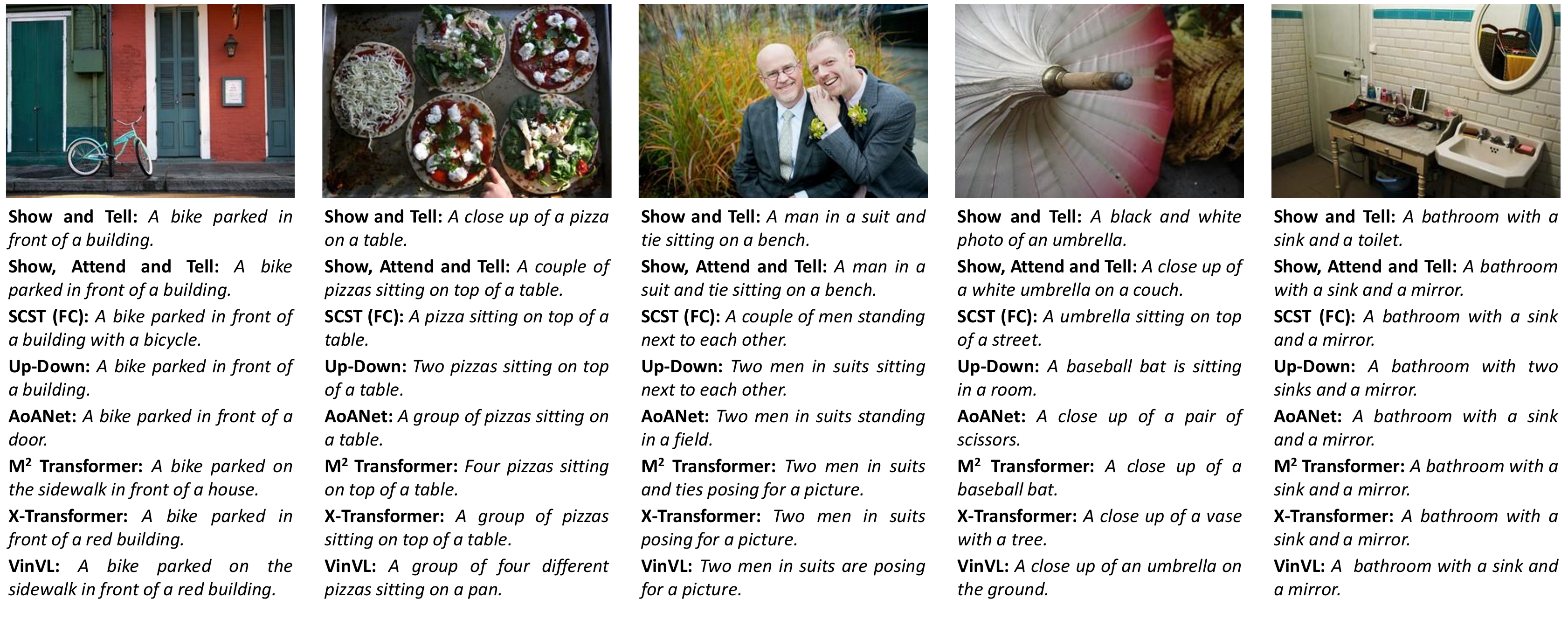} \\
\end{tabular}
}
\vspace{-0.3cm}
\caption{Additional qualitative examples from eight popular captioning models on COCO test images.}
\label{fig:qualitatives_02}
\end{figure*}

For a deeper qualitative analysis, in Fig.~\ref{fig:qualitatives_01}-\ref{fig:qualitatives_02}, we report examples of captions generated by eight popular approaches on images from the COCO dataset. It can be observed that the produced captions have similar length and structure (\ie~the main subject is mentioned first, then the main action, and finally additional details concerning other visual entities in the scene). This mimics the characteristics of the majority of the ground-truth captions in COCO. Another aspect that emerges is the lack of counting capabilities (consider \eg~the first example in the second row and the second example in the bottom row of Fig.~\ref{fig:qualitatives_02}). Current approaches struggle in mentioning the right amount of instances of the same entities and generally refer to multiple instances as \textit{a group of}. Finally, it is worth mentioning the difficulty in describing unusual concepts, both situations and visual entities (consider \eg~the first two examples in Fig.~\ref{fig:qualitatives_01}), which is a symptom of the lack of generalization capability. In fact, in such cases, unusual concepts are described as more represented concepts in the training set. For example, the \textit{ferret} in the top-right of Fig.~\ref{fig:qualitatives_01} is described as a \textit{cat} or a \textit{mouse}, and the historically-dressed \textit{man} in the last example of the second row of Fig.~\ref{fig:qualitatives_02} is described as a \textit{woman}. This issue is less evident for VinVL, enforcing the role of pre-training to achieve better generalization capabilities.

Finally, in Fig.~\ref{fig:attention1}, we report a visualization of the attention states corresponding to the captions generated by two methods based on image regions, \ie~Up-Down~\cite{anderson2018bottom}, which performs attention over image regions, and $\mathcal{M}^2$~Transformer~\cite{cornia2020meshed}, which performs self-attention encoding. For visualization, we use the normalized attribution scores obtained for each image region via the Integrated Gradients approach~\cite{sundararajan2017axiomatic} and projected between 0 and 1 by applying a contrast stretching function. In particular, we show the attended regions for each generated word and outline the region with the highest attribution score. With a focus on visual words, it can be observed that, for both approaches, the regions with the highest scores are coherent with the produced word. However, thanks to the more complex relations modeled via self-attention, $\mathcal{M}^2$ Transformer generally pays more attention to fewer, more precise regions compared to Up-Down (consider \eg~the region contributing the most to outputting \textit{tracks} in the third example, or \textit{skateboard} in the last one).

\begin{table*}
\caption{Performance analysis of representative image captioning approaches in terms of different evaluation metrics on the Flickr30K datatset. The $\dagger$ marker indicates models trained by us with ResNet-152 features, while the $\ddagger$ marker indicates unofficial implementations. For all the metrics, the higher the value, the better ($\uparrow$).}
\label{tab:results_f30k}
\setlength{\tabcolsep}{.3em}
\renewcommand{\arraystretch}{1.1}
\rowcolors{7}{}{lightgray}
\resizebox{\linewidth}{!}{
\begin{tabular}{>{\color{black}}l>{\color{black}}c>{\color{black}}c>{\color{black}}c>{\color{black}}c>{\color{black}}c>{\color{black}}c>{\color{black}}c>{\color{black}}c>{\color{black}}c>{\color{black}}c>{\color{black}}c>{\color{black}}c>{\color{black}}c>{\color{black}}c>{\color{black}}c>{\color{black}}c>{\color{black}}c>{\color{black}}c>{\color{black}}c>{\color{black}}c>{\color{black}}c>{\color{black}}c>{\color{black}}c>{\color{black}}c>{\color{black}}c>{\color{black}}c}
\toprule
& & & & \multicolumn{6}{c}{\textbf{Standard Metrics}} & & & \multicolumn{4}{c}{\textbf{Diversity Metrics}} & & & \multicolumn{3}{c}{\textbf{Embedding-based Metrics}} & & & \multicolumn{4}{c}{\textbf{Learning-based Metrics}}  \\
\cmidrule{5-10} \cmidrule{13-16} \cmidrule{19-21} \cmidrule{24-27}
& & \textbf{Trained on} & & \textbf{B-1} & \textbf{B-4} & \textbf{M} & \textbf{R} & \textbf{C} & \textbf{S} & & & \textbf{Div-1} & \textbf{Div-2} & \textbf{Vocab} & \textbf{\%Novel} & & & \textbf{WMD} & \textbf{Alignment} & \textbf{Coverage} & & & \textbf{TIGEr} & \textbf{BERT-S} & \textbf{CLIP-S} & \textbf{CLIP-S$^\text{Ref}$}\\
\midrule
Show and Tell$^\dagger$~\cite{vinyals2015show} & & COCO & & 51.0 & 11.4 & 13.1 & 34.8 & 22.8 & 7.6 & & & 0.037 & 0.093 & 331 & 94.8 & & & 8.6 & 0.019 & 61.9 & & & 52.9 & 90.6 & 0.604 & 0.656 \\
Show, Attend and Tell$^\dagger$~\cite{xu2015show} & & COCO & & 57.3 & 14.7 & 15.1 & 38.8 & 29.4 & 9.4 & & & 0.044 & 0.124 & 402 & 96.3 & & & 9.5 & 0.053 & 63.7 & & & 53.0 & 91.1 & 0.638 & 0.686 \\
Up-Down$^\ddagger$~\cite{anderson2018bottom} & & COCO & & 65.5 & 19.5 & 18.6 & 44.0 & 42.6 & 12.5 & & & 0.047 & 0.131 & 463 & 98.1 & & & 11.1 & 0.105 & 68.9 & & & 53.6 & 91.9 & 0.682 & 0.719 \\
$\mathcal{M}^2$ Transformer~\cite{cornia2020meshed} & & COCO & & 67.9 & 21.0 & 19.4 & 45.3 & 47.4 & 13.0 & & & 0.048 & 0.150 & 470 & 98.9 & & & 11.7 & 0.106 & 67.0 & & & 53.7 & 91.8 & 0.680 & 0.721 \\
VinVL~\cite{zhang2021vinvl} & & COCO & & \textbf{74.3} & \textbf{28.4} & \textbf{23.5} & \textbf{51.1} & \textbf{75.2} & \textbf{16.8} & & & \textbf{0.066} & \textbf{0.188} & \textbf{651} & \textbf{98.8} & & & \textbf{15.0} & \textbf{0.147} & \textbf{72.2} & & & \textbf{53.6} & \textbf{93.1} & \textbf{0.754} & \textbf{0.787} \\
\midrule
Show and Tell$^\dagger$~\cite{vinyals2015show} & & Flickr30K & & 64.1 & 21.5 & 18.3 & 44.4 & 41.7 & 12.2 & & & 0.037 & 0.075 & 373 & 84.5 & & & 11.2 & 0.090 & 64.2 & & & 53.5 & 92.1 & 0.658 & 0.701 \\
Show, Attend and Tell$^\dagger$~\cite{xu2015show} & & Flickr30K & & 65.6 & 23.6 & 19.2 & 45.4 & 49.1 & 13.3 & & & 0.045 & 0.096 & 454 & 90.1 & & & 11.8 & 0.089 & 64.1 & & & 53.4 & 92.1 & 0.679 & 0.717 \\
Up-Down$^\ddagger$~\cite{anderson2018bottom} & & Flickr30K & & \textbf{72.4} & 28.3 & 21.6 & 49.5 & 63.3 & 15.9 & & & 0.061 & 0.155 & 587 & 95.6 & & & 13.1 & 0.119 & 65.5 & & & 53.5 & 92.7 & 0.720 & 0.755 \\
ORT~\cite{herdade2019image} & & Flickr30K & & 72.2 & \textbf{30.1} & \textbf{22.8} & 50.4 & \textbf{68.8} & \textbf{16.9} & & & 0.072 & 0.171 &  \textbf{738} &  \textbf{96.1} & & &  \textbf{13.7} &  \textbf{0.129} &  \textbf{67.2} & & & 53.5 & 92.7 &  \textbf{0.728} & 0.760 \\
$\mathcal{M}^2$ Transformer~\cite{cornia2020meshed} & & Flickr30K & & \textbf{72.4} & 29.8 & 22.4 & \textbf{50.6} & 68.4 & 16.2 & & &  \textbf{0.079} &  \textbf{0.196} & 728 & 93.8 & & & 13.6 & 0.120 & 65.0 & & & \textbf{53.6} &  \textbf{92.9} & 0.724 &  \textbf{0.763} \\
\bottomrule
\end{tabular}
}
\vspace{-0.2cm}
\end{table*}

\begin{figure*}
\footnotesize
\centering
\setlength{\tabcolsep}{.05em}
\renewcommand*{\arraystretch}{0.5}
\resizebox{\linewidth}{!}{
\begin{tabular}{cccccccc}
\multicolumn{8}{c}{\textbf{Up-Down~\cite{anderson2018bottom}}} \\
\includegraphics[width=0.124\linewidth]{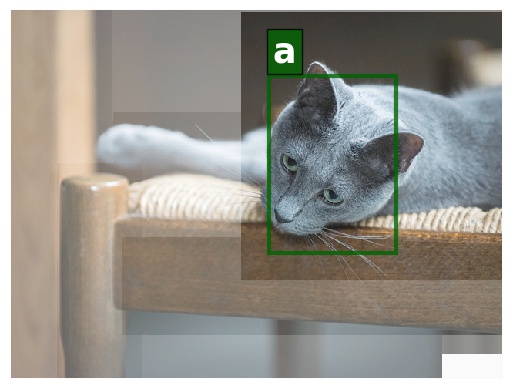} & 
\includegraphics[width=0.124\linewidth]{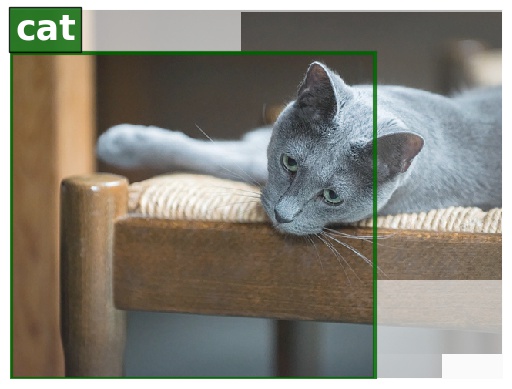} & 
\includegraphics[width=0.124\linewidth]{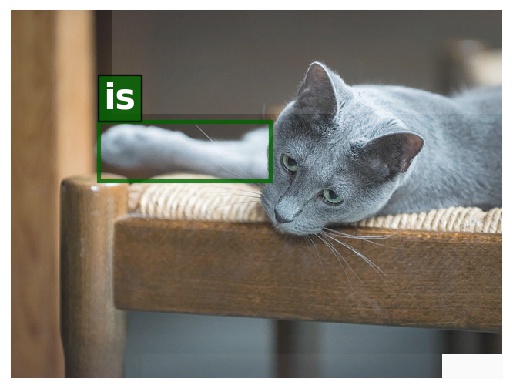} & 
\includegraphics[width=0.124\linewidth]{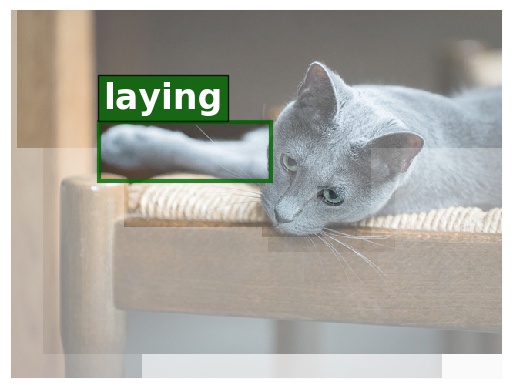} & 
\includegraphics[width=0.124\linewidth]{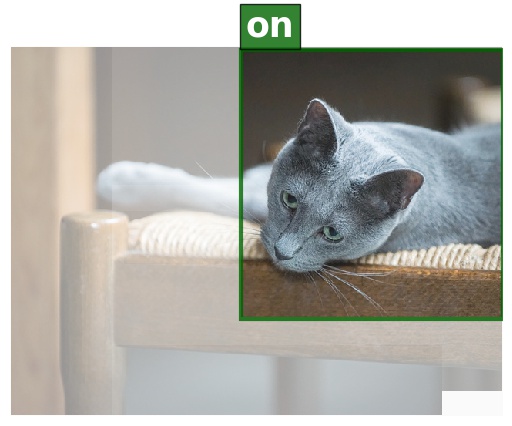} &
\includegraphics[width=0.124\linewidth]{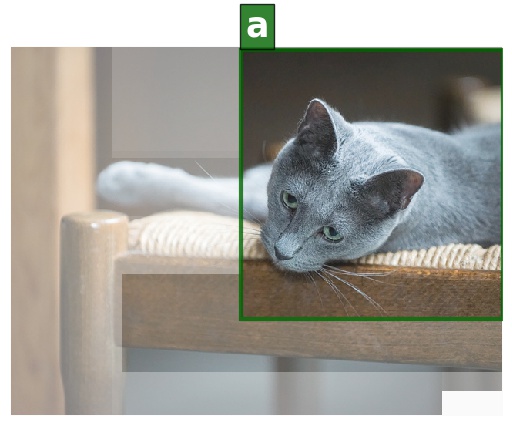} & 
\includegraphics[width=0.124\linewidth]{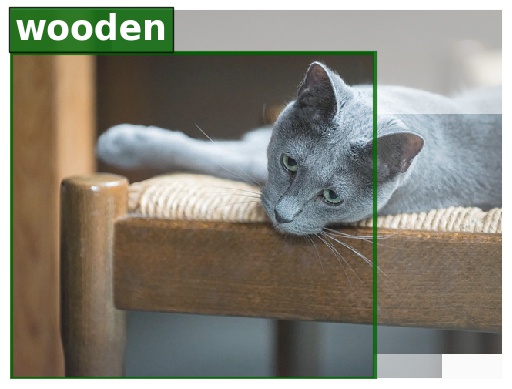} & 
\includegraphics[width=0.124\linewidth]{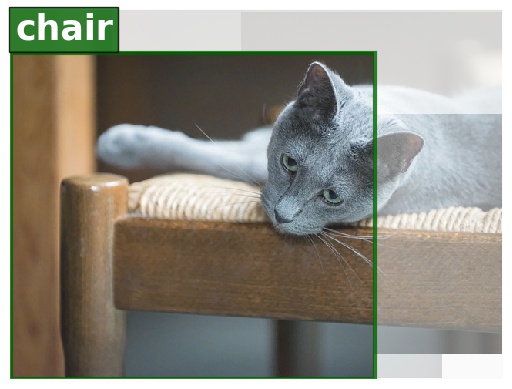} \\
\addlinespace[0.1cm]
\multicolumn{8}{c}{\textbf{$\mathbf{\mathcal{M}^2}$ Transformer~\cite{cornia2020meshed}}} \\
\includegraphics[width=0.124\linewidth]{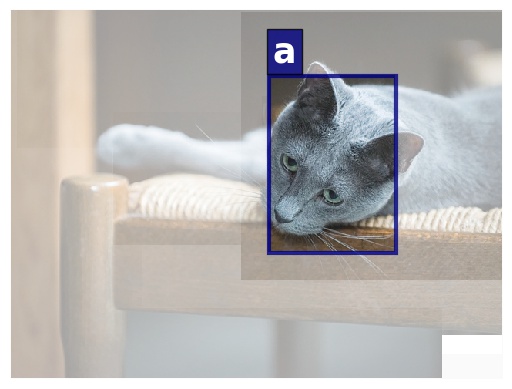} & 
\includegraphics[width=0.124\linewidth]{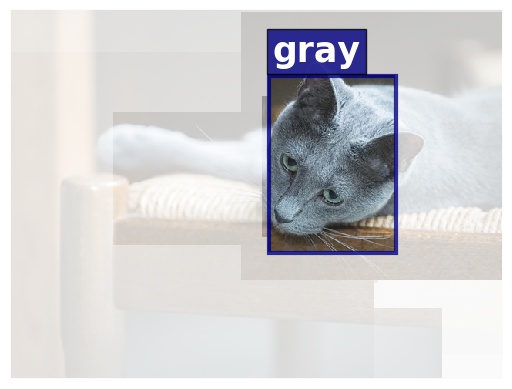} & 
\includegraphics[width=0.124\linewidth]{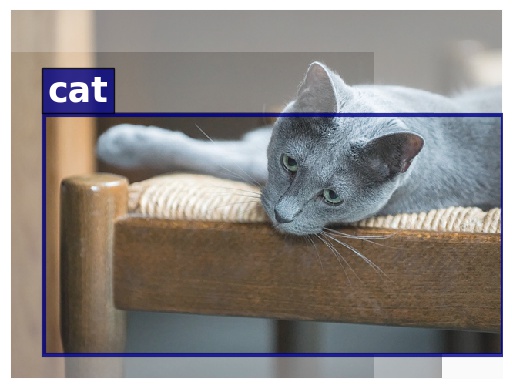} & 
\includegraphics[width=0.124\linewidth]{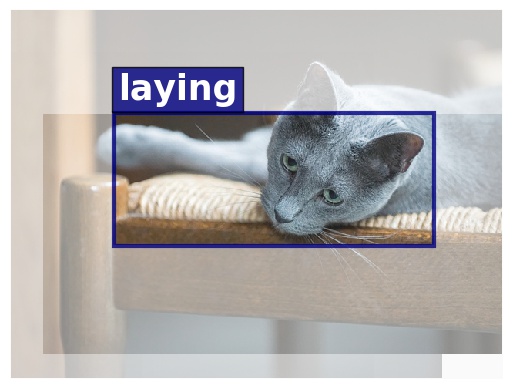} & 
\includegraphics[width=0.124\linewidth]{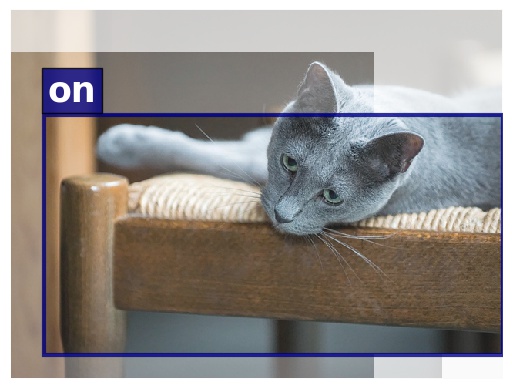} &
\includegraphics[width=0.124\linewidth]{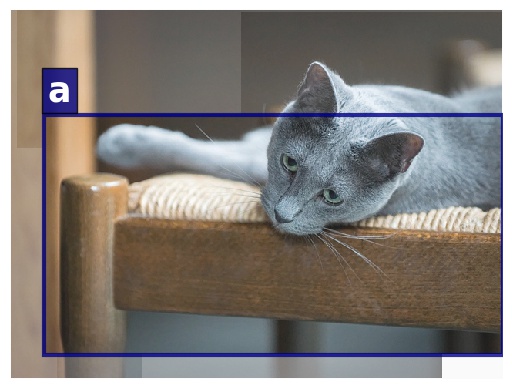} & 
\includegraphics[width=0.124\linewidth]{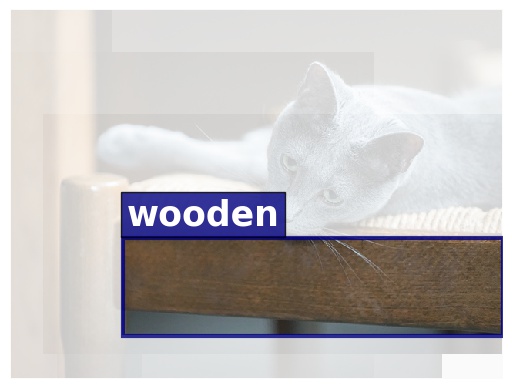} & 
\includegraphics[width=0.124\linewidth]{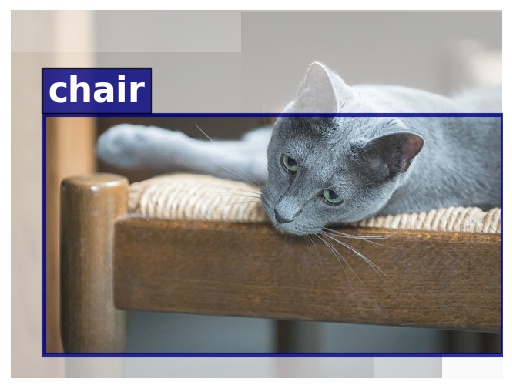} \\

\addlinespace[0.2cm]
\multicolumn{8}{c}{\textbf{Up-Down~\cite{anderson2018bottom}}} \\
\includegraphics[width=0.124\linewidth]{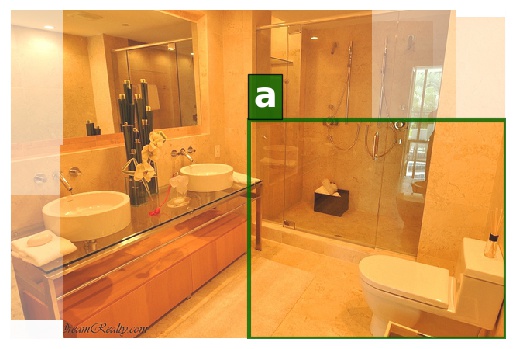} & 
\includegraphics[width=0.124\linewidth]{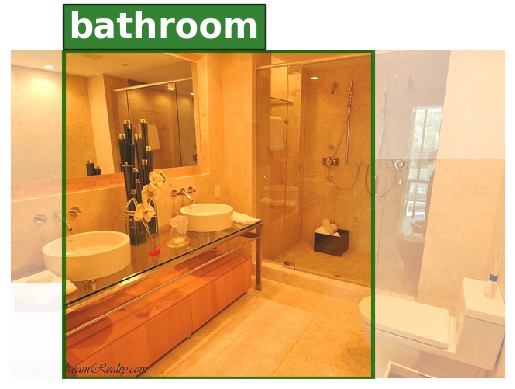} & 
\includegraphics[width=0.124\linewidth]{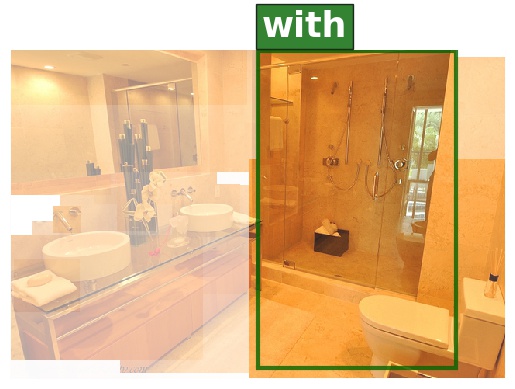} & 
\includegraphics[width=0.124\linewidth]{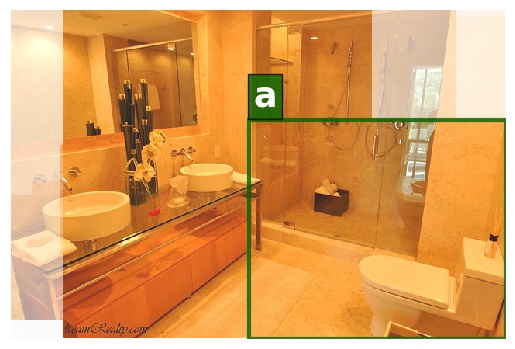} & 
\includegraphics[width=0.124\linewidth]{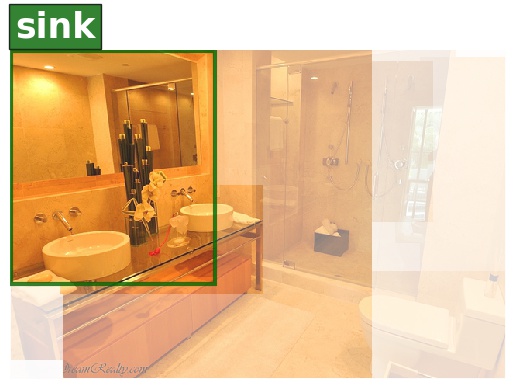} &
\includegraphics[width=0.124\linewidth]{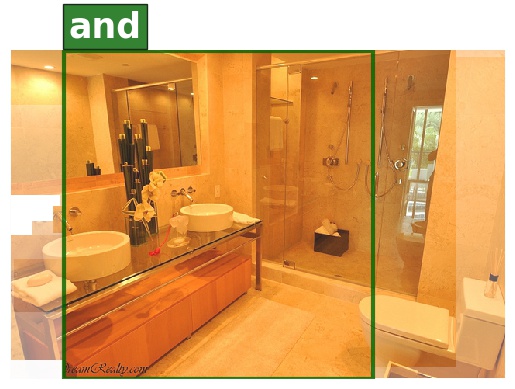} & 
\includegraphics[width=0.124\linewidth]{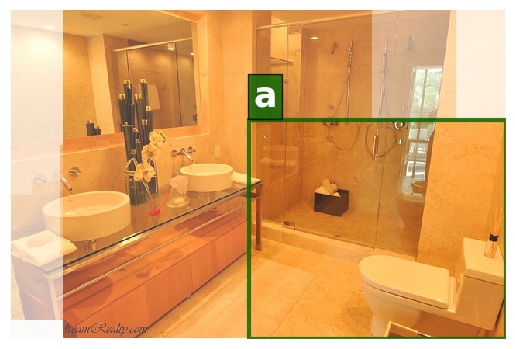} & 
\includegraphics[width=0.124\linewidth]{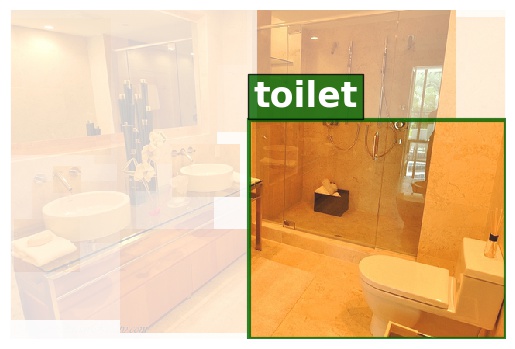} \\
\addlinespace[0.1cm]
\multicolumn{8}{c}{\textbf{$\mathbf{\mathcal{M}^2}$ Transformer~\cite{cornia2020meshed}}} \\
\includegraphics[width=0.124\linewidth]{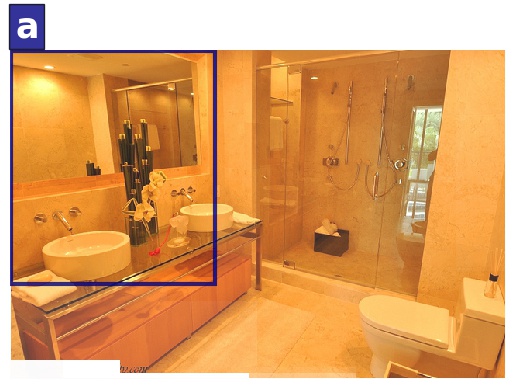} & 
\includegraphics[width=0.124\linewidth]{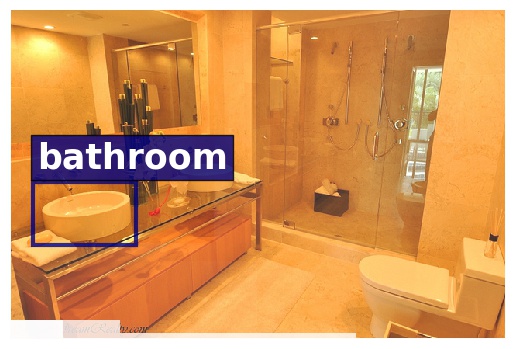} & 
\includegraphics[width=0.124\linewidth]{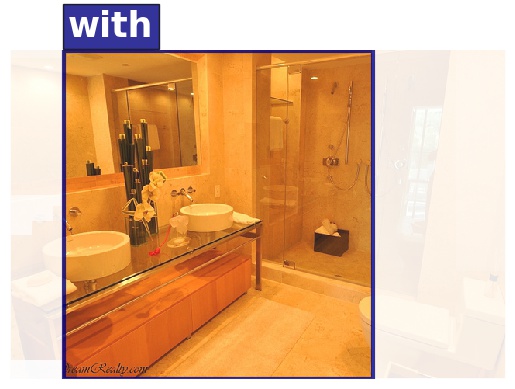} & 
\includegraphics[width=0.124\linewidth]{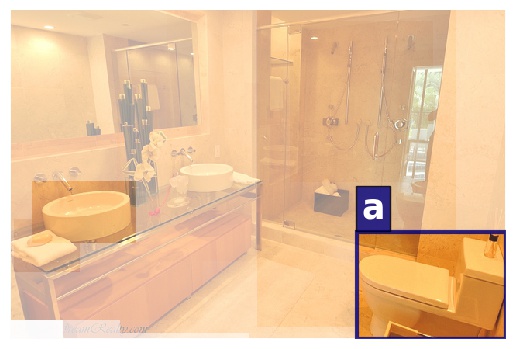} & 
\includegraphics[width=0.124\linewidth]{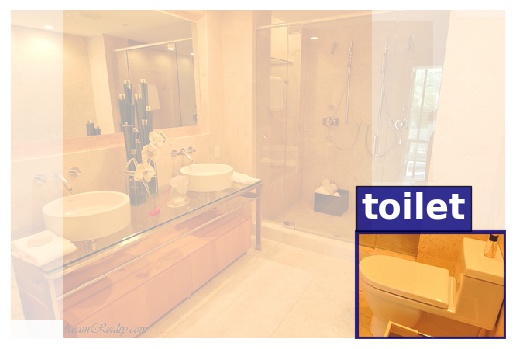} &
\includegraphics[width=0.124\linewidth]{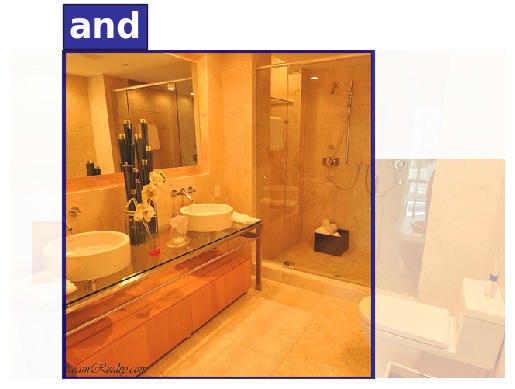} & 
\includegraphics[width=0.124\linewidth]{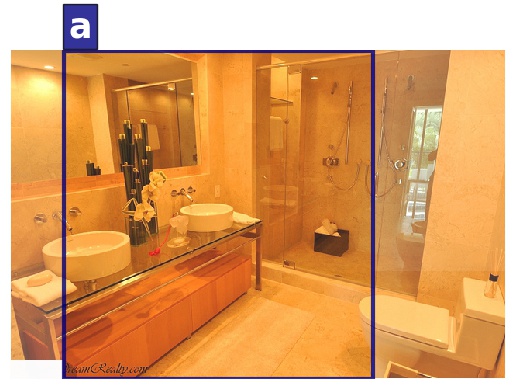} & 
\includegraphics[width=0.124\linewidth]{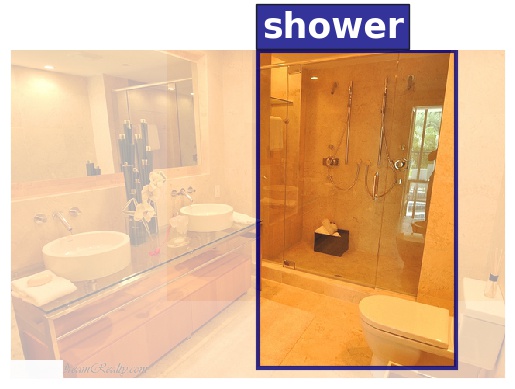} \\

\addlinespace[0.2cm]
\multicolumn{8}{c}{\textbf{Up-Down~\cite{anderson2018bottom}}} \\
\includegraphics[width=0.124\linewidth]{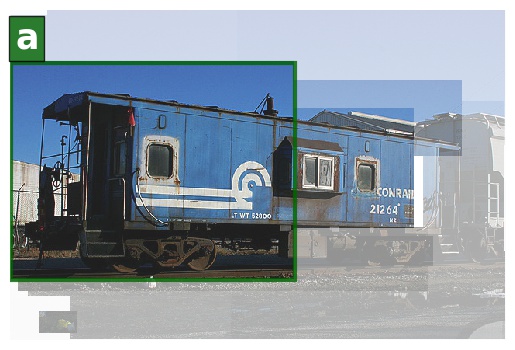} & 
\includegraphics[width=0.124\linewidth]{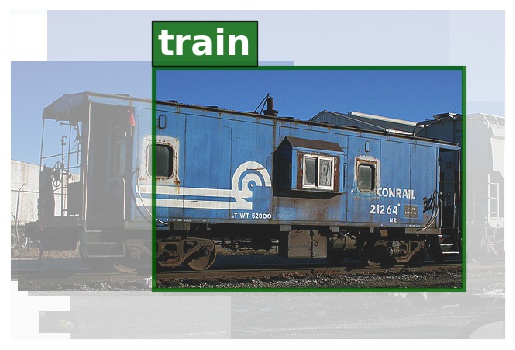} & 
\includegraphics[width=0.124\linewidth]{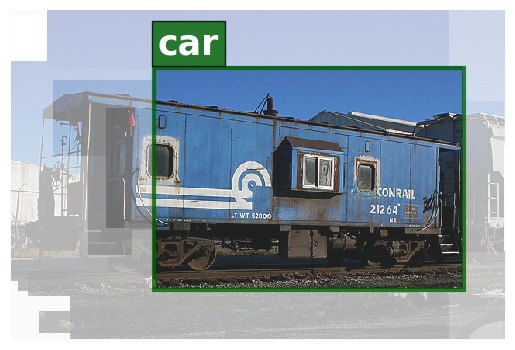} & 
\includegraphics[width=0.124\linewidth]{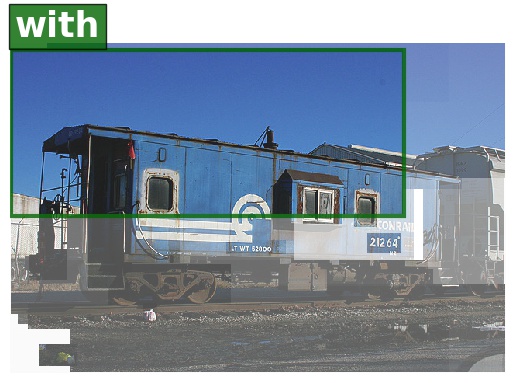} & 
\includegraphics[width=0.124\linewidth]{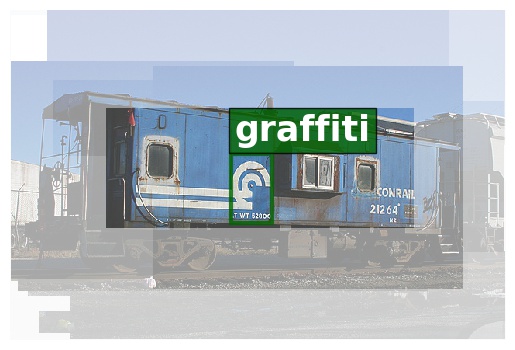} &
\includegraphics[width=0.124\linewidth]{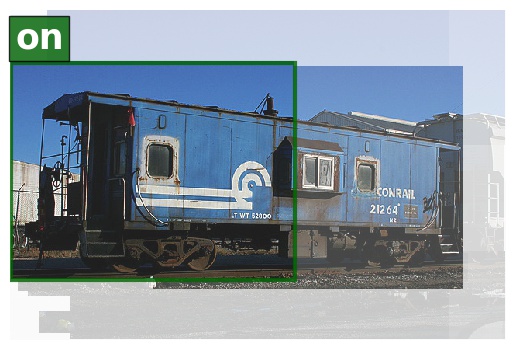} & 
\includegraphics[width=0.124\linewidth]{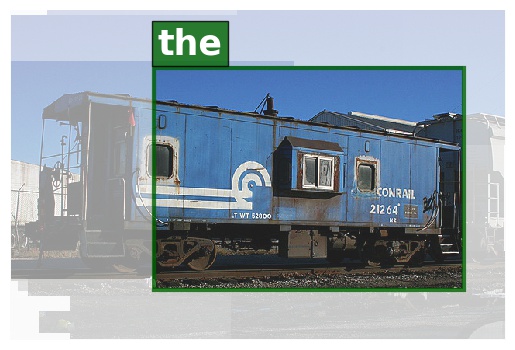} & 
\includegraphics[width=0.124\linewidth]{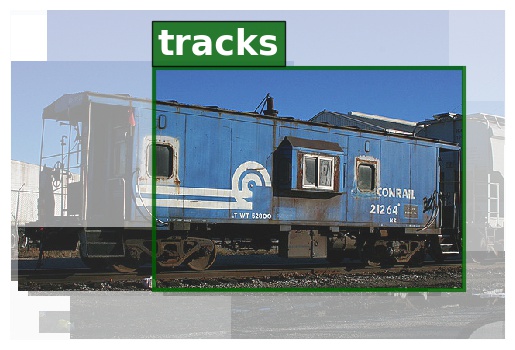} \\
\addlinespace[0.1cm]
\multicolumn{8}{c}{\textbf{$\mathbf{\mathcal{M}^2}$ Transformer~\cite{cornia2020meshed}}} \\
\includegraphics[width=0.124\linewidth]{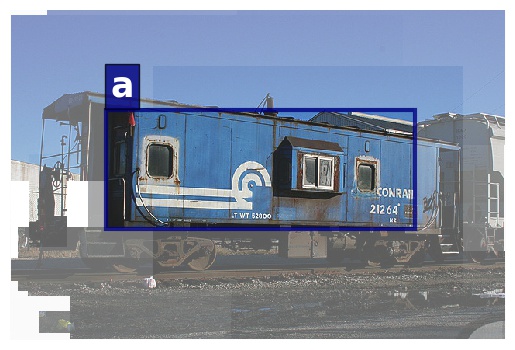} & 
\includegraphics[width=0.124\linewidth]{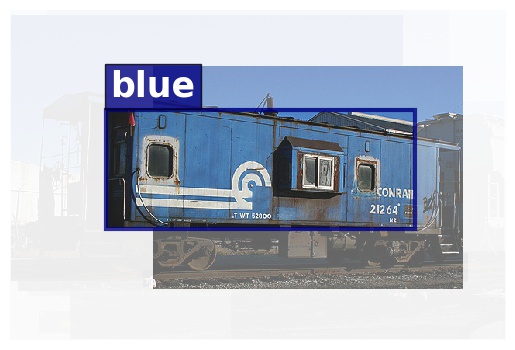} & 
\includegraphics[width=0.124\linewidth]{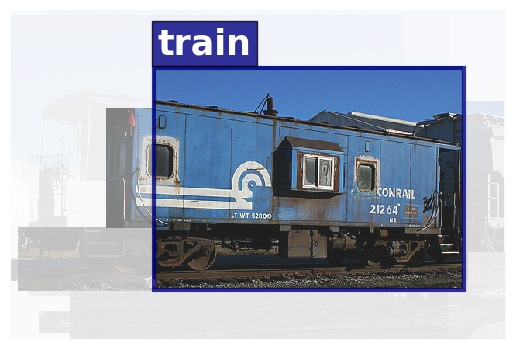} & 
\includegraphics[width=0.124\linewidth]{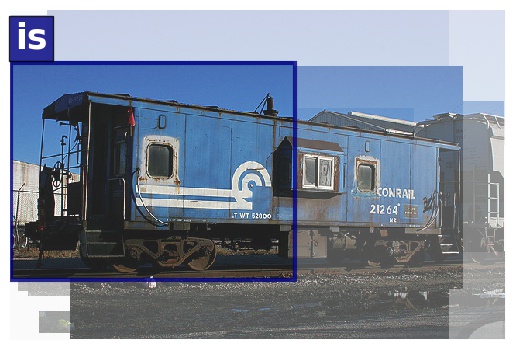} & 
\includegraphics[width=0.124\linewidth]{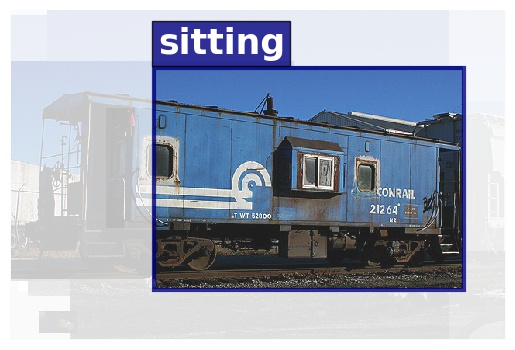} &
\includegraphics[width=0.124\linewidth]{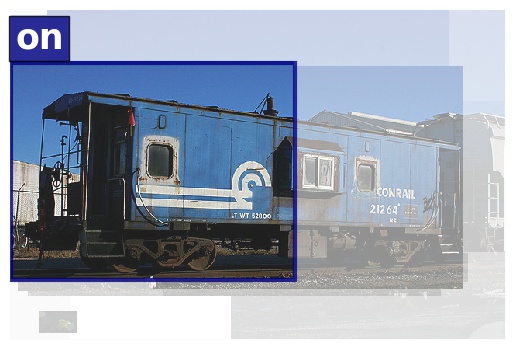} & 
\includegraphics[width=0.124\linewidth]{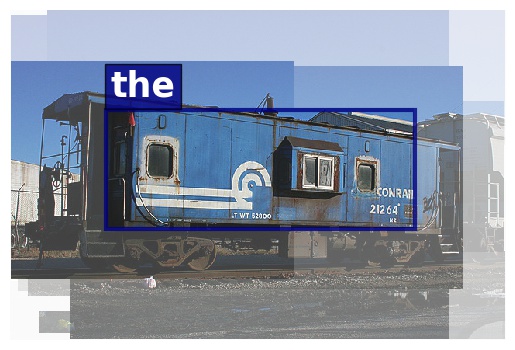} & 
\includegraphics[width=0.124\linewidth]{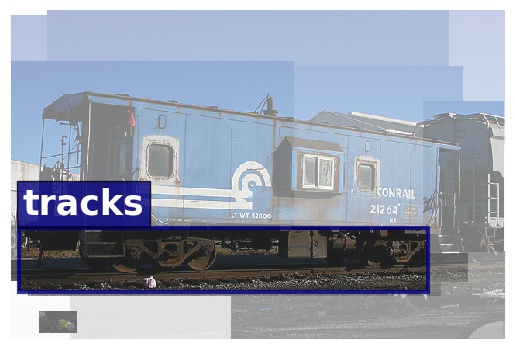} \\

\addlinespace[0.2cm]
\multicolumn{8}{c}{\textbf{Up-Down~\cite{anderson2018bottom}}} \\
\includegraphics[width=0.124\linewidth]{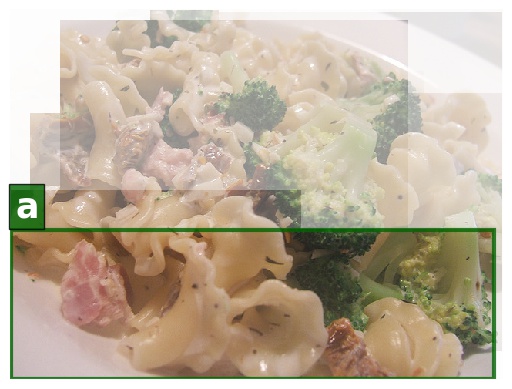} & 
\includegraphics[width=0.124\linewidth]{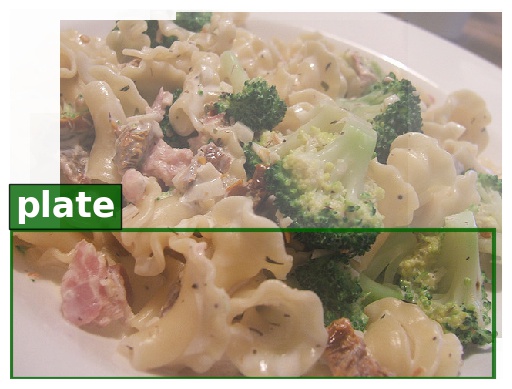} & 
\includegraphics[width=0.124\linewidth]{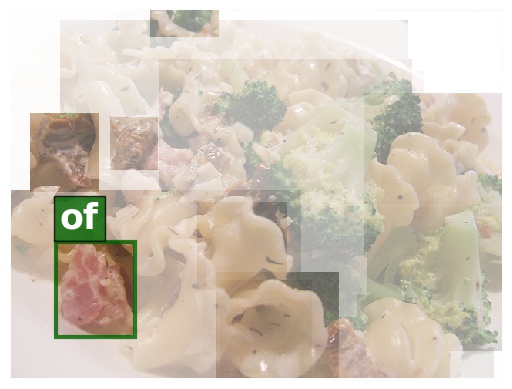} & 
\includegraphics[width=0.124\linewidth]{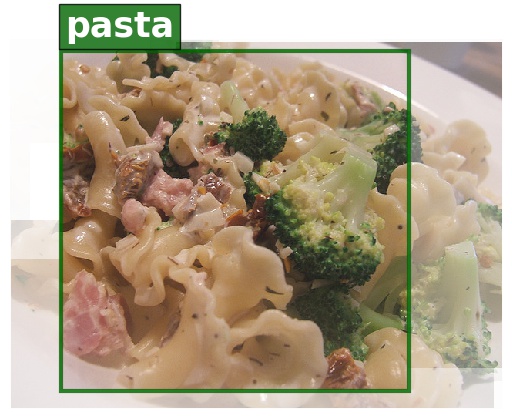} & 
\includegraphics[width=0.124\linewidth]{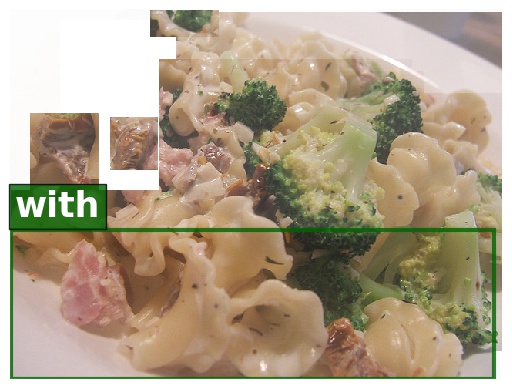} &
\includegraphics[width=0.124\linewidth]{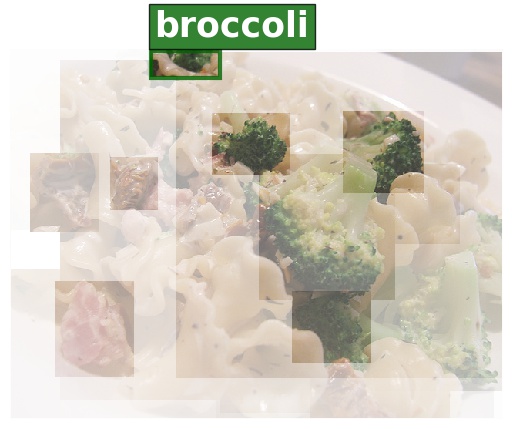} & 
\includegraphics[width=0.124\linewidth]{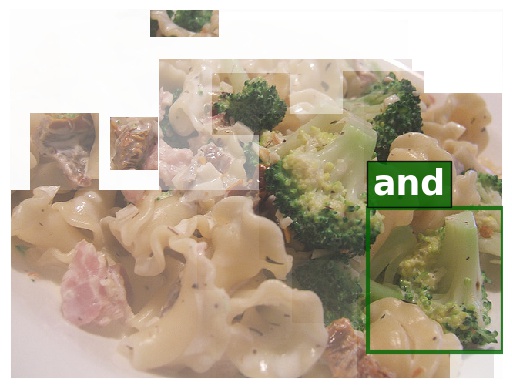} & 
\includegraphics[width=0.124\linewidth]{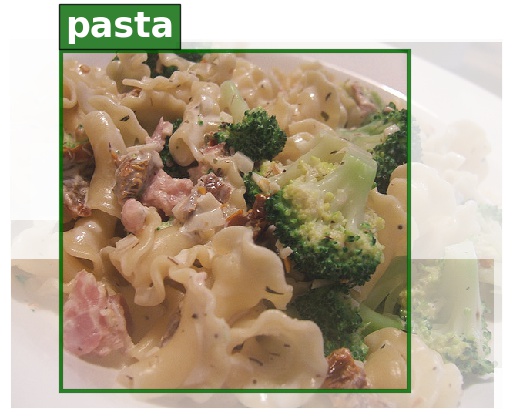} \\
\addlinespace[0.1cm]
\multicolumn{8}{c}{\textbf{$\mathbf{\mathcal{M}^2}$ Transformer~\cite{cornia2020meshed}}} \\
\includegraphics[width=0.124\linewidth]{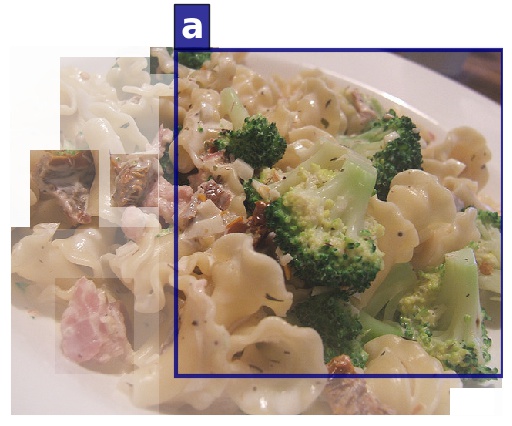} & 
\includegraphics[width=0.124\linewidth]{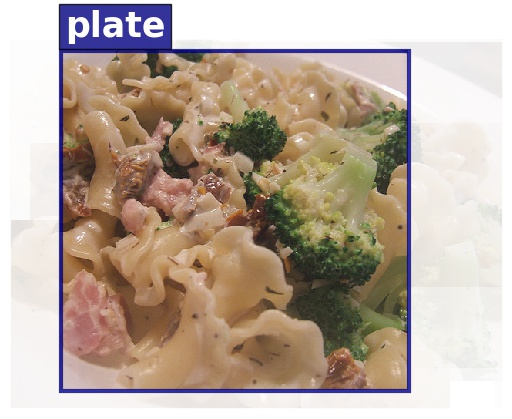} & 
\includegraphics[width=0.124\linewidth]{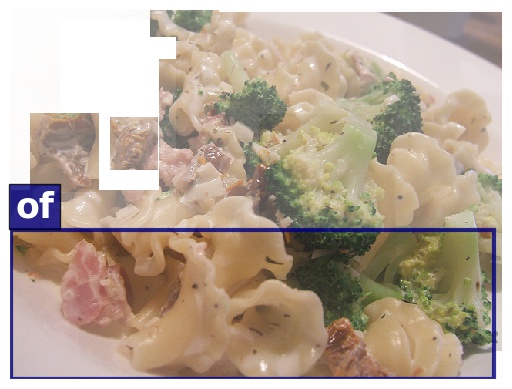} & 
\includegraphics[width=0.124\linewidth]{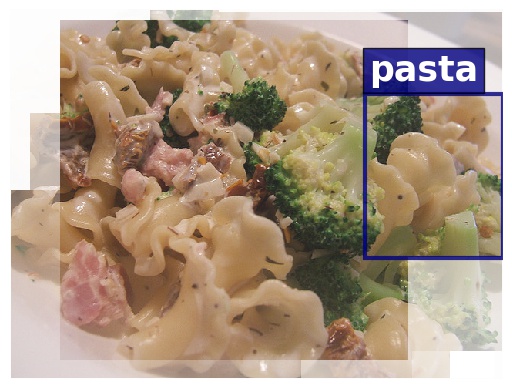} & 
\includegraphics[width=0.124\linewidth]{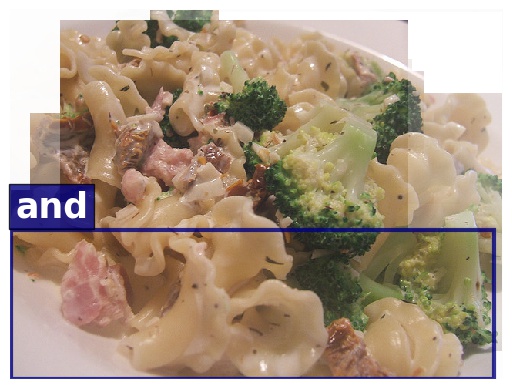} &
\includegraphics[width=0.124\linewidth]{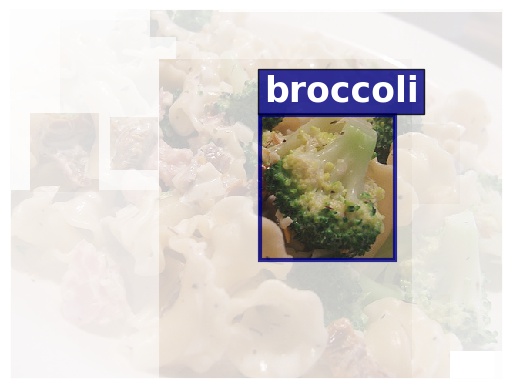} & 
\includegraphics[width=0.124\linewidth]{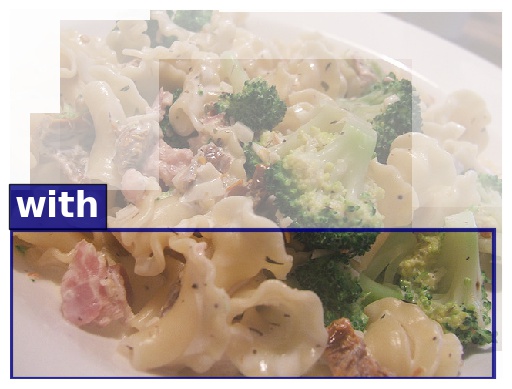} & 
\includegraphics[width=0.124\linewidth]{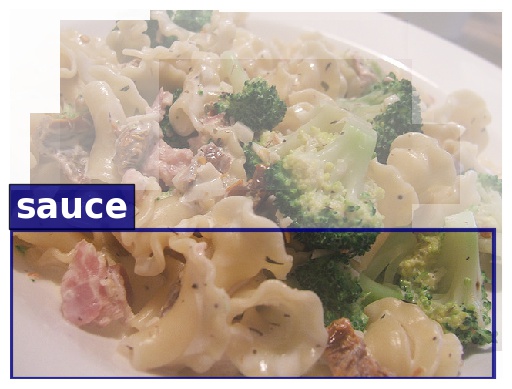} \\

\addlinespace[0.2cm]
\multicolumn{8}{c}{\textbf{Up-Down~\cite{anderson2018bottom}}} \\
\includegraphics[width=0.124\linewidth]{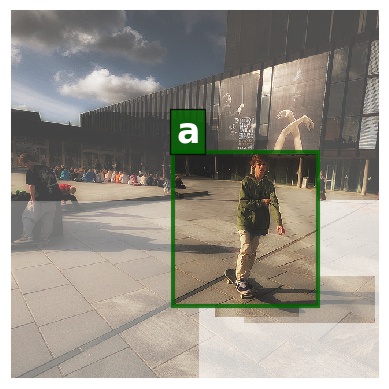} & 
\includegraphics[width=0.124\linewidth]{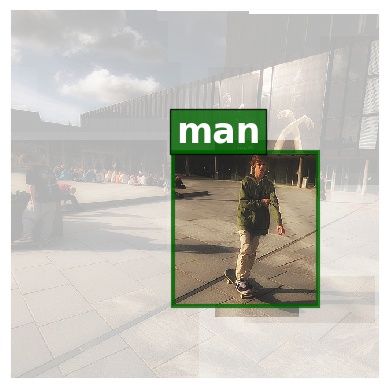} & 
\includegraphics[width=0.124\linewidth]{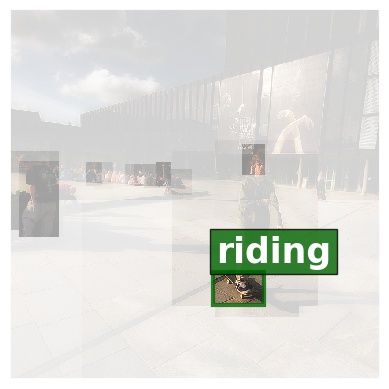} & 
\includegraphics[width=0.124\linewidth]{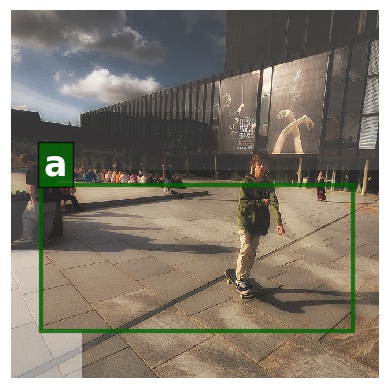} & 
\includegraphics[width=0.124\linewidth]{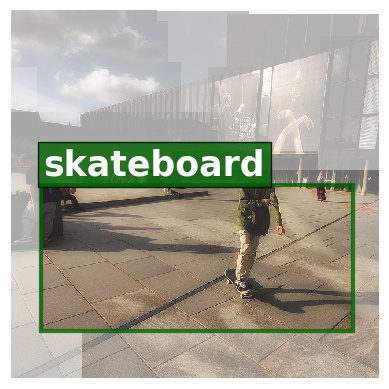} &
\includegraphics[width=0.124\linewidth]{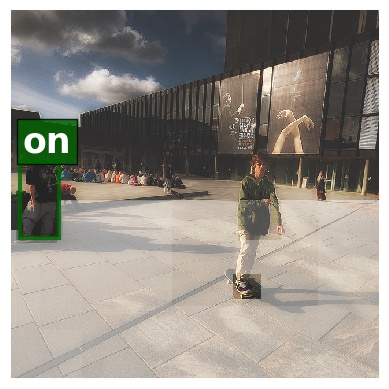} & 
\includegraphics[width=0.124\linewidth]{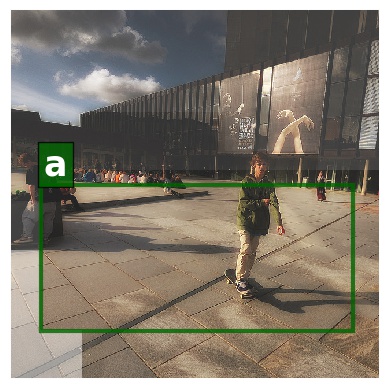} & 
\includegraphics[width=0.124\linewidth]{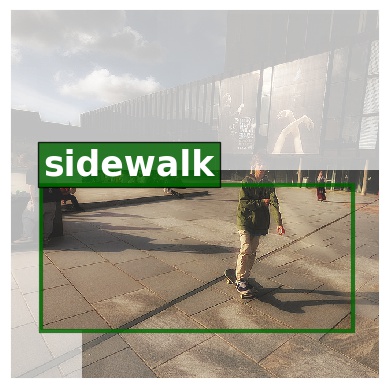} \\
\addlinespace[0.1cm]
\multicolumn{8}{c}{\textbf{$\mathbf{\mathcal{M}^2}$ Transformer~\cite{cornia2020meshed}}} \\
\includegraphics[width=0.124\linewidth]{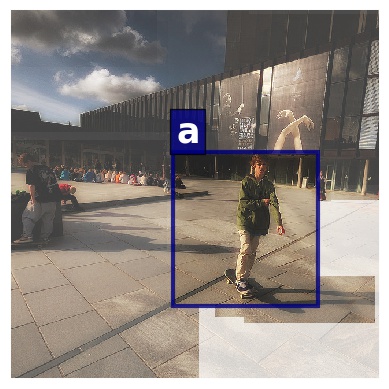} & 
\includegraphics[width=0.124\linewidth]{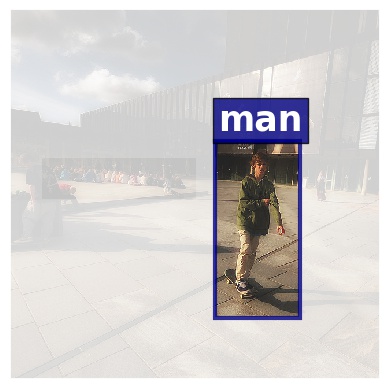} & 
\includegraphics[width=0.124\linewidth]{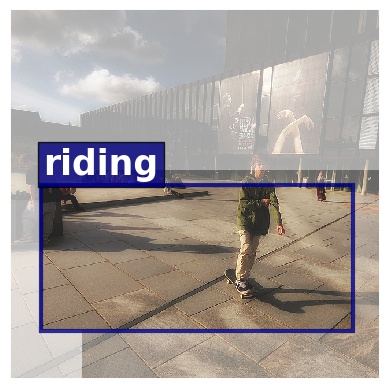} & 
\includegraphics[width=0.124\linewidth]{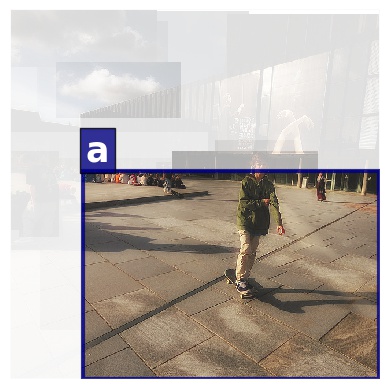} & 
\includegraphics[width=0.124\linewidth]{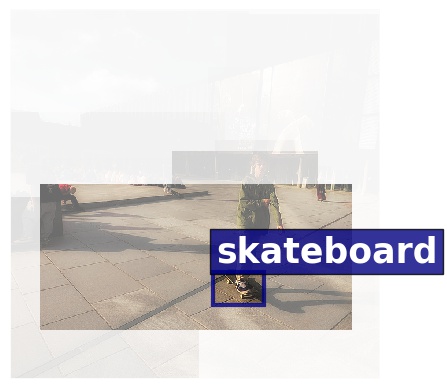} &
\includegraphics[width=0.124\linewidth]{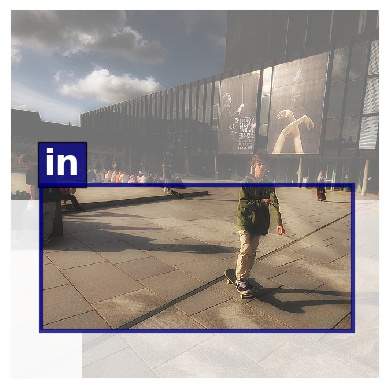} & 
\includegraphics[width=0.124\linewidth]{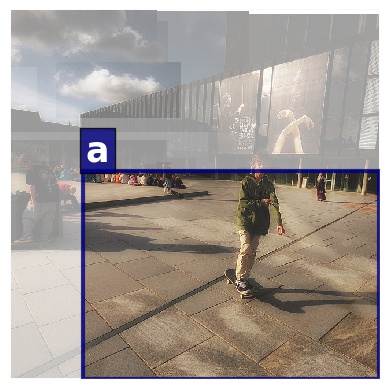} & 
\includegraphics[width=0.124\linewidth]{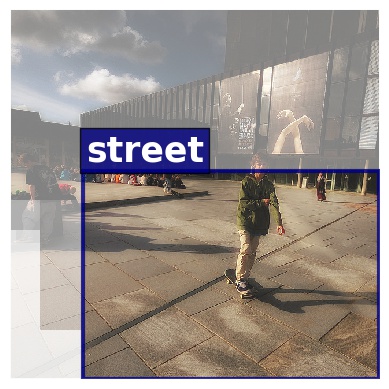} \\
\end{tabular}
}
\caption{Visualization of attention states for sample captions generated by Up-Down~\cite{anderson2018bottom} and $\mathcal{M}^2$ Transformer~\cite{cornia2020meshed}. For each generated word, we show the attended image regions, outlining the region with the maximum output attribution in green and blue, respectively.}
\label{fig:attention1}
\end{figure*}